\documentclass[letterpaper]{article} 
\usepackage{aaai2027}  
\usepackage[hyphens]{url}  
\usepackage{graphicx} 
\usepackage{natbib}  
\usepackage{caption} 
\usepackage{algorithm}
\usepackage{algorithmic}

\usepackage{newfloat}
\usepackage{listings}
\DeclareCaptionStyle{ruled}{labelfont=normalfont,labelsep=colon,strut=off} 
\floatstyle{ruled}
\newfloat{listing}{tb}{lst}{}
\floatname{listing}{Listing}

\usepackage{booktabs}

\usepackage{subcaption}
\usepackage{tcolorbox}
\tcbuselibrary{skins, breakable, theorems}
\usepackage{amsmath}
\usepackage{amssymb}
\usepackage{amsthm}
\usepackage{booktabs}
\usepackage{framed} 
\usepackage{multirow}
\usepackage{enumitem}
\usepackage{caption}
\usepackage{etoolbox}
\usepackage{xspace}
\usepackage{accents}
\usepackage{subfig}
\usepackage{colortbl}
\usepackage{xcolor}
\usepackage{arydshln}
\usepackage{array} 
\usepackage{booktabs}
\usepackage{makecell}
\usepackage{tabularx} 
\usepackage{calc}

\usepackage{graphicx}

\usepackage{lipsum}

\title{Slides2MindMap: Reconstructing Cognitively Efficient Knowledge Hierarchies from Lecture Slides}
\author{
    Yuzhi Wang\textsuperscript{\rm 1}, 
    Rongjun Ye\textsuperscript{\rm 1},
    Shengyuan Chen\textsuperscript{\rm 1}\corresponding,
    Huachi Zhou\textsuperscript{\rm 1},
    Jiaqi Bai\textsuperscript{\rm 1},
    Chuang Zhou\textsuperscript{\rm 1},
    Zhicong Hong\textsuperscript{\rm 2},
    Xiao Huang\textsuperscript{\rm 1}
}  

\affiliations{
    \textsuperscript{\rm 1}The Hong Kong Polytechnic University,
    \textsuperscript{\rm 2}Shanghai Jiaotong University\\

    \{yu-zhi.wang, rongjun.ye, chuang-qqzj.zhou\}@connect.polyu.hk,\\
    \{sheng-yuan.chen, hc0223.zhou, jacky.bai, xiao.huang\}@polyu.edu.hk,
    arrow23@sjtu.edu.cn
}

\begin{document}

\maketitle

\begin{abstract}
Generating mind maps from lecture slides can help learners efficiently assimilate fragmented knowledge, promising substantial benefits for intelligent education. However, dedicated automatic generation and evaluation frameworks remain underexplored and challenging, requiring a global-local knowledge focus balance and handling large-scale, heterogeneous slides. We formulate the \textbf{Slides2MindMap} task, which aims to reconstruct cognitively efficient knowledge hierarchies from a course's slide deck collection. For systematic evaluation, we introduce \textbf{S2M-Bench}, a benchmark comprising 12,774 slide pages with expert-annotated mind maps spanning 24 university courses. S2M-Bench includes a cognitive-science-grounded evaluation framework that integrates ground-truth-based comparison, structure conformity analysis, and VLM-as-a-Judge. To address this task, we propose \textbf{AutoMindMap}, an agentic framework inspired by the Structure Building Framework. AutoMindMap comprises \textit{Skeleton Laying} for global scaffold anchoring, \textit{Iterative Knowledge Integration} augmented by context-aware summarization, and \textit{Dual-Stage Refinement} with a local-global decoupling mechanism. The framework reconciles local knowledge faithfulness with global coherence, and adapts to slide-specific features. Experiments on S2M-Bench demonstrate that AutoMindMap outperforms baselines and achieves superior robustness across different models and scenarios, underscoring its pedagogical application value.

\end{abstract}

\begin{links}
    \link{Code}{https://github.com/YZWANG02/Slides2MindMap}
\end{links}

\section{Introduction} 

Mind maps play a significant role in modern education scenarios, revealing the progressive and hierarchical knowledge logic, and offering a visual scaffold that mirrors the associative nature of human cognition in learning~\citep{buzan2006mind}. Grounded in Advance Organizer Theory~\citep{ausubel1960use} and Cognitive Load Theory~\citep{sweller2011cognitive}, mind maps effectively facilitate the assimilation of fragmented knowledge~\citep{raiyn2016role}, and enable more pedagogical scenarios such as lecture planning~\citep{zhang2025awaking} and exercise generation~\citep{dong2024clr}. Despite the pedagogical value, constructing high-quality mind maps from course materials remains labor-intensive and expertise-demanding, motivating the pursuit of automated generation~\citep{zhang2024coreference}, particularly for \textbf{slide-oriented mind mapping}~\citep{low2025boommapper}. Unlike well-structured textbooks, slides are often interspersed with visually noisy and pedagogy-driven content, lacking explicit hierarchical organization and narrative continuity~\citep{tanaka2023slidevqa}. The nature risks learner misinterpretation and cognitive overload, and complicates automated knowledge extraction~\citep{kasinathan2024presentation}, thereby underscoring both the application necessity and the technical difficulty of the task.

\begin{figure}[t]
\centering
\includegraphics[width=1\linewidth]{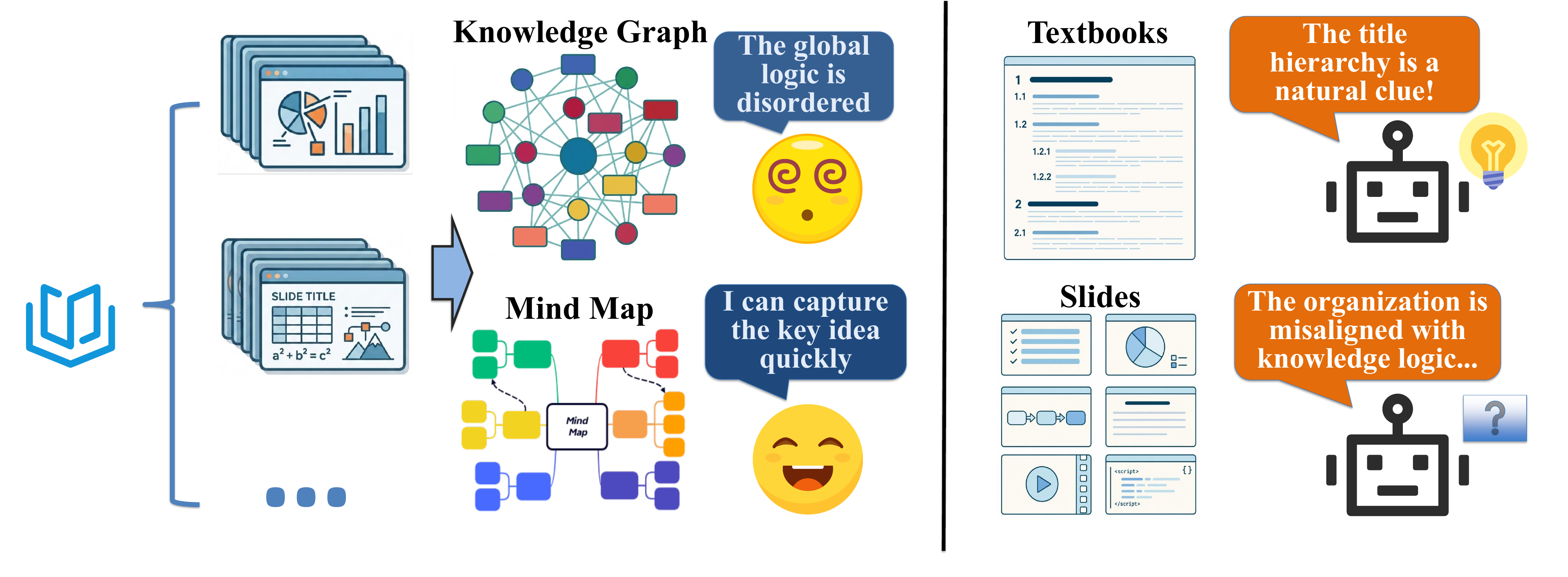}

\caption{Illustration of Slides2MindMap. \textit{Left}: mind maps provide more intuitive structure for visual learning, but need to strike a global-local balance. \textit{Right}: the flexible organization of slides makes explicit hierarchy extraction difficult.}

\label{fig:intro}
\end{figure}

Early supervised methods for mind map generation are only within simplified textual scenarios, and suffer from limited generalization and scalability~\citep{hu2021efficient,zhang2024coreference}. Recent large language models (LLMs) bring new opportunities for the task due to the powerful zero-shot extraction and summarization capacities~\citep{zhang2025systematic}. However, dedicated generation and evaluation frameworks remain underexplored, and fundamental obstacles still exist, as shown in Figure~\ref{fig:intro}. \textbf{First, mind maps present a construction paradigm gap with previous knowledge representation frameworks}. Previous ones, such as knowledge graphs (KGs), focus on dense triple extraction and high-frequency entity anchoring. Such a structure targets retrieval and inference tasks, but lacks conciseness and overall hierarchical logic for \textit{efficient human visual understanding}~\citep{zhu2024llms, zhang2025survey}. Comparatively, a cognitively efficient mind map should organize concepts into progressive and meaningful hierarchies around a central theme, distilling dense relational networks into an intuitive, branched structure that emphasizes core knowledge logic~\citep{hua2019exploring}. This principle demands a \textit{\textbf{balance} between local knowledge faithfulness and global logical coherence}, which existing LLM-based extraction and summarization methods struggle to strike.

\begin{figure}[t]
\centering
\includegraphics[width=1\linewidth]{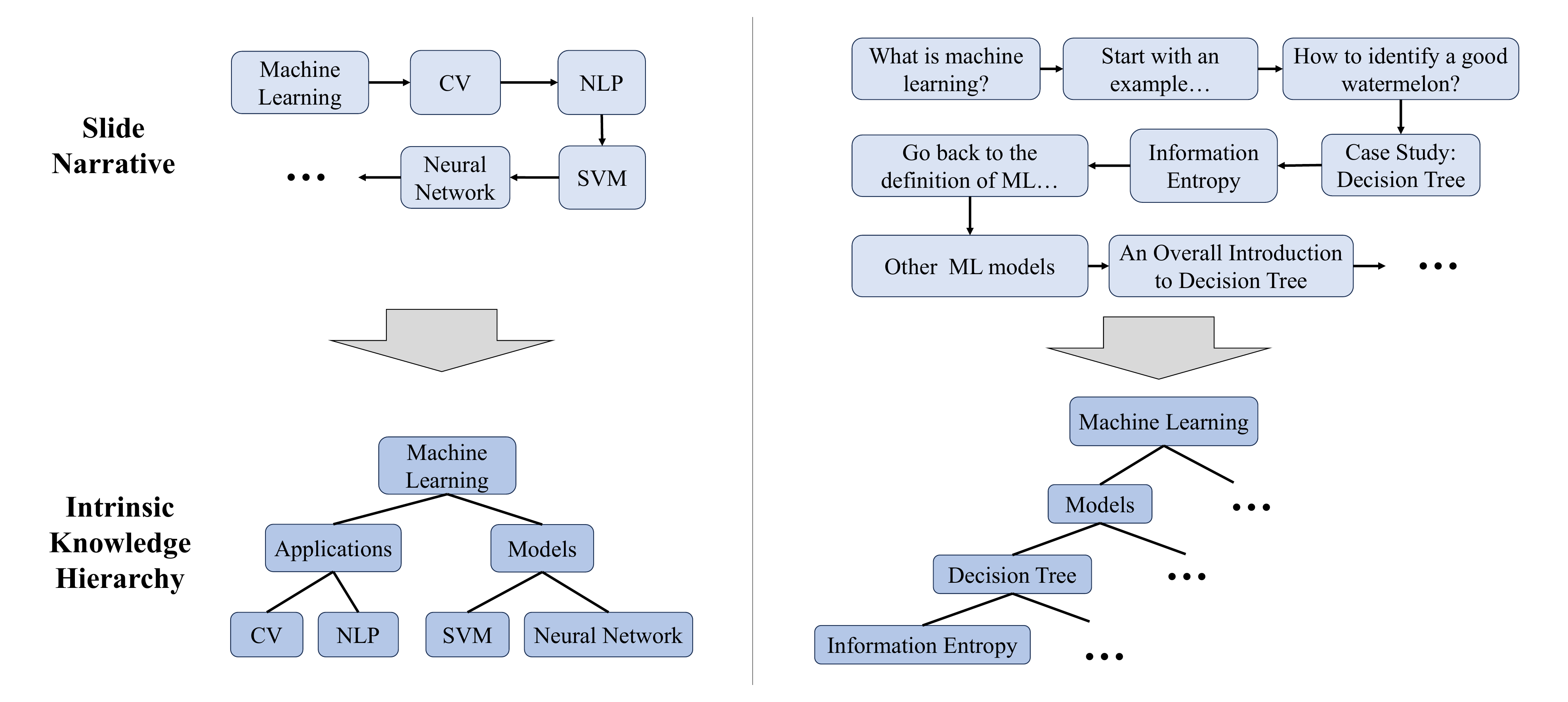}

\caption{Two typical cases of the heterogeneity. \textit{Left}: intermediate taxonomies of descendant concepts might be implicit and need to be inferred. \textit{Right}: a lot of noise and knowledge interleaving are for pedagogical purposes, and hierarchy reconstruction needs further reasoning.}

\label{fig:hete}
\end{figure}

\textbf{This challenge is compounded by slide-specific features}~\citep{tanaka2023slidevqa}. (1)~Slides show flexible organizations, lacking explicit hierarchical logic in linear and fragmented structures. As shown in Figure~\ref{fig:hete}, the narratives serve pedagogical functions (e.g., guidance, interpretation, association, and consolidation), and exhibit interleaving, repetition, and jumps~\citep{feng2020genre}, presenting a \textbf{systematic heterogeneity} with intrinsic knowledge logic. (2)~The slide collection of each course tends to be \textbf{large-scale} (e.g., with about 150K tokens in Figure~\ref{tab:benchstat1}), comprising numerous decks and hundreds of slides with implicit and interlaced knowledge association~\citep{low2025boommapper}, which further complicates comprehensive knowledge understanding, especially for end-to-end methods~\citep{liu2024lost}. Although recent document hierarchy indexing (DHI) methods like PageIndex~\citep{zhang2025pageindex} and BookRAG~\citep{wang2025bookrag} can generate analogous hierarchical structures, these methods rely heavily on well-structured documents and explicit formatting cues, thus are brittle in flexible and long-context slide scenarios. Accordingly, building coherent knowledge hierarchies from slides requires \textit{further reasoning and reconstruction rather than direct extraction and merging}.

The above observations raise \textbf{two central questions}: (1)~How to comprehensively evaluate the mind maps generated from slides, considering knowledge faithfulness, structure compliance, task openness, and cognitive efficiency? (2)~How to automatically build mind maps that faithfully reconstruct knowledge structures from fragmented and flexibly organized slides, while presenting cognitively efficient structures for visual understanding? To answer the questions, we first formulate the \textbf{\underline{S}lides\underline{2M}indMap} task, aiming to reconstruct explicit, coherent, and faithful knowledge hierarchies from a course's large-scale slide deck collection. For comprehensive evaluation, we introduce \textbf{S2M-Bench}, the first dedicated benchmark focusing on hierarchical knowledge reconstruction from authentic course slide collections, and targeting human cognitive effects. S2M-Bench comprises 12,774 slide pages across 24 university-level courses, with expert-annotated mind maps averaging 121.38 concepts per course. The courses are categorized by knowledge type and content modality to emphasize scenario diversity. Given the semi-open nature of the task, the evaluation framework is grounded in cognitive science and integrates ground-truth-based comparison, quantitative structure conformity, and VLM-as-a-Judge~\citep{chen2024mllmasajudgeassessingmultimodalllmasajudge}.

To address the task, we propose \textbf{AutoMindMap}, an LLM-based agentic framework inspired by the \textit{Structure Building Framework} (SBF) Theory~\citep{gernsbacher2013language} in cognitive psychology. We discard the traditional knowledge extraction paradigm and treat mind map construction as a \textit{dynamic discourse comprehension process} that mirrors human cognitive models. AutoMindMap emulates foundation laying, mapping, and shifting processes in SBF through a three-stage design: \textbf{(1)~Skeleton Laying} emulates \textit{foundation laying}, capturing salient elements to scaffold an overall knowledge skeleton, and plan guidelines for further map expansion. The planning-then-constructing strategy anchors global structure logic to prevent collapse. \textbf{(2)~Iterative Knowledge Integration} emulates \textit{mapping-shifting}, maintaining a reading memory state as cross-discourse semantic bridges through context-aware summaries, then incrementally expanding the map through atomic tools and structural repair. The design alleviates knowledge fragmentation, logic breaks, and invalid structure. \textbf{(3)~Dual-Stage Refinement} decouples macro-level (e.g., global balance and coherence) and micro-level (e.g., factual accuracy) issues for separate refinement, to avoid global-local mutual structural destruction. Macro-level refinement is performed with rendered map images for intuitive structural insight. Overall, the SBF-inspired design preserves global and local consistency, achieving secure \textit{harnessing} through structural anchoring, tool-based expansion, and refinement.

Our contribution can be summarized as follows: (1)~We establish S2M-Bench, the first dedicated benchmark for evaluating the lecture-slide-oriented automated knowledge hierarchy reconstruction task. (2)~We propose AutoMindMap, an agentic framework that effectively addresses the task by emulating the human discourse cognitive model. (3)~Extensive experiments on S2M-Bench show that AutoMindMap significantly outperforms baselines and exhibits strong generalization and robustness across different models and scenarios.

\section{S2M-Bench}
\label{sec:s2mbench}

\begin{table*}[htbp]
\centering
\small
\setlength{\tabcolsep}{5pt}
\begin{tabular}{lccccccccc}

\toprule
\textbf{Category} & \textbf{Courses} & \textbf{Decks/Co.} & \textbf{Pages/Co.} & \textbf{Pages/Deck} & \textbf{Tokens/Co.} & \textbf{Tokens/Page} & \textbf{Nodes/Co.} & \textbf{Edges/Co.} \\
\midrule
 Math \& Programming  & 7  & 11.29 & 516.14 & 45.73 & 146.24K & 283.32 & 121.71 & 136.57 \\
Concept Understanding  & 6  & 10.83 & 658.50 & 60.78 & 170.39K & 258.76 & 154.67 & 167.17 \\
 Synthetic Application   & 11 & 11.18 & 470.55 & 42.08 & 138.68K & 294.72 & 103.00 & 111.82 \\
\midrule
Text-dominated  & 6  & 9.33  & 388.00 & 41.57 & 106.05K & 273.33 & 138.50 & 147.33 \\
 Image-dominated  & 6  & 11.67 & 620.33 & 53.17 & 165.06K & 266.08 & 111.00 & 120.50 \\
Hybrid  & 12 & 11.75 & 557.50 & 47.45 & 162.07K & 290.71 & 118.00 & 131.83 \\
\midrule
 \textbf{Overall} & \textbf{24} & \textbf{11.12} & \textbf{530.83} & \textbf{47.72} & \textbf{148.81K} & \textbf{280.34} & \textbf{121.38} & \textbf{132.88} \\
\bottomrule

\end{tabular}

\caption{Statistics for different course categories. All columns are averages except \textit{Courses}. \textit{Co.} abbreviates \textit{Course}. The tokenization is based on \textit{o200k\_base}.}
\label{tab:benchstat1}
\end{table*}

This section elaborates S2M-Bench, a benchmark for evaluating the Slides2MindMap task. The sub-sections cover the problem definition, dataset construction and analysis, and the evaluation metrics.

\subsection{Problem Definition}
\textbf{Definition 1} (Mind Map).
A mind map $\mathcal{M}$~\citep{buzan2006mind} is a connected directed acyclic graph (DAG) defined as
$
\mathcal{M} = (V, E, r),
$
where $V$ is the set of concept nodes, $r \in V$ is an only \textbf{central} root node representing the core topic. $E \subseteq V \times V$ is a set of hierarchical edges, where $(v_i, v_j) \in E$ indicates that $v_j$ is a child of $v_i$, multiple parents of one node (termed \textbf{cross-links}) are permitted. Hierarchical edges indicate dominance relationships (e.g., hypernym–hyponym, whole–part, and entity–attribute). This paper excludes the map visual design~\citep{hua2019exploring} and only focuses on the knowledge structure.

\textbf{Definition 2} (Slides2Mindmap).
Given a course $C$ comprising $m$ slide decks $\mathcal{D}=\{D_1, \dots, D_m\}$, where $\mathcal{D}$ is not necessarily well-formed (e.g., without prerequisite or standard file name), and each deck $D_i$ consists of several pages containing heterogeneous elements (e.g., titles, images, paragraphs, equations, tables). The \textbf{Slides2MindMap} task aims to generate a mind map $\mathcal{M}$ with the course name as $r$, such that $\mathcal{M}$ can faithfully generalize the intrinsic knowledge hierarchy of $C$, and present a coherent and cognitively-efficient global structure for visual understanding.

\subsection{Dataset Construction and Quality Control}

The collected courses span university-level computer science and mathematics. To ensure \textbf{data quality}, we converted all decks to PDF, removed slides with formatting anomalies, and verified page completeness. To keep \textbf{particularity}, we deliberately selected courses that exhibit enough scale, flexibly organized structures, and rich hierarchical knowledge, while excluding those with excessively structured narratives or abstract concepts. To maximize \textbf{scenario diversity}, we categorized all courses along two dimensions (i.e., knowledge type and content modality) inspired by Bloom's taxonomy of educational objectives~\citep{armstrong2010bloom} and multi-modal education theory~\citep{hassett2009theories}. Following the categorization criteria (detailed in Appendix~\ref{app:dataset details}), we further selected 24 high-quality courses totaling 267 decks and 12,774 slides. Table~\ref{tab:benchstat1} summarizes the dataset categorization and statistics. For each course, we invited expert annotators, consisting of the course lecturer and PhD-level teaching assistants, to construct mind maps with unified instructions and rubrics. To ensure \textbf{annotation reliability}, we conducted cross-validation and multiple refinement rounds after initial annotation to eliminate redundant, ambiguous, and over-dense concepts, retaining only high-confidence elements that faithfully reflect the source material. This rigorous rule minimizes subjectivity. The annotation consumes over 200 hours. These human-annotated maps serve as \textbf{ground-truth} in subsequent experiments. Further analysis and taxonomy illustrations of the dataset are deferred to Appendix~\ref{app:dataset details}.

\subsection{Evaluation Metrics}
Although high-quality human annotation can be used as an evaluation reference, the characteristics of mind maps determine that the knowledge faithfulness, structure compliance, open-ended nature, and cognitive efficiency need to be comprehensively considered. Thus, we design a three-faceted evaluation framework.

\subsubsection{Ground-Truth (GT)-based Comparison.}
Let $\mathcal{M}^{\text{gt}} = (V^{\text{gt}}, E^{\text{gt}}, r^{\text{gt}})$ and $\mathcal{M}^{\text{pred}} = (V^{\text{pred}}, E^{\text{pred}}, r^{\text{pred}})$ denote the ground-truth and generated maps, respectively. We first obtain a matched node-pair set $\mathcal{C} \subseteq V^{\text{pred}} \times V^{\text{gt}}$ via a matching algorithm (detailed in Appendix~\ref{app:matchingalgo}). To emphasize the trade-off capacity, we design the F1-score from three aspects: \textbf{(1)~Hierarchy-aware Node F1-Score~(HN-F1)} assigns each matched pair a weight that penalizes depth misalignment based on standard F1-score calculation, thus better reflecting the ability to identify the knowledge hierarchy:
$
w(v_i^{\text{gt}}, v_i^{\text{pred}}) = 1 - \left| \frac{d(v_i^{\text{gt}})}{d^{\text{gt}}} - \frac{d(v_i^{\text{pred}})}{d^{\text{pred}}} \right|,
$
where the depth $d(v)$ denotes the shortest-path distance from the root $r$, and $d^{\text{gt}} = \max_{v \in V^{\text{gt}}} d(v)$, $d^{\text{pred}} = \max_{v \in V^{\text{pred}}} d(v)$ are the respective maximum depths. 
\textbf{(2)~Edge Connectivity F1-Score~(EC-F1)} evaluates edge-level similarity and reflects the ability to identify hierarchical relations between concepts. Considering the open-ended nature, we use directed reachability instead of strict edge matching to tolerate reasonable intermediate nodes. \textbf{(3)~Matched-Edge Connectivity F1-Score~(MEC-F1)} further restricts the denominator of EC-F1 to edges whose both endpoints are in the matched set $\mathcal{C}$, thus isolating the evaluation to the structurally shared concept backbone. The metric is more stringent and better reflects the quality of the aligned hierarchical skeleton. The detailed calculations of the above three metrics are in Appendix~\ref{app:gtcal}.

\subsubsection{Quantitative Structure Conformity.}
This aspect measures whether the generated mind map is \textit{structurally compliant} through quantifiable indicators. \textbf{(1)~Pass@1} serves as a \textit{hard constraint}: a generated map \textit{passes} if containing exactly one root, no isolated nodes, and no cycles. This metric verifies only the most \textit{fundamental structural requirements} of a valid mind map~\citep{buzan2006mind}. We test the strictest condition that only runs the algorithm once. \textbf{(2)~Structural Load Factor (SLF)} serves as a \textit{soft constraint}, which is grounded in the principles of working memory capacity~\citep{miller1956magical} and progressive differentiation~\citep{ausubel1960use}, quantifying cognitive load imposed by structural anomalies. The metric is derived from existing structural constraint rubrics~\citep{kedaj2014effective} and is defined as the average proportion of \textit{\underline{ab}normal} leaf nodes and non-leaf nodes:
$
\text{SLF} = \frac{1}{2}(\frac{|V_{\text{leaf}}^{\text{ab}}|}{|V_{\text{leaf}}|}+\frac{|V_{\text{non-leaf}}^{\text{ab}}|}{|V_{\text{non-leaf}}|}),
$
where a leaf node $v$ is abnormal if $d(v)=1$ or $d(v))\ge 7$, and abnormal non-leaf nodes refer to the nodes with more than 7 direct child nodes. 

\subsubsection{VLM-as-a-Judge.}
Considering the open-ended nature and the visual pedagogical effectiveness, we employ a vision-language model (VLM), GPT-5.4, as an evaluator to simulate the human learner, scoring mind maps using a 5-level Likert scale~\citep{nemoto2014likert} based on both \textit{textual relation} lists and \textit{rendered images}. Grounded in established cognitive science theories (detailed in Appendix~\ref{app:theory}) and adapted from existing rubrics~\citep{hua2019exploring}, we propose three dimensions: \textbf{(1)~Content Accuracy (CA)} measures whether the method can extract correct, concise, and unambiguous concept representations with reasonable taxonomies from complex documents. \textbf{(2)~Structure Quality (SQ)} examines structural soundness. Guided by principles of cognitive balance~\citep{Sweller2019CognitiveAA} and working memory capacity, this metric focuses on coherent hierarchy progressiveness, branch balance, and appropriate scale. \textbf{(3) Pedagogical Efficiency (PE)} measures cognitive-level utility for human learners, including alignment with human cognitive habits to minimize confusion, featuring clear visual rendering, appropriate scale and granularity, relevant cross-links, and distinct structural emphasis on core knowledge for efficient comprehension.

Detailed criteria of VLM-as-a-Judge evaluation are in Appendix~\ref{app:rubrics}. We make further analysis on the evaluation reliability in Appendix~\ref{app:Evaluation Reliability Analysis}.

\section{AutoMindMap}

\begin{figure*}[t]
\centering
\includegraphics[width=1\linewidth]{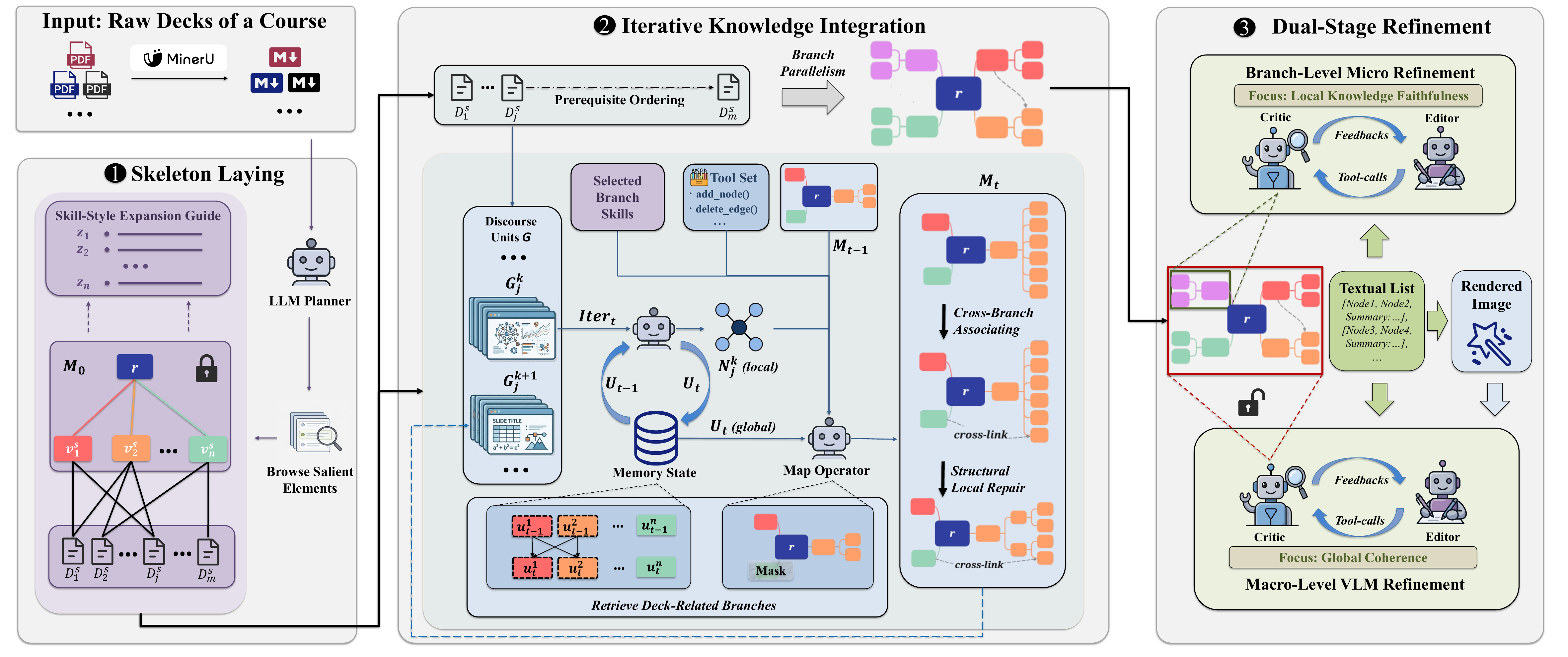}

\caption{Overall framework of \textit{AutoMindMap}. The main pipeline receives the parsed Markdown files of a course, then lays the initial skeleton with key associative information. The mind map is incrementally expanded with memory and tool-augmented integration. The map is further refined by an editor-critic framework that decouples macro-level and micro-level repair.}

\label{fig:frame}
\end{figure*}

The challenges encountered in Slides2MindMap bear similarities to human discourse comprehension. The SBF theory reveals that humans organize large-scale discourse into a coherent knowledge system through a cognitive model. The model involves laying a foundational understanding before reading, incrementally mapping coherent information into the understanding system, and shifting to new substructures when coherence breaks. Drawing inspiration from this cognitive model, AutoMindMap mirrors these modules through three concrete stages:\textbf{ Skeleton Laying},\textbf{ Iterative Knowledge Integration}, and \textbf{Dual-Stage Refinement}; each is further adapted to slide-specific obstacles through tailored mechanisms described below. These modules form a top-down global structuring and bottom-up gradual refinement to pursue a global-local balance. The overall framework is shown in Figure~\ref{fig:frame}.

Before the pipeline, we first employ MinerU~\citep{wang2024mineru} to parse raw PDF decks into Markdown format. Images are transformed into descriptive summaries extracted by the VLM, and each page is segmented and wrapped into “\textit{<page\_i> ... </page\_i>}” style to facilitate understanding.

\subsection{Skeleton Laying}
The stage mirrors the foundation laying in SBF, generating the course's overall understanding and planning the subsequent stages before the construction. For a course $C$ consisting of decks $\mathcal{D} = \{D_1, \dots, D_m\}$, we extract salient elements $K_i$ (e.g., file meta data, page titles) from each deck $D_i$, and ask LLM planner to produce: (1) Backbone second-level nodes $\mathcal{V}^s = \{v_1^s, \dots, v_n^s\}$ directly connecting to $r$, to formulate the \textbf{initial skeleton}, which represents the top-level and most abstract concept set with proper size that outlines the course. The skeleton is locked before the final macro refinement. (2) A bipartite graph $B_{(V,D)} \subseteq \mathcal{V}^s \times \mathcal{D}$, mapping each $v_i^s \in \mathcal{V}^s$ to relevant decks to build content logical association between decks and second-level nodes. (3) A set $\mathcal{Z}=\{z_1,\dots,z_n\}$ where each $z_i$ is an expansion guideline for $v_i^s$ and serves as a skill-style~\citep{xu2026agent} principle for downstream branching. $\mathcal{Z}$ includes anticipated descendant structure, risks to be avoided, and actionable suggestions. (4) A prerequisite sequence $\mathcal{S} = (D_1^s, \dots, D_m^s)$ guides sequential and parallel processing. 

$\mathcal{V}^s$ and $\mathcal{Z}$ scaffold a global anchor for subsequent generation, and prevent the subsequent map expansion from structure shift and loss of overall knowledge architecture sight; $B_{(V,D)}$ and $\mathcal{S}$ bridge cross-deck logic, avoiding the structural disorder caused by document-knowledge heterogeneity.

\subsection{Iterative Knowledge Integration}
The integration mirrors the mapping (intra-branch) and shifting (cross-branch) processes. Concepts are incrementally assimilated via summarization-enhanced atomic tool calls and local structural adjustment. Following $\mathcal{S}$, we ask the LLM to segment pages based on discourse semantics ~\citep{feng2020genre} instead of format clues. Each deck $D_j$ is partitioned into contiguous page groups $\mathcal{G}_j = \{G_j^1,\dots, G_j^q\}$ based on $K_j$, where each group constitutes a coherent discourse unit (e.g., with the same pedagogical topic or rhetorical move). This strategy respects narrative boundaries and mitigates logic breaks. Then the following steps are executed iteratively:

Firstly, for group $G_j^k$ processed in $t^{th}$ iteration, LLM-based extraction and summarization are performed sequentially:
\begin{equation}
    (\mathcal{N}_j^k,\ \mathcal{U}_{t}) = \Gamma(G_j^k,\ \mathcal{U}_{t-1}),
\end{equation}
where $\Gamma$ denotes the LLM operation, $\mathcal{N}_j^k = (V_j^k, E_j^k)$ is the extracted local concept relation subgraph, $\mathcal{U}_t = \{u_t^i \mid (v_i^s, D_j) \in B_{(V,D)}\}$ is a dynamic memory state that evolves across iterations. Each $u_t^i$ is anchored to $D_j$-relevant $v^s_i$ branches and retains the essence of prior content, reconstructing knowledge logic continuity from the pedagogy sequence. Critically, $\mathcal{U}_t$ provides the contextual scaffold for interpreting $\mathcal{N}_j^k$, and the synergies bridge the branch-level fragmented local discourses~\citep{o1998updating}. This branch-isolated design avoids context explosion and collapse, while preserving intra-branch coherence.

Second, the integration of $\mathcal{N}_j^k$ into the global mind map $\mathcal{M}_t$ is formalized as
\begin{equation}
\mathcal{M}_{t} = \Theta(\mathcal{N}_j^k,\ Z_t,\ \mathcal{M}_{t-1}',\ \mathcal{U}_{t}),
\end{equation}
where $\Theta$ is the integration operation, $Z_t = \{z_i \mid (v_i^s, D_j) \in B_{(V,D)}\}$ controls the update with prior guidance, and only retrieves deck-related $z_i$. $\mathcal{M_\text{0}}$ is the initial skeleton. $\mathcal{M}_{t-1}' \subseteq \mathcal{M}_{t-1}$ is a \textit{partial view} of the whole map that exposes only the branches $\mathcal{M}_{t-1}(v_i^s)$ under $v_i^s$ with $(v_i^s, D_j) \in B_{(V,D)}$. $\mathcal{M}_{t-1}'$ masks unrelated branches and constrains the LLM's context to the most relevant structural and semantic neighborhood, avoiding distraction by unrelated noise knowledge. Then $\mathcal{M}_t$ is further improved by the following two steps:

\textbf{Cross-Branch Associating} aims to mitigate the risk of cross-link loss caused by $\mathcal{M}_{t-1}'$. For each new updated node $v_{w}$, we retrieve candidate node set $V_{c}$ from the nodes outside $\mathcal{M}_{t-1}'$ that each node $v_{c}^i \in V_{c}$ satisfies $ v_{c}^i \in TopK(\text{Sim}(v_{w}, v))_{v \notin \mathcal{M}_{t-1}'}$ and $\text{Sim}(v_{w}, v_{c}^i)\geq \gamma_{sim}$, where $\text{Sim}(\cdot,\cdot)$ is performed on the text embeddings \textit{cosine similarity} and token-level \textit{Jaccard similarity} separately to capture lexical and semantic association. The neighbor subgraphs of $v_{w}$ and $v_{c}^i$ are then provided to the LLM to judge inter-branch conceptual dependencies and perform potential merging and cross-linking operations.

\textbf{Structural Local Repair} checks structural legitimacy (e.g., loops and broken structure) and health (e.g., abnormal nodes and confusing relations) issues through rules and the LLM, to identify risky structures around each $v_{w}$ for further repair and prevent the accumulation of structural errors.

Then the pipeline will return to the first step to process the next group (or next deck). For acceleration, decks that have disjoint second-level node assignments are processed in parallel, and decks under the same second-level node are processed based on $S$. 

Notably, all operations concerning the mind map update are implemented via reasoning-enhanced atomic tool calls (listed in Appendix~\ref{app:tool}), with compliance checks on parameter validity and structural constraints, and stage-aware tool isolation. The mechanism constrains the mind map generation to verifiable and controllable graph editing, mitigating collapsing structures caused by hallucination.

\subsection{Dual-Stage Refinement}
Even with careful incremental integration and checking, the final mind map may exhibit residual structural imperfections such as overgrown branches, guideline violations, and suboptimal global balance. However, direct refinement may risk local-global mutual structural destruction, which can further cause repeated modifications and structure collapse. We address the issues through an extra \textbf{Editor-Critic} refinement framework, which decouples micro-level local anomaly repair and macro-level global coherence improvement.

First, each second-level branch $\mathcal{M}(v_i^s)$ is refined independently. A critic agent examines the branch and produces feedback; then an editor agent executes structural edits via tool calls: 
\begin{equation}
    F_i^{(t)} = \pi_{\text{critic-mic}}\bigl(\mathcal{M}(v_i^s)^{(t)},\ z_i\bigr),
\end{equation}
\begin{equation}
    \mathcal{M}(v_i^s)^{(t+1)} = \pi_{\text{editor-mic}}\bigl(\mathcal{M}(v_i^s)^{(t)},\ F_i^{(t)}\bigr),
\end{equation}
where $F_i^{(t)}$ is a list of actionable feedback, $\mathcal{M}(v_i^s)^{(t+1)}$ is the refined result. The tools used by the editor are the same as the integration step. This stage focuses on \textbf{\underline{mic}ro-level} repair (e.g., local conflict resolution, guideline adherence, and branch-level pruning). The second stage addresses \textbf{\underline{mac}ro-level} issues such as overall branch imbalance and structural incoherence, which are difficult to assess from textual relations alone. We further render the entire mind map as an image $I^{(t)} = \text{Vis}(\mathcal{M}^{(t)})$, and feed \textbf{both the image and the textual list} to a VLM-based critic; the editor then revises the whole map:
\begin{equation}
    F^{(t)} = \pi_{\text{critic-mac}}(\mathcal{M}^{(t)},\ I^{(t)}),
\end{equation}
\begin{equation}
   \mathcal{M}^{(t+1)} = \pi_{\text{editor-mac}}(\mathcal{M}^{(t)},\ F^{(t)},\ I^{(t)}).
\end{equation}
In macro-level refinement, the rendered image provides intuitive spatial awareness that text alone cannot convey. Structure-irrelevant pure visual design (e.g., color) is excluded in the refinement.

The refinements are repeated until reaching the maximum rounds or no issues are reported. Here the superscript $t$ indicates the refinement round, distinguished from the previous subscript. The two-stage refinement reflects a bottom-up philosophy that prevents mutual destruction caused by mixing local editing and global editing.

\section{Experiments}

\begin{table*}
\small
\setlength{\tabcolsep}{8pt}
\centering
\label{tab:results}
\begin{tabular}{@{}l l ccc cc cccc@{}}
\toprule
& \multirow{2}{*}{\textbf{Methods}} & \multicolumn{3}{c}{\textbf{GT-based}} & \multicolumn{2}{c}{\textbf{Structure Conformity}} & \multicolumn{4}{c}{\textbf{VLM-Judge}} \\
\cmidrule(lr){3-5} \cmidrule(lr){6-7} \cmidrule(lr){8-11}
& & \textbf{HN-F1} & \textbf{EC-F1}&\textbf{MEC-F1} & \textbf{Pass@1} & \textbf{SLF $\downarrow$} & \textbf{CA} & \textbf{SQ} & \textbf{PE} & \textbf{Avg.} \\
\midrule
\multicolumn{2}{@{}c}{\textbf{Human}} & - & - & - & 100.00 & 1.05 & 3.96 & 3.92 & 3.79 & 3.89 \\
\midrule
\multirow{5}{*}{\textbf{Baselines}}
& \textbf{Direct}      & \underline{33.43} & \underline{20.67} & 64.50 & 79.16 & 20.22 & {3.17} & {3.00} & {2.88} & {3.01} \\

& \textbf{Chunk-Merge} & 25.03 & 16.21 &\underline{68.10} & 83.33 & \underline{5.26} & 2.96 & 2.92 & 2.50 & 2.79 \\

& \textbf{Highlights-only}  & 23.63 & 14.50 & {66.25} & 95.83 & 19.12 & \underline{3.21} & \underline{3.13}& \underline{3.00}& \underline{3.11} \\

& \textbf{BookRAG}         & {25.12} & 10.12 & 47.85 & \textbf{\underline{100.00}}& {12.56} & 2.29 & 2.21 & 1.92 & 2.14 \\

& \textbf{PageIndex}        & 24.15 & 11.66 & 57.11 & \textbf{\underline{100.00}}& 14.19 & 2.75 & 2.50 & 2.08 & 2.44 \\
\midrule

\multirow{2}{*}{\textbf{AutoMindMap}}
& \textbf{+ GPT-4.1-mini}    & 37.80 & 22.96 & 74.01 & \textbf{100.00}& \textbf{1.82}  & 3.58 & 3.63 & \textbf{3.71} & 3.64 \\

& \textbf{+ GPT-5.4-mini}    &\textbf{39.88} & \textbf{25.12} & \textbf{74.34} & \textbf{100.00}& 1.93  & \textbf{3.83} & \textbf{3.71} & 3.63 & \textbf{3.72} \\
\bottomrule
\end{tabular}
\caption{Experimental results of different methods. DeepSeek-V4-Flash serves as the backbone model. \textit{Bold} and \textit{underlined} results indicate the global optimum and the baseline optimum (except for Human), respectively. Avg. is the average result of VLM-judge metrics. Human annotations serve as the ground truth; thus, GT-based metrics are inapplicable.}
\label{tab:comparison}
\end{table*}

\begin{figure*}[t]
\centering
\includegraphics[width=\linewidth]{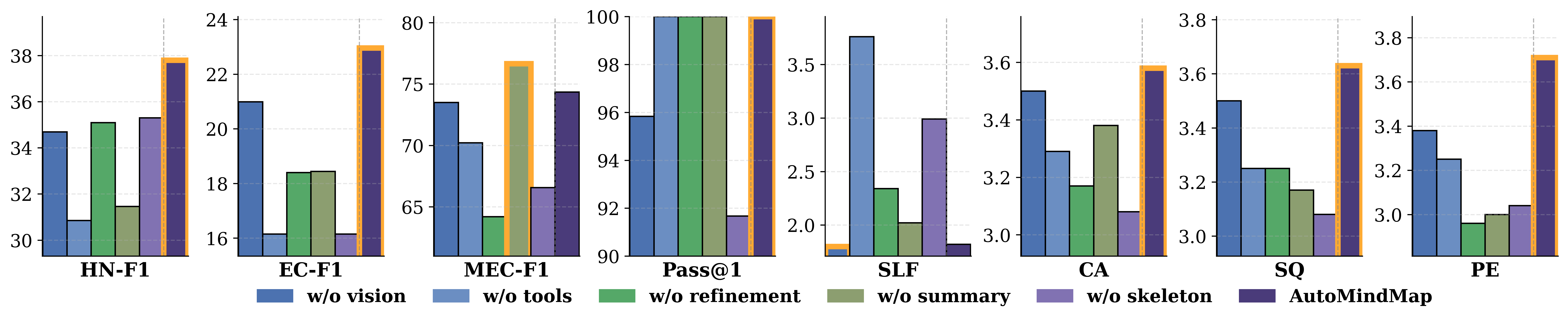}

\caption{Ablation study result. The corresponding detailed data is in Table~\ref{tab:abl}.}

\label{fig:abl}
\end{figure*}

\begin{figure}[t!]
\centering
\includegraphics[width=\linewidth]{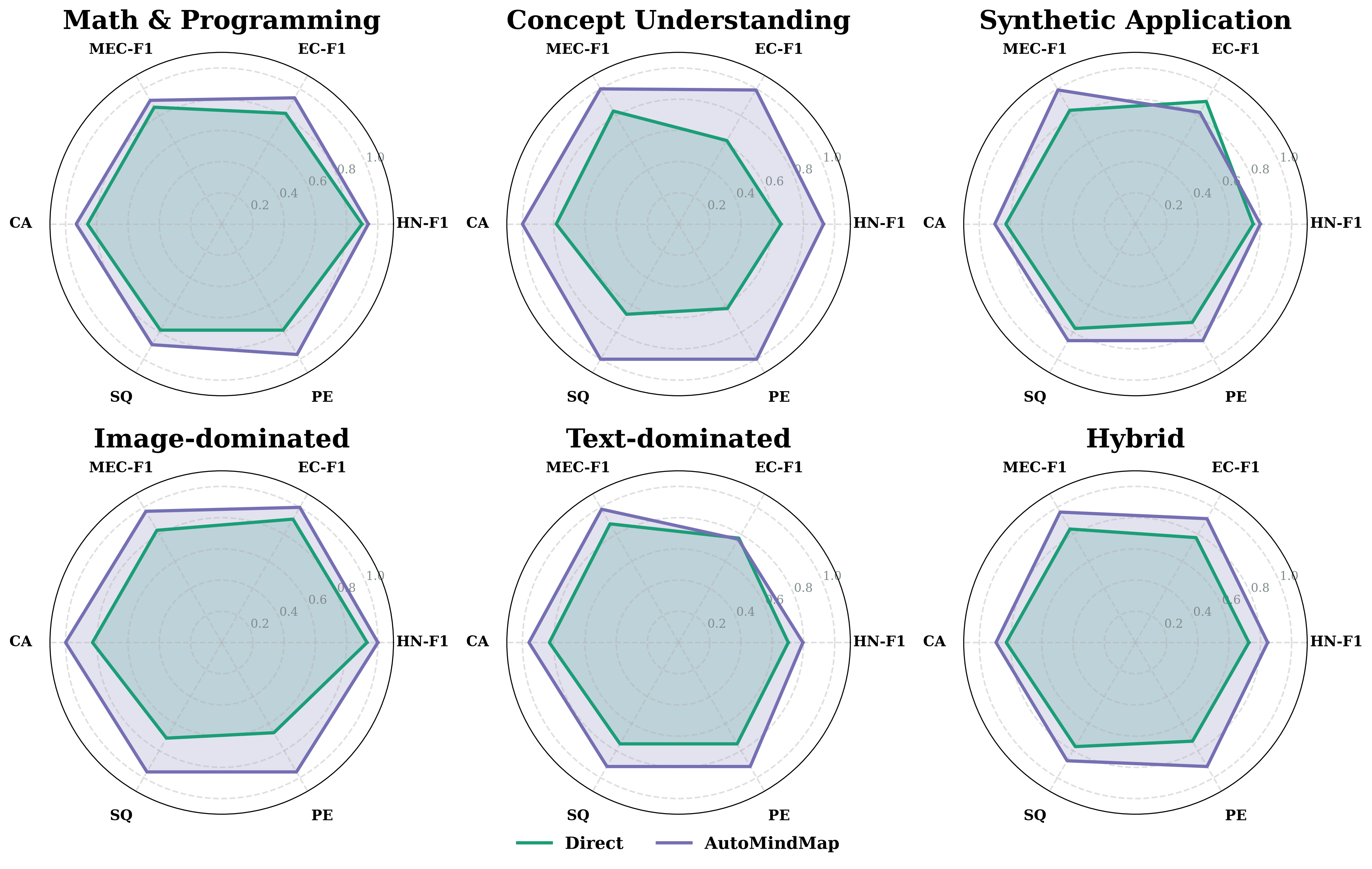}

\caption{Comparison between \textit{AutoMindMap} and \textit{Direct} on different courses categories. Each metric is normalized by its maximum value across all results, resulting in a [0,1] scale. AutoMindMap performs better on each category, especially in \textit{Concept Understanding}, showing the advantage in courses with more complex knowledge hierarchy. Appendix~\ref{app:moreexp} provides more detailed analysis.}

\label{fig:radar}
\end{figure}

We compare AutoMindMap against existing baselines and conduct ablation studies to validate the contribution of each module. Additional experiments are discussed in the \textbf{Appendix}, including the performance across different LLMs and course categories, the running cost and stability (Appendix~\ref{app:moreexp}), the evaluation reliability analysis (Appendix~\ref{app:Evaluation Reliability Analysis}), the statistic of structural indicator distribution (Appendix~\ref{app:morestruct}) and the case study of rendering effects (Appendix~\ref{app:casestudy}).

\subsection{Experiment Setup}
\subsubsection{Baselines.} Since no prior work directly addresses the Slides2MindMap task, we design several heuristic methods and select representative DHI methods as baselines: \textbf{(1)~Human}. Human-annotated ground-truth data serves as a competitive reference in structure conformity and VLM-Judge metrics. \textbf{(2)~Direct}. The LLM directly outputs all \textit{(concept, sub-concept)} tuples given the full course content and constraints as prompts. \textbf{(3)~Chunk-Merge}. Each deck is treated as a chunk (with heuristic segmentation for special long decks); local mind maps are generated per chunk, then merged and refined by the LLM and statistical rules, following the mainstream LLM-based KG construction (KGC) and \textit{Map-Reduce} paradigm~\citep{edge2024local, dean2008mapreduce}. \textbf{(4)~Highlights-only}. The LLM receives only salient elements (e.g., titles) from the parsed Markdown to infer hierarchy. \textbf{(5)~BookRAG}~\citep{wang2025bookrag}. A tree-hierarchy indexing framework based on MinerU's parsing output. We only use its tree indexing algorithm, excluding the graph construction and retrieval parts. \textbf{(6)~PageIndex}~\citep{zhang2025pageindex}. A TOC-style vector-free indexing method that organizes hierarchy via formatting cues and content understanding. Notably, mainstream KGC methods~\citep{zhu2024llms} are excluded because the generated KG structure is far from a mind map, and even beyond the basic constraints (i.e., $\text{pass@1}=0\%$). Training-based mind mapping methods~\citep{hu2021efficient,zhang2024coreference} are also excluded, as our data scale and scenarios are beyond their scope (i.e., sentence-level short text summarization).

\subsubsection{Implementation.} Since AutoMindMap is training-free and to ensure evaluation effectiveness, all 24 courses in S2M-Bench are treated as the test set. All raw files are parsed into Markdown via MinerU for all non-Human methods. DeepSeek-V4-Flash serves as the backbone LLM for baselines and the vision-free modules of AutoMindMap; GPT-4.1-mini and GPT-5.4-mini are evaluated separately as the VLM for visual refinement in AutoMindMap. More detailed experimental implementation is in Appendix~\ref{app:moreexpset}. We conducted further experiments with more models and parameters, which are presented in Appendix~\ref{app:moreexp}.

\subsection{Experiment Results}

\subsubsection{Main Experiments.} As shown in Table~\ref{tab:comparison}, \textbf{AutoMindMap} (with either VLM) avoids illegal generation and consistently outperforms all baselines (except for Human) across every metric, while the stronger VLM (GPT-5.4-mini) yields further improvements. \textbf{Human} annotations achieve the highest scores, confirming the annotation quality, and AutoMindMap still has a gap with human experts. Among baselines, \textbf{Direct} and \textbf{Highlights-only} perform relatively better, as they directly generate the entire map, preserving global structural consistency at the cost of missing inconspicuous concepts and generating invalid structure. \textbf{Chunk-Merge} maintains competitive local knowledge faithfulness but suffers from poor global coherence due to the post-merging mechanism, reflected in lower VLM-Judge scores. \textbf{BookRAG} and \textbf{PageIndex} reliably produce valid tree structures (Pass@1 = 100\%) but perform worst on both GT-based and VLM-Judge metrics, confirming that retrieval-oriented DHI methods are ill-suited for the Slides2MindMap task. All the automated baselines except for Chunk-Merge have obviously high SLF, which indicates the risk of increased cognitive load of the map. We conduct further analysis on fine-grained GT-based metrics and visualization cases in Appendix~\ref{app:gtdetail} and Appendix~\ref{app:casestudy}.

\subsubsection{Ablation Study.}
Figure~\ref{fig:abl} presents the ablation results, where we compare the full AutoMindMap against five variants: \textbf{(1)~w/o vision}, which supplies only textual descriptions of the map during refinement; \textbf{(2)~w/o tools}, which directly generates JSON tuples for map updates; \textbf{(3)~w/o refinement}, which ablates the decoupled refinement mechanism; \textbf{(4)~w/o summary}, which directly transforms the long corpus into a mind map without context-aware summarization for discourse-based grouping; \textbf{(5)~w/o skeleton}, which builds the map directly from the root without any prior scaffolding and skill. We only use GPT-4.1-mini for visual refinement. Notably, AutoMindMap is not strictly optimal on every metric (e.g., slightly underperforms on MEC-F1 and SLF), indicating that some components trade off performance on specific dimensions for overall quality improvement. The trade-off is discussed in Appendix~\ref{app:gtdetail} and Appendix~\ref{app:metcor}. We combine the visual generation effect and provide a more in-depth study on the function of each ablated module in Appendix~\ref{app:casestudy}.

\subsubsection{Further Analysis.}
AutoMindMap exhibits robust performance across different LLMs and diverse course categories (see Figure~\ref{fig:radar}), low variance under stochastic multi-runs, and eligible runtime cost and visual understanding effects. These detailed experiments are in Appendix~\ref{app:moreexp} and Appendix~\ref{app:casestudy}. For the evaluation framework of S2M-Bench, in Appendix~\ref{app:Evaluation Reliability Analysis}, we confirm that VLM-Judge exhibits high agreement with human and cross-model judges, the robustness of the evaluation system, and that GT-based metrics are influenced by task openness. Appendix~\ref{app:morestruct} further shows that the structural statistical feature distribution of the maps generated by AutoMindMap is closer to that of humans.

\section{Conclusion}

In this work, we formulate a new task of Slides2MindMap, which aims to generate cognitively efficient concept mind maps from large-scale course lecture slides. We further propose a benchmark, S2M-Bench, to comprehensively evaluate the Slides2MindMap task, and an LLM-based agentic framework, AutoMindMap, which outperforms the existing baselines in S2M-Bench. Based on the Structure Building Framework Theory, AutoMindMap involves a pipeline consisting of skeleton laying, iterative knowledge integration, and dual-stage refinement, balancing local knowledge faithfulness and global logic coherence, and adapting to the large-scale fragmented slide content. However, our study has some limitations (see Appendix~\ref{app:limit}). These issues can be further explored. AutoMindMap has been deployed on relevant intelligent education platforms and will be fully open-source. We hope that this work can advance the field of Educational AI, especially knowledge structure visualization.

\bibliography{aaai2027}

@article{raiyn2016role,
  title={The Role of Visual Learning in Improving Students' High-Order Thinking Skills.},
  author={Raiyn, Jamal},
  journal={Journal of education and practice},
  volume={7},
  number={24},
  pages={115--121},
  year={2016},
  publisher={ERIC}
}

@book{buzan2006mind,
  title={The mind map book},
  author={Buzan, Tony and Buzan, Barry},
  year={2006},
  publisher={Pearson Education}
}

@article{ausubel1960use,
  title={The use of advance organizers in the learning and retention of meaningful verbal material.},
  author={Ausubel, David P},
  journal={Journal of educational psychology},
  volume={51},
  number={5},
  pages={267},
  year={1960},
  publisher={American Psychological Association}
}

@incollection{sweller2011cognitive,
  title={Cognitive load theory},
  author={Sweller, John},
  booktitle={Psychology of learning and motivation},
  volume={55},
  pages={37--76},
  year={2011},
  publisher={Elsevier}
}

@article{wang2024survey,
  title={A survey of models for cognitive diagnosis: New developments and future directions},
  author={Wang, Fei and Gao, Weibo and Liu, Qi and Li, Jiatong and Zhao, Guanhao and Zhang, Zheng and Huang, Zhenya and Zhu, Mengxiao and Wang, Shijin and Tong, Wei and others},
  journal={arXiv preprint arXiv:2407.05458},
  year={2024}
}

@inproceedings{low2025boommapper,
  title={BoomMapper: An Innovative Web-Based System for Transforming PowerPoint Presentations into Mind Maps},
  author={Low, Coey and Kasinathan, Vinothini and Mustapha, Aida},
  booktitle={International Workshop on Learning Technology for Education Challenges},
  pages={128--139},
  year={2025},
  organization={Springer}
}

@inproceedings{zhang2024coreference,
  title={Coreference graph guidance for mind-map generation},
  author={Zhang, Zhuowei and Hu, Mengting and Bai, Yinhao and Zhang, Zhen},
  booktitle={Proceedings of the AAAI Conference on Artificial Intelligence},
  volume={38},
  number={17},
  pages={19623--19631},
  year={2024}
}

@inproceedings{kasinathan2024presentation,
  title={A Presentation Mining Framework: From Text Mining to to Mind Mapping},
  author={Kasinathan, Vinothini and Mustapha, Aida},
  booktitle={International Conference on Soft Computing and Data Mining},
  pages={233--243},
  year={2024},
  organization={Springer}
}

@inproceedings{tanaka2023slidevqa,
  title={Slidevqa: A dataset for document visual question answering on multiple images},
  author={Tanaka, Ryota and Nishida, Kyosuke and Nishida, Kosuke and Hasegawa, Taku and Saito, Itsumi and Saito, Kuniko},
  booktitle={Proceedings of the AAAI Conference on Artificial Intelligence},
  volume={37},
  number={11},
  pages={13636--13645},
  year={2023}
}

@inproceedings{hu2021efficient,
  title={Efficient mind-map generation via sequence-to-graph and reinforced graph refinement},
  author={Hu, Mengting and Guo, Honglei and Zhao, Shiwan and Gao, Hang and Su, Zhong},
  booktitle={Proceedings of the 2021 Conference on Empirical Methods in Natural Language Processing},
  pages={8130--8141},
  year={2021}
}

@article{zhu2024llms,
  title={Llms for knowledge graph construction and reasoning: Recent capabilities and future opportunities},
  author={Zhu, Yuqi and Wang, Xiaohan and Chen, Jing and Qiao, Shuofei and Ou, Yixin and Yao, Yunzhi and Deng, Shumin and Chen, Huajun and Zhang, Ningyu},
  journal={World Wide Web},
  volume={27},
  number={5},
  pages={58},
  year={2024},
  publisher={Springer}
}

@article{zhang2025systematic,
  title={A systematic survey of text summarization: From statistical methods to large language models},
  author={Zhang, Haopeng and Yu, Philip S and Zhang, Jiawei},
  journal={ACM Computing Surveys},
  volume={57},
  number={11},
  pages={1--41},
  year={2025},
  publisher={ACM New York, NY}
}

@article{zhang2025survey,
  title={A survey of graph retrieval-augmented generation for customized large language models},
  author={Zhang, Qinggang and Chen, Shengyuan and Bei, Yuanchen and Yuan, Zheng and Zhou, Huachi and Hong, Zijin and Chen, Hao and Xiao, Yilin and Zhou, Chuang and Dong, Junnan and others},
  journal={arXiv preprint arXiv:2501.13958},
  year={2025}
}

@article{davies2011concept,
  title={Concept mapping, mind mapping and argument mapping: what are the differences and do they matter?},
  author={Davies, Martin},
  journal={Higher education},
  volume={62},
  number={3},
  pages={279--301},
  year={2011},
  publisher={Springer}
}

@article{hua2019exploring,
  title={Exploring the psychometric properties of the mind-map scoring rubric},
  author={Hua, Cheng and Wind, Stefanie A},
  journal={Behaviormetrika},
  volume={46},
  number={1},
  pages={73--99},
  year={2019},
  publisher={Springer}
}

@article{zhang2025pageindex,
  author = {Mingtian Zhang and Yu Tang and PageIndex Team},
  title = {PageIndex: Next-Generation Vectorless, Reasoning-based RAG},
  journal = {PageIndex Blog},
  year = {2025},
  month = {September},
  note = {https://pageindex.ai/blog/pageindex-intro},
}

@article{wang2025bookrag,
  title={BookRAG: A Hierarchical Structure-aware Index-based Approach for Retrieval-Augmented Generation on Complex Documents},
  author={Wang, Shu and Zhou, Yingli and Fang, Yixiang},
  journal={arXiv preprint arXiv:2512.03413},
  year={2025}
}

@book{gernsbacher2013language,
  title={Language comprehension as structure building},
  author={Gernsbacher, Morton Ann},
  year={2013},
  publisher={Psychology Press}
}

@article{armstrong2010bloom,
  title={Bloom’s taxonomy},
  author={Armstrong, Patricia},
  journal={Vanderbilt University Center for Teaching},
  volume={12},
  number={05},
  pages={2023},
  year={2010},
  publisher={Vanderbilt University Nashville, TN, USA}
}

@article{miller1956magical,
  title={The magical number seven, plus or minus two: Some limits on our capacity for processing information.},
  author={Miller, George A},
  journal={Psychological review},
  volume={63},
  number={2},
  pages={81},
  year={1956},
  publisher={American Psychological Association}
}

@article{kedaj2014effective,
  title={Effective mind maps in e-learning},
  author={Kedaj, Petr and Pavl{\'\i}{\v{c}}ek, Josef and Hanzl{\'\i}k, Petr and others},
  journal={Acta Informatica Pragensia},
  volume={3},
  number={3},
  pages={239--250},
  year={2014}
}

@inproceedings{nemoto2014likert,
  title={Likert-scale questionnaires},
  author={Nemoto, Tomoko and Beglar, David},
  booktitle={JALT 2013 conference proceedings},
  volume={108},
  number={1},
  pages={1--6},
  year={2014}
}

@article{wang2024mineru,
  title={Mineru: An open-source solution for precise document content extraction},
  author={Wang, Bin and Xu, Chao and Zhao, Xiaomeng and Ouyang, Linke and Wu, Fan and Zhao, Zhiyuan and Xu, Rui and Liu, Kaiwen and Qu, Yuan and Shang, Fukai and others},
  journal={arXiv preprint arXiv:2409.18839},
  year={2024}
}

@incollection{feng2020genre,
  title={Genre, pedagogy, and PowerPoint design: A multimodal move analysis of linguistics lecture slides},
  author={Feng, Dezheng William},
  booktitle={Approaches to specialized genres},
  pages={177--197},
  year={2020},
  publisher={Routledge}
}

@article{o1998updating,
  title={Updating a situation model: a memory-based text processing view.},
  author={O'Brien, Edward J and Rizzella, Michelle L and Albrecht, Jason E and Halleran, Jennifer G},
  journal={Journal of Experimental Psychology: Learning, Memory, and Cognition},
  volume={24},
  number={5},
  pages={1200},
  year={1998},
  publisher={American Psychological Association}
}

@article{dean2008mapreduce,
  title={MapReduce: simplified data processing on large clusters},
  author={Dean, Jeffrey and Ghemawat, Sanjay},
  journal={Communications of the ACM},
  volume={51},
  number={1},
  pages={107--113},
  year={2008},
  publisher={ACM New York, NY, USA}
}

@article{edge2024local,
  title={From local to global: A graph rag approach to query-focused summarization},
  author={Edge, Darren and Trinh, Ha and Cheng, Newman and Bradley, Joshua and Chao, Alex and Mody, Apurva and Truitt, Steven and Metropolitansky, Dasha and Ness, Robert Osazuwa and Larson, Jonathan},
  journal={arXiv preprint arXiv:2404.16130},
  year={2024}
}

@inproceedings{fabbri2018tutorialbank,
  title={Tutorialbank: A manually-collected corpus for prerequisite chains, survey extraction and resource recommendation},
  author={Fabbri, Alexander Richard and Li, Irene and Trairatvorakul, Prawat and He, Yijiao and Ting, Weitai and Tung, Robert and Westerfield, Caitlin and Radev, Dragomir},
  booktitle={Proceedings of the 56th Annual Meeting of the Association for Computational Linguistics (Volume 1: Long Papers)},
  pages={611--620},
  year={2018}
}

@inproceedings{han2025beyond,
  title={Beyond Linear Digital Reading: An LLM-Powered Concept Mapping Approach for Reducing Cognitive Load},
  author={Han, Junzhi and Choi, Jinho D},
  booktitle={Proceedings of the 20th Workshop on Innovative Use of NLP for Building Educational Applications (BEA 2025)},
  pages={805--817},
  year={2025}
}

@inproceedings{niu2025tree,
  title={Tree-KG: An expandable knowledge graph construction framework for knowledge-intensive domains},
  author={Niu, Songjie and Yang, Kaisen and Zhao, Rui and Liu, Yichao and Li, Zonglin and Wang, Hongning and Chen, Wenguang},
  booktitle={Proceedings of the 63rd Annual Meeting of the Association for Computational Linguistics (Volume 1: Long Papers)},
  pages={18516--18529},
  year={2025}
}

@inproceedings{li2019should,
  title={What should i learn first: Introducing lecturebank for nlp education and prerequisite chain learning},
  author={Li, Irene and Fabbri, Alexander R and Tung, Robert R and Radev, Dragomir R},
  booktitle={Proceedings of the AAAI Conference on Artificial Intelligence},
  volume={33},
  number={01},
  pages={6674--6681},
  year={2019}
}

@article{rawat2023logical,
  title={Logical concept mapping and social media analytics relating to cyber criminal activities for ontology creation},
  author={Rawat, Romil},
  journal={International Journal of Information Technology},
  volume={15},
  number={2},
  pages={893--903},
  year={2023},
  publisher={Springer}
}

@article{ding2024automated,
  title={Automated construction of theme-specific knowledge graphs},
  author={Ding, Linyi and Zhou, Sizhe and Xiao, Jinfeng and Han, Jiawei},
  journal={arXiv preprint arXiv:2404.19146},
  year={2024}
}

@article{xiao2025eduvqa,
  title={EduVQA: A multimodal Visual Question Answering framework for smart education},
  author={Xiao, Jiongen and Zhang, Zifeng},
  journal={Alexandria Engineering Journal},
  volume={122},
  pages={615--624},
  year={2025},
  publisher={Elsevier}
}

@inproceedings{zhang2025awaking,
  title={Awaking the slides: a tuning-free and knowledge-regulated AI tutoring system via language model coordination},
  author={Zhang-Li, Daniel and Zhang, Zheyuan and Yu, Jifan and Lim, Joy Jia Yin and Tu, Shangqing and Gong, Linlu and Wang, Haohua and Liu, Zhiyuan and Liu, Huiqin and Hou, Lei and others},
  booktitle={Proceedings of the 31st ACM SIGKDD Conference on Knowledge Discovery and Data Mining V. 1},
  pages={2872--2883},
  year={2025}
}

@article{dhindsa2011constructivist,
  title={Constructivist-visual mind map teaching approach and the quality of students’ cognitive structures},
  author={Dhindsa, Harkirat S and Roger Anderson, O},
  journal={Journal of Science Education and Technology},
  volume={20},
  number={2},
  pages={186--200},
  year={2011},
  publisher={Springer}
}

@inproceedings{Yu2020MOOCCubeAL,
  title={MOOCCube: A Large-scale Data Repository for NLP Applications in MOOCs},
  author={Jifan Yu and Gan Luo and Tong Xiao and Qingyang Zhong and Yuquan Wang and Wenzheng Feng and Junyi Luo and Chenyu Wang and Lei Hou and Juan-Zi Li and Zhiyuan Liu and Jie Tang},
  booktitle={Annual Meeting of the Association for Computational Linguistics},
  year={2020},
}

@inproceedings{Yu_2019,
   title={Course Concept Expansion in MOOCs with External Knowledge and Interactive Game},
   DOI={10.18653/v1/p19-1421},
   booktitle={Proceedings of the 57th Annual Meeting of the Association for Computational Linguistics},
   publisher={Association for Computational Linguistics},
   author={Yu, Jifan and Wang, Chenyu and Luo, Gan and Hou, Lei and Li, Juanzi and Liu, Zhiyuan and Tang, Jie},
   year={2019},
   pages={4292–4302} }

@misc{aburasheed2025llmassistedknowledgegraphcompletion,
      title={LLM-Assisted Knowledge Graph Completion for Curriculum and Domain Modelling in Personalized Higher Education Recommendations}, 
      author={Hasan Abu-Rasheed and Constance Jumbo and Rashed Al Amin and Christian Weber and Veit Wiese and Roman Obermaisser and Madjid Fathi},
      year={2025},
      eprint={2501.12300},
      archivePrefix={arXiv},
      primaryClass={cs.HC},
}

@article{https://doi.org/10.1002/widm.1332,
author = {Ferreira-Mello, Rafael and André, Máverick and Pinheiro, Anderson and Costa, Evandro and Romero, Cristobal},
title = {Text mining in education},
journal = {WIREs Data Mining and Knowledge Discovery},
volume = {9},
number = {6},
pages = {e1332},
doi = {https://doi.org/10.1002/widm.1332},
year = {2019}
}

@INPROCEEDINGS{11444950,
  author={Zhang, Enming and Li, Yuzhe and Liu, Yuliang and Zhu, Yingying and Bai, Xiang},
  booktitle={2025 IEEE/CVF International Conference on Computer Vision (ICCV)}, 
  title={Towards Comprehensive Lecture Slides Understanding: Large-Scale Dataset and Effective Method}, 
  year={2025},
  volume={},
  number={},
  pages={4455-4464},
  doi={10.1109/ICCV51701.2025.00424}}

@InProceedings{Chen2024a,
  author    = {Zhe Chen and Heyang Liu and Wenyi Yu and Guangzhi Sun and Hongcheng Liu and Ji Wu and Chao Zhang and Yu Wang and Yanfeng Wang},
  booktitle = {Proceedings of the 62nd Annual Meeting of the Association for Computational Linguistics (Volume 1: Long Papers), {ACL} 2024, Bangkok, Thailand, August 11-16, 2024},
  title     = {M{\({^3}\)}{AV}: A Multimodal, Multigenre, and Multipurpose Audio-Visual Academic Lecture Dataset},
  year      = {2024},
  editor    = {Lun{-}Wei Ku and Andre Martins and Vivek Srikumar},
  pages     = {9041--9060},
  publisher = {Association for Computational Linguistics},
  bibsource = {dblp computer science bibliography, https://dblp.org},
  doi       = {10.18653/V1/2024.ACL-LONG.489},
}

@INPROCEEDINGS{10376585,
  author={Lee, Dong Won and Ahuja, Chaitanya and Liang, Paul Pu and Natu, Sanika and Morency, Louis-Philippe},
  booktitle={2023 IEEE/CVF International Conference on Computer Vision (ICCV)}, 
  title={Lecture Presentations Multimodal Dataset: Towards Understanding Multimodality in Educational Videos}, 
  year={2023},
  volume={},
  number={},
  pages={20030-20041},
  doi={10.1109/ICCV51070.2023.01838}}

@article{Viera2005UnderstandingIA,
  title={Understanding interobserver agreement: the kappa statistic.},
  author={Anthony J. Viera and Joanne Mills Garrett},
  journal={Family medicine},
  year={2005},
  volume={37 5},
  pages={
          360-3
        },
}

@misc{chen2024mllmasajudgeassessingmultimodalllmasajudge,
      title={MLLM-as-a-Judge: Assessing Multimodal LLM-as-a-Judge with Vision-Language Benchmark}, 
      author={Dongping Chen and Ruoxi Chen and Shilin Zhang and Yinuo Liu and Yaochen Wang and Huichi Zhou and Qihui Zhang and Yao Wan and Pan Zhou and Lichao Sun},
      year={2024},
      eprint={2402.04788},
      archivePrefix={arXiv},
      primaryClass={cs.CL},
      url={https://arxiv.org/abs/2402.04788}, 
}

@article{Sen1968EstimatesOT,
  title={Estimates of the Regression Coefficient Based on Kendall's Tau},
  author={Pranab Kumar Sen},
  journal={Journal of the American Statistical Association},
  year={1968},
  volume={63},
  pages={1379-1389},
  url={https://api.semanticscholar.org/CorpusID:122323572}
}

@INPROCEEDINGS{175815,
  author={Johnson, B. and Shneiderman, B.},
  booktitle={Proceeding Visualization '91}, 
  title={Tree-maps: a space-filling approach to the visualization of hierarchical information structures}, 
  year={1991},
  volume={},
  number={},
  pages={284-291},
  doi={10.1109/VISUAL.1991.175815}}

@article{Sweller2019CognitiveAA,
  title={Cognitive Architecture and Instructional Design: 20 Years Later},
  author={John Sweller and Jeroen J. G. van Merri{\"e}nboer and Fred Paas},
  journal={Educational Psychology Review},
  year={2019},
  volume={31},
  pages={261 - 292},
}

@inproceedings{Spiro1988CognitiveFT,
  title={Cognitive flexibility theory : advanced knowledge acquisition in ill-structured domains},
  author={Rand J. Spiro},
  year={1988},
  url={https://api.semanticscholar.org/CorpusID:54151398}
}

@article{hassett2009theories,
  title={Theories and practices of multimodal education: The instructional dynamics of picture books and primary classrooms},
  author={Hassett, Dawnene D and Curwood, Jen Scott},
  journal={The Reading Teacher},
  volume={63},
  number={4},
  pages={270--282},
  year={2009},
  publisher={Wiley Online Library}
}

@article{xu2026agent,
  title={Agent skills for large language models: Architecture, acquisition, security, and the path forward},
  author={Xu, Renjun and Yan, Yang},
  journal={arXiv preprint arXiv:2602.12430},
  year={2026}
}

@article{liu2024lost,
  title={Lost in the middle: How language models use long contexts},
  author={Liu, Nelson F and Lin, Kevin and Hewitt, John and Paranjape, Ashwin and Bevilacqua, Michele and Petroni, Fabio and Liang, Percy},
  journal={Transactions of the association for computational linguistics},
  volume={12},
  pages={157--173},
  year={2024}
}

@article{gonnermann2026llm,
  title={LLM-Based Educational Simulation: Evaluating Temporal Student Persona Stability Across ADHD Profiles},
  author={Gonnermann-M{\"u}ller, Jana and Haase, Jennifer and Leins, Nicolas and Kosch, Thomas and Pokutta, Sebastian},
  journal={arXiv preprint arXiv:2605.06307},
  year={2026}
}

@article{dong2024clr,
  title={Clr-bench: Evaluating large language models in college-level reasoning},
  author={Dong, Junnan and Hong, Zijin and Bei, Yuanchen and Huang, Feiran and Wang, Xinrun and Huang, Xiao},
  journal={arXiv preprint arXiv:2410.17558},
  year={2024}
}

\appendix

\setcounter{secnumdepth}{2}

\clearpage
\tableofcontents

\section{Related Work}
\label{ap:relatedwork}

\begin{table*}
\small
\setlength{\tabcolsep}{5pt}
\centering

\label{tab:Kappa}
\begin{tabular}{lccc}
\toprule
\textbf{Datasets} & \textbf{Decks} & \textbf{Pages} & \textbf{Tasks}   \\
\midrule
\textbf{LectureBank}~\citep{li2019should} & 1,352 & 51,939 & Prerequisite Graph Construction  \\
\textbf{SlideVQA}~\citep{tanaka2023slidevqa} & 2,619 & 52,380 & Visual Question Answering  \\
\textbf{LPM}~\citep{10376585} &334 & 9,031 & Image Retrieval \\
\textbf{$\text{M}^3$AV}~\citep{Chen2024a} &1,113 & 24,936 & Multimodal Video Understanding \\
\textbf{LecSlides-370K}~\citep{11444950}  & 25,542 & 370,078 & Visual Question Answering \& Summary  \\
\midrule
\textbf{S2M-Bench} & 267 & 12,774 & Mind Map Construction  \\
\bottomrule
\end{tabular}
\caption{Comparisons among different lecture slides datasets.}
\label{tab:dscompare}
\end{table*}

\noindent \textbf{Educational Knowledge Extraction and Mind Map.} Extracting structured knowledge from pedagogical materials facilitates numerous intelligent education applications~\citep{Yu2020MOOCCubeAL}, including cognitive diagnosis~\citep{wang2024survey}, recommendation~\citep{fabbri2018tutorialbank}, question answering~\citep{xiao2025eduvqa}, lecture planning~\citep{zhang2025awaking}, and visual learning~\citep{dhindsa2011constructivist}. Prior studies have explored prerequisite relations~\citep{li2019should}, concept maps~\citep{rawat2023logical}, mind maps~\citep{zhang2024coreference}, educational knowledge graphs~\citep{niu2025tree}, and concept expansion~\citep{Yu_2019}, etc., and the development of recent large language models (LLMs) further facilitates these tasks~\citep{zhang2025awaking,han2025beyond,aburasheed2025llmassistedknowledgegraphcompletion}. For mind map generation, previous studies are few and focus on the identification of concept dominance relationships from simplified or well-structured text; the evaluations are limited to accuracy-based or downstream task-oriented metrics~\citep{zhang2024coreference, hu2021efficient, https://doi.org/10.1002/widm.1332, ding2024automated}. These studies present limited generalization and scalability, especially in more open-ended generation and evaluation. Although the Human-Computer Interaction field has more related studies, they only use LLMs as end-to-end mind map generators and are limited to statistical evaluation on user experiments~\citep{gonnermann2026llm,han2025beyond}. In conclusion, there is still a lack of focus on automated mind map generation in large-scale slide scenarios, and more comprehensive evaluation methods considering knowledge faithfulness, structure compliance, task openness, and cognitive efficiency. These gaps remain critical in current educational scenarios where slides dominate.

\noindent \textbf{Lecture Slides Datasets.} Several datasets have been proposed for lecture slide-based tasks, which mainly involve pedagogical applications, such as structured knowledge extraction~\citep{li2019should}, multimodal content understanding~\citep{10376585}, and question answering~\citep{tanaka2023slidevqa}. Among representative studies, LectureBank~\citep{li2019should} aims at prerequisite graph learning; SlideVQA~\citep{tanaka2023slidevqa} and LecSlides-370K~\citep{11444950} target visual question answering and summarization over thousands of slides; LPM~\citep{10376585} and $\text{M}^3$AV~\citep{Chen2024a} introduce other multimodal data sources like videos, audio, textbooks, etc., to support multimodal retrieval and understanding. Table~\ref{tab:dscompare} summarizes the comparisons among these datasets. These datasets treat slides as a corpus or visual sources for retrieval, instead of framing knowledge structuring as a standalone generation task. S2M-Bench fills this gap by providing expert-annotated mind maps as a structured knowledge abstraction over each entire course, which is the first benchmark that systematically evaluates \textbf{automated mind map generation}, a human visual-learning-targeted hierarchical knowledge reconstruction task from authentic, large-scale slide collections. 

\noindent \textbf{Document Hierarchy Indexing.} Recent methods for table-of-contents (TOCs) style document hierarchy indexing (DHI), such as Tree-KG~\citep{niu2025tree}, PageIndex~\citep{zhang2025pageindex}, and BookRAG~\citep{wang2025bookrag}, preserve the native hierarchy of well-structured documents by exploiting explicit formatting cues. While these methods can produce structures visually reminiscent of mind maps, their reliance on formatting signals renders them brittle when applied to slides, due to the systematic misalignment between surface document organization and the intrinsic concept hierarchical structure. Thus, retrieval-oriented DHI methods are ill-suited for pedagogical mind map generation. A fundamentally different generation paradigm is required.

\section{Limitations and Future Work}
\label{app:limit}
\noindent \textbf{Benchmark Comprehensiveness}. Due to the high cost of manual labeling and the limited access to multi-disciplinary educational resources, S2M-Bench only includes a modest-scale dataset with a focus on computer science and mathematics. Future work can further expand the dataset to more subjects. However, we still categorize the dataset based on multiple dimensions, which alleviates the data insufficiency problem. Besides, given the high cost and potential ethical concerns, we don't conduct experiments in real educational scenarios to collect human feedback. Human feedback is the most intuitive evaluation and can be considered in the future.

\noindent \textbf{Generation Efficiency}. AutoMindMap does not show a particularly high advantage in cost efficiency (e.g., token consumption and running time), while also focusing only on individual course generation. Future work can further propose more token-saving and faster generation methods, and introduce self-evolution mechanisms to allow agents to reuse the generation experience in different courses. 

\noindent \textbf{The Openness of the Task Definition}. In the educational field, the visual design effect of mind maps is also a key evaluation indicator. However, Slides2MindMap is only defined within concept structure organization, and excludes visual rendering modules (i.e., visual rendering is implemented with external tools instead of an internal module). Future work can consider building mind maps with a more open form that involves images, annotations, links, and typography design, etc. The task can be further modeled as a text-to-image generation task.

\noindent \textbf{Dependence on Document Parsing Tools}. The generation quality of AutoMindMap depends on the parsing results of MinerU, which is an inevitable drawback of this work. However, parsing the layout of PDF documents is a separate task with its own challenges. We still hope that document-native end-to-end frameworks can be proposed.

\begin{figure*}[t]
    \centering
    \begin{subfigure}{0.63\linewidth}
        \centering
        \includegraphics[width=\linewidth]{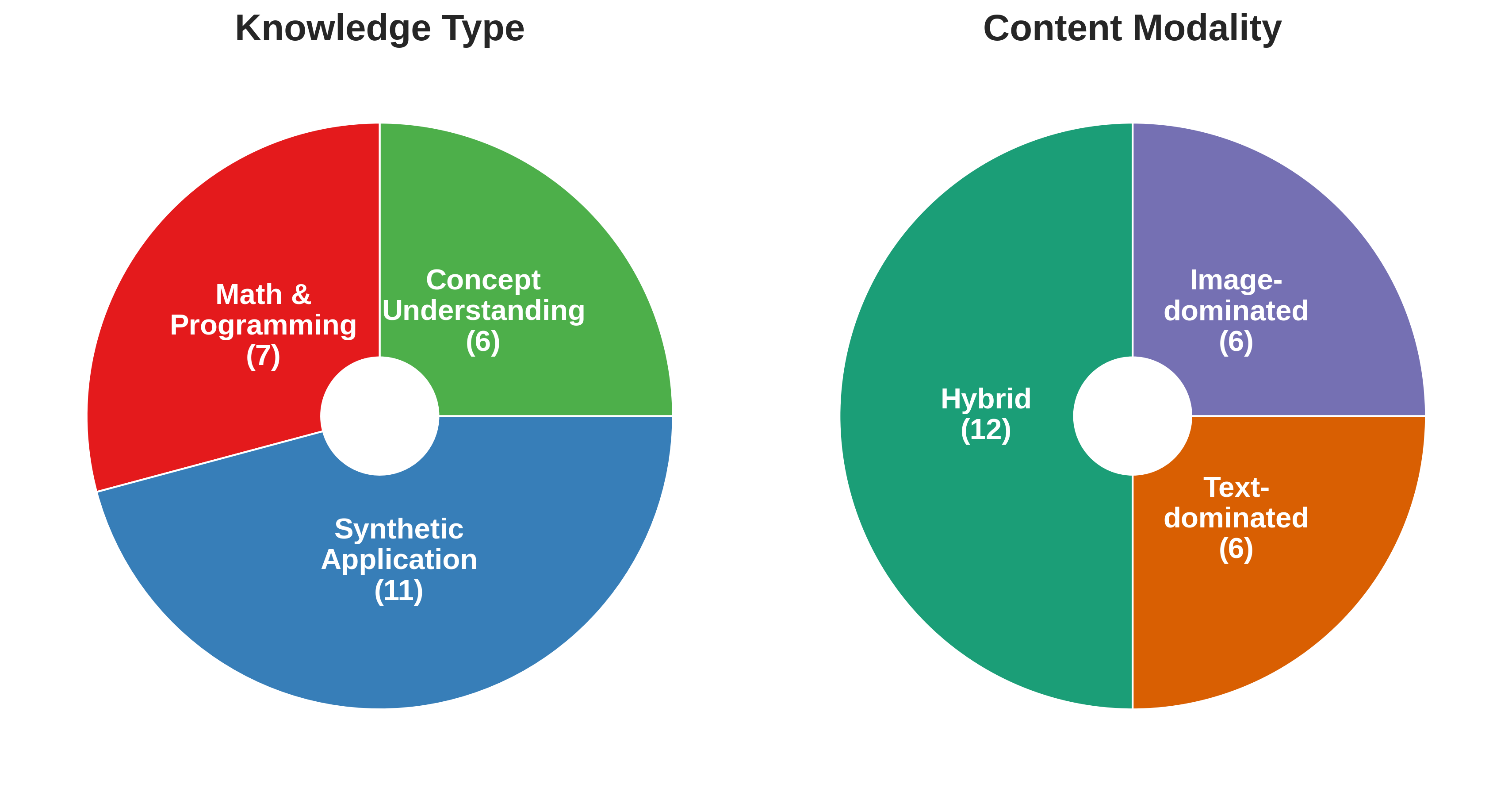}
        \caption{Course categories.}
        \label{fig:ring_pies}
    \end{subfigure}
    \hfill
    \begin{subfigure}{0.35\linewidth}
        \centering
        \includegraphics[width=\linewidth]{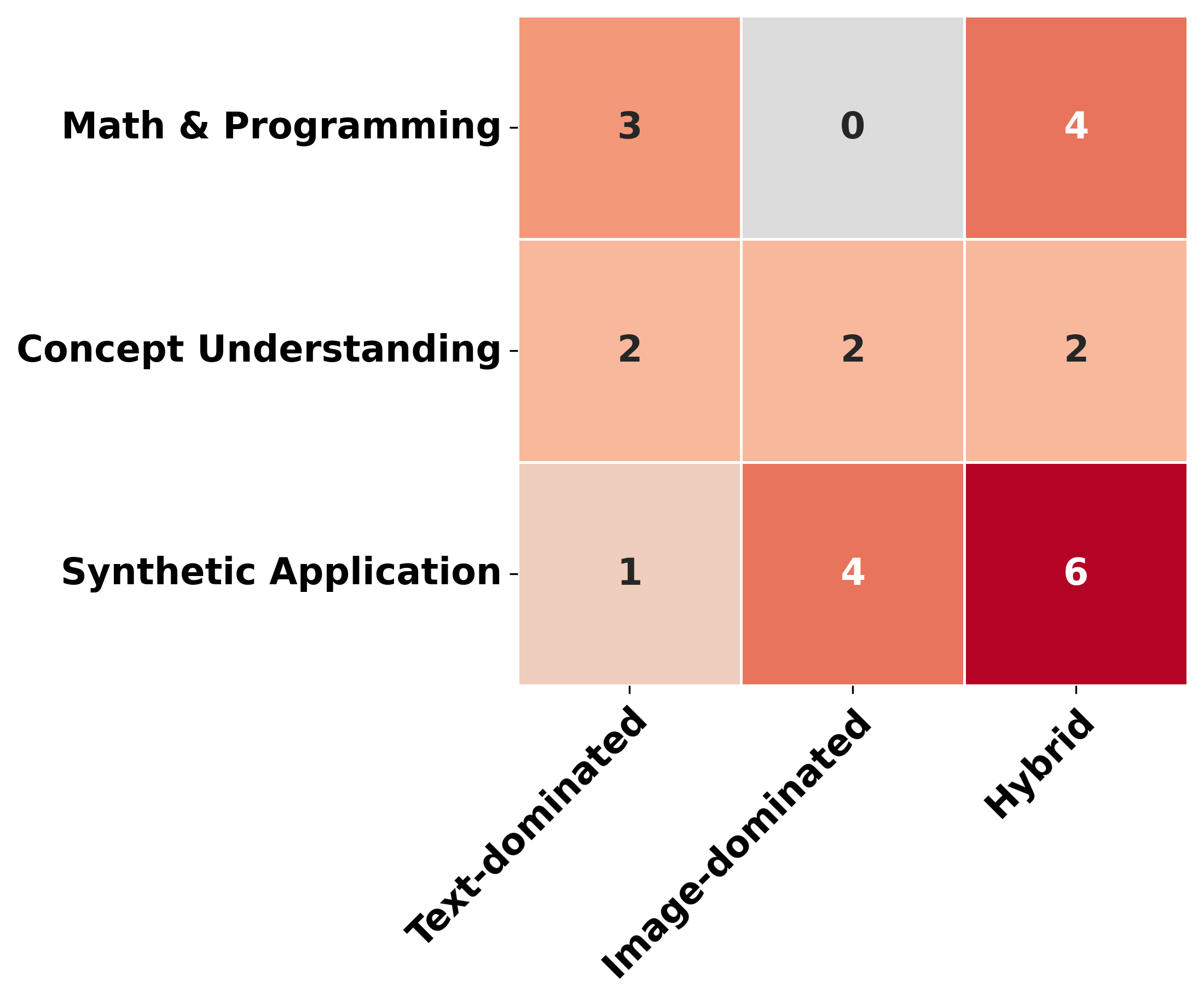}
        \caption{Category overlaps.}
        \label{fig:ring_pies}
    \end{subfigure}
    \hfill
    \begin{subfigure}{0.205\linewidth}
        \centering
        \includegraphics[width=\linewidth]{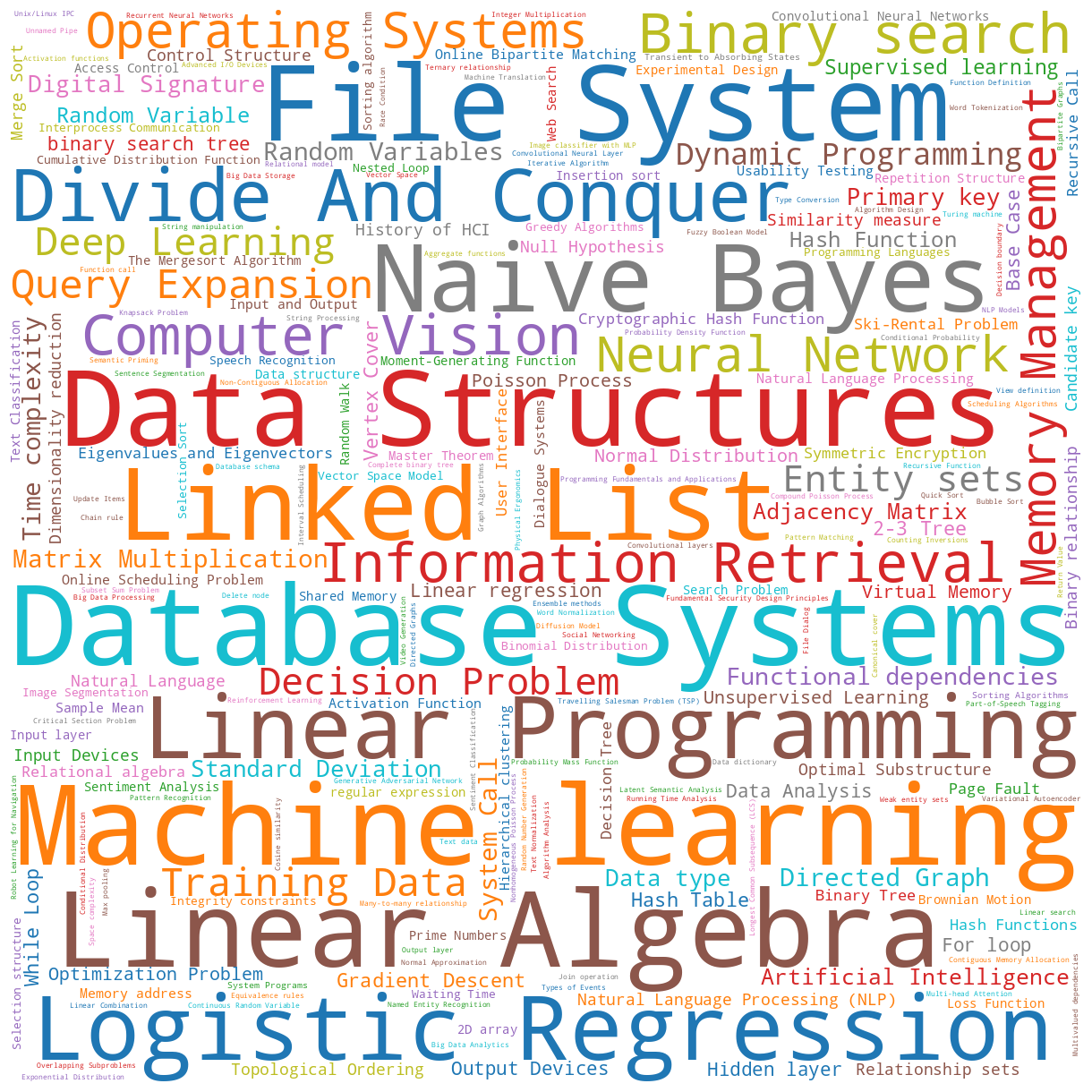}
        \caption{Word cloud of the concepts in S2M-Bench.}
        \label{fig:histograms}
    \end{subfigure}
    \hfill
    \begin{subfigure}{0.78\linewidth}
        \centering
        \includegraphics[width=\linewidth]{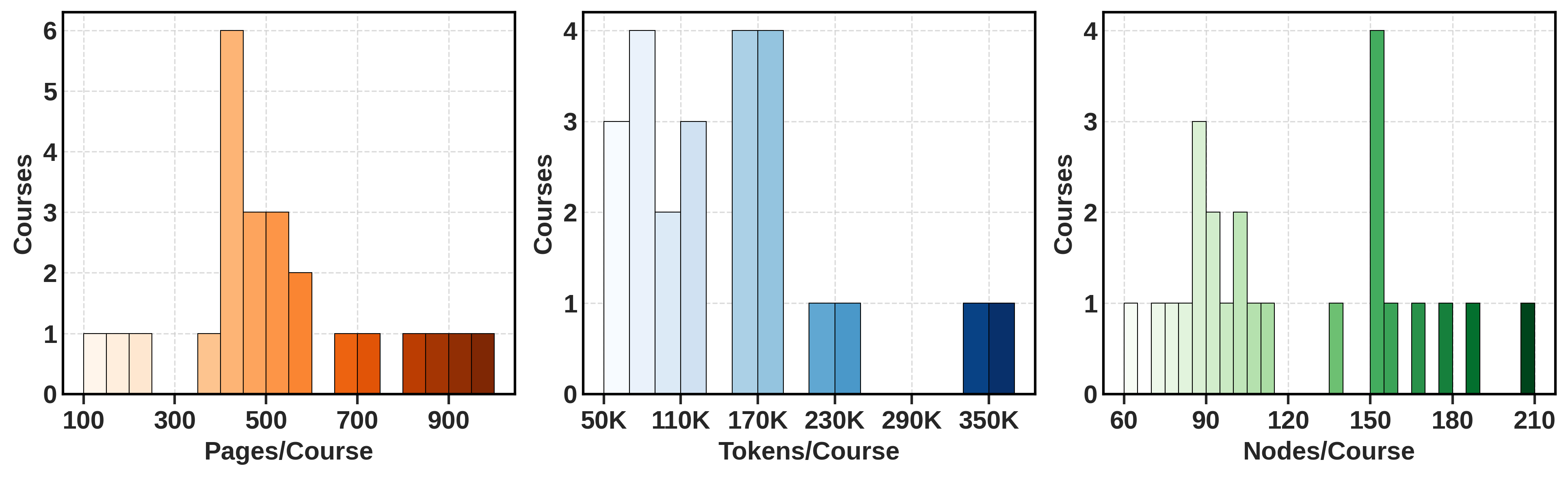}
        \caption{Distribution of pages, tokens, and concept nodes per course.}
        \label{fig:wordcloud}
    \end{subfigure}
    \caption{Further Statistic of S2M-Bench.}
    \label{fig:benchstat}
\end{figure*}

\begin{figure*}
    \centering
    \begin{subfigure}{\linewidth}
        \centering
        \includegraphics[width=\linewidth]{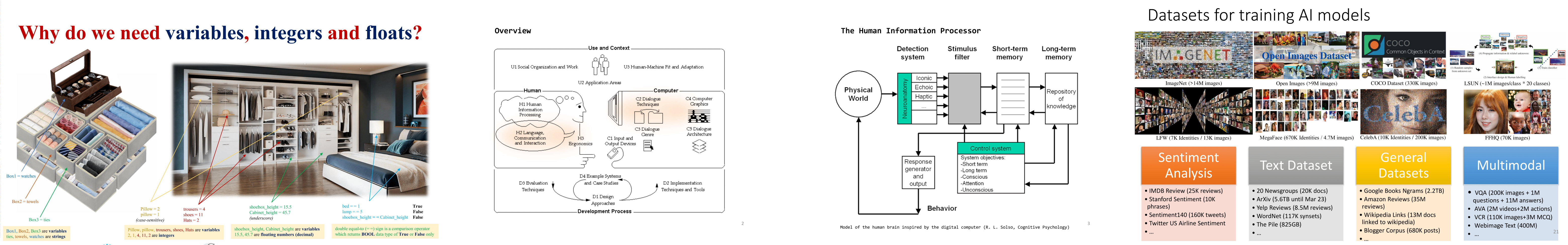}
        \caption{Image-dominated Slides.}
        \label{fig:ring_pies}
    \end{subfigure}
    \hfill
    \begin{subfigure}{\linewidth}
        \centering
        \includegraphics[width=\linewidth]{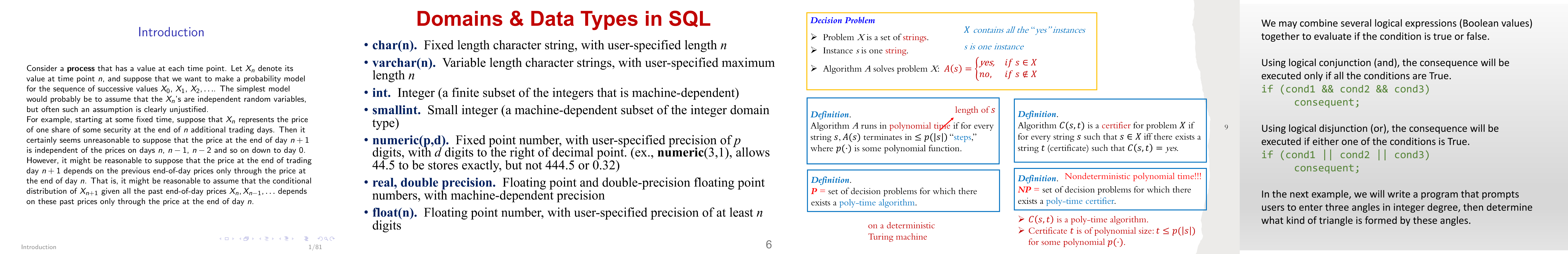}
        \caption{Text-dominated Slides.}
        \label{fig:histograms}
    \end{subfigure}
    \hfill
    \begin{subfigure}{\linewidth}
        \centering
        \includegraphics[width=\linewidth]{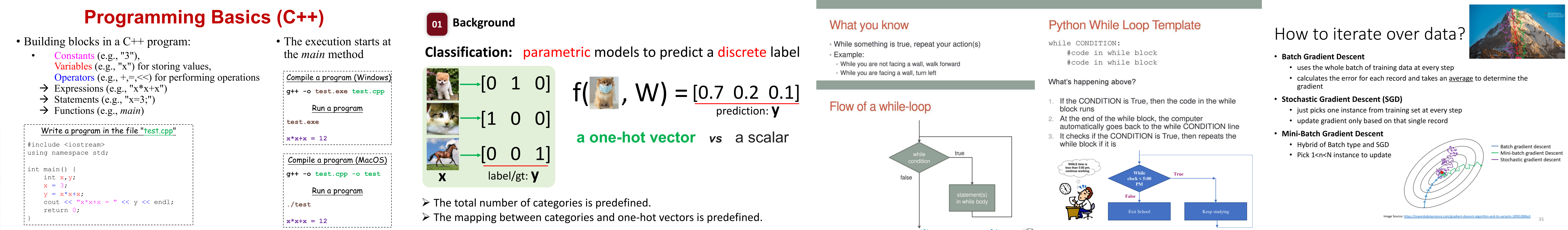}
        \caption{Hybrid Slides.}
        \label{fig:wordcloud}
    \end{subfigure}
    \caption{Examples of different categories in terms of content modality.}
    \label{fig:example_slides}
\end{figure*}

\section{Dataset Details}
\label{app:dataset details}
We further provide statistics on the dataset collected in S2M-Bench. S2M-Bench includes a total of 24 courses, 267 decks, 12,774 pages, and 3,571,497 tokens. The ground truth (GT) mind maps are constructed by human annotators for each course, with a total of 2,913 concept nodes. In Figure~\ref{fig:benchstat}, we visualize the course category distribution, category overlaps, word cloud, and the distribution of pages/course, tokens/course, and concept-nodes/course in human-labeled ground truth. In Figure~\ref{fig:example_slides}, we select some examples for each category in terms of the content modality. We also illustrate the principle of categorization in Table~\ref{tab:catdec}. The categorization of content modality is based on the overall style of all the slides of a course, rather than a single slide.

\begin{table*}
\small
\setlength{\tabcolsep}{5pt}
\centering

\begin{tabularx}{\textwidth}{l X}
\toprule
\textbf{Category} & \textbf{Description} \\
\midrule
Math \& Programming
  & Encompasses pure programming and foundational mathematics, featuring numerous examples and exercises, e.g., Linear Algebra, Calculus, C++ Programming, Python Programming. \\
\midrule
Concept Understanding
  & Centers on the memorization and comprehension of concepts and principles, with minimal computation or programming, e.g., Operating System, Computer Organization and Architecture. \\
\midrule
Synthetic Application
  & Integrates conceptual understanding and computational tasks, oriented toward specific application scenarios, e.g., Stochastic Processes for Investment, Machine Learning. \\
\midrule
Image-dominated
  & Primarily visual, with images (including formulas and charts) as the main elements; text appears only as captions or short phrases. \\
\midrule
Text-dominated
  & Comprises extensive textual paragraphs or long sentences across most slides. \\
\midrule
Hybrid
  & Contains both substantial text blocks and significant image elements in comparable proportions. \\

\bottomrule
\end{tabularx}
\caption{Description of each course category.}
\label{tab:catdec}
\end{table*}

\section{More Experiment Settings}
\label{app:moreexpset}
For the baselines, the original implementation of BookRAG and PageIndex only uses the original titles as the index names and neglects the name conciseness, which is unfair in our evaluation method. Thus, we further rephrase the node names of their results into concise concept representations through LLMs (i.e., the backbone models).

We conduct experiments on a server with 64 Intel(R) Xeon(R) Gold 6426Y CPUs and 8 NVIDIA GeForce RTX 4090 GPUs. All the LLM calls are through the official API of OpenAI and DeepSeek. All text embeddings used in our experiments are based on \textbf{text-embedding-3-small}, and all image rendering, including the visual refinement stage and VLM-Judge evaluation, are based on \textbf{Graphviz}. The output DPI is set to 35 (for refinement) and 100 (for evaluation), respectively. In cross-branch retrieval, $K=3$ and $\gamma_{sim}=0.6$. The refinement is set to two rounds. The LLM temperature is 0.1 during generation, and 0 during evaluation. All the prompts used for AutoMindMap, baselines, and evaluation are provided in the Appendix.

\section{Evaluation Metrics}

\subsection{Procedure for Concept Nodes Match}
\label{app:matchingalgo}
A prerequisite for GT-based metrics is to establish a matched node-pair set $\mathcal{C} \subseteq V^{\text{pred}} \times V^{\text{gt}}$ that aligns concepts with semantically equivalent meanings despite surface-form variations. To accommodate the inherent diversity in concept naming (e.g., abbreviations, synonyms, and paraphrases), we design a multi-stage matching algorithm that progressively refines matches while preserving precision. The algorithm initializes two candidate pools: $\mathcal{P}_{\text{gt}} = V^{\text{gt}}$ and $\mathcal{P}_{\text{pred}} = V^{\text{pred}}$. At each successful match, the matched pair is added to $\mathcal{C}$, and both nodes are removed from their respective pools (i.e., each node can be matched only once).

The matching proceeds in three stages: \textbf{(1) Exact match.} We perform case-insensitive, singular/plural-normalized string equality between nodes in $\mathcal{P}_{\text{gt}}$ and $\mathcal{P}_{\text{pred}}$. This captures identical concept names with minor inflectional variants. \textbf{(2) Abbreviations match.} For each remaining ground-truth node $v^{\text{gt}} \in \mathcal{P}_{\text{gt}}$, we invoke an LLM to examine its name and parent/child context. The LLM is prompted to infer potential full forms or common abbreviations (e.g., "ML" and "Machine Learning") and decide on a match. This stage handles semantic equivalence that is not captured by surface similarity. \textbf{(3) LLM-assisted similarity match.} For the remaining $v^{\text{gt}} \in \mathcal{P}_{\text{gt}}$, we compute two complementary similarity scores against every remaining predicted node $v^{\text{pred}} \in \mathcal{P}_{\text{pred}}$: \textbf{word-level Jaccard similarity}, and \textbf{cosine similarity of text embeddings}. For each similarity measure, we retain the top-$k$ predicted nodes (with $k=3$) whose score exceeds a threshold $\tau = 0.6$, forming a candidate pool of at most $2k$ candidates. The LLM then reviews these candidates along with their structural neighbors (parent/child nodes) to select the most appropriate match or reject all. If a match is accepted, the pair is added to $\mathcal{C}$ and removed from the pools; otherwise, the ground-truth node remains unmatched and is ultimately discarded. The LLM prompts are provided in Appendix~\ref{app:prompteval}.

\subsection{Calculation of GT-based Metrics}
\label{app:gtcal}

\noindent \textbf{HN-F1.} Let $\mathcal{C} \subseteq V^{\text{pred}} \times V^{\text{gt}}$ be the matched node-pair set. For each pair $(v_i^{\text{gt}}, v_i^{\text{pred}}) \in \mathcal{C}$, define
\begin{equation}
w(v_i^{\text{gt}}, v_i^{\text{pred}}) = 1 - \left| \frac{d(v_i^{\text{gt}})}{d^{\text{gt}}} - \frac{d(v_i^{\text{pred}})}{d^{\text{pred}}} \right|,
\end{equation}
where $d(\cdot)$ is the distance from the root and $d^{\text{gt}} = \max_{v \in V^{\text{gt}}} d(v)$, $d^{\text{pred}} = \max_{v \in V^{\text{pred}}} d(v)$. Then the corresponding precision, recall, and the F1-score are calculated by
\begin{equation}
\text{HN-P} = \frac{\sum_{(v_i^{\text{gt}}, v_i^{\text{pred}}) \in \mathcal{C}} w(v_i^{\text{gt}}, v_i^{\text{pred}})}{|V^{\text{pred}}|}, 
\end{equation}
\begin{equation}
\text{HN-R} = \frac{\sum_{(v_i^{\text{gt}}, v_i^{\text{pred}}) \in \mathcal{C}} w(v_i^{\text{gt}}, v_i^{\text{pred}})}{|V^{\text{gt}}|}, 
\end{equation}
\begin{equation}
\text{HN-F1} = 2 \cdot \frac{\text{HN-P} \cdot \text{HN-R}}{\text{HN-P} + \text{HN-R}}.
\end{equation}

\noindent \textbf{EC-F1.} An edge $(v_i^{\text{gt}}, v_j^{\text{gt}}) \in E^{\text{gt}}$ is \textit{covered} if both endpoints are matched in $\mathcal{C}$ and there exists a directed path from $v_i^{\text{pred}}$ to $v_j^{\text{pred}}$ in $\mathcal{M}^{\text{pred}}$. Coverage for $(v_i^{\text{pred}}, v_j^{\text{pred}}) \in E^{\text{pred}}$ is defined symmetrically. Then:
\begin{equation}
\text{EC-P} = \frac{|\{ e \in E^{\text{pred}} \mid e \text{ is covered} \}|}{|E^{\text{pred}}|},
\end{equation}
\begin{equation}
\text{EC-R} = \frac{|\{ e \in E^{\text{gt}} \mid e \text{ is covered} \}|}{|E^{\text{gt}}|}, 
\end{equation}
\begin{equation}
\text{EC-F1} = 2 \cdot \frac{\text{EC-P} \cdot \text{EC-R}}{\text{EC-P} + \text{EC-R}}.
\end{equation}

\noindent \textbf{MEC-F1.} Let $E_{\text{map}}^{\text{gt}} = \{ (u,v) \in E^{\text{gt}} \mid u,v \text{ are matched in } \mathcal{C} \}$ and define $E_{\text{map}}^{\text{pred}}$ analogously. Then:
\begin{equation}
\text{MEC-P} = \frac{|\{ e \in E_{\text{map}}^{\text{pred}} \mid e \text{ is covered} \}|}{|E_{\text{map}}^{\text{pred}}|}, 
\end{equation}
\begin{equation}
\text{MEC-R} = \frac{|\{ e \in E_{\text{map}}^{\text{gt}} \mid e \text{ is covered} \}|}{|E_{\text{map}}^{\text{gt}}|}, 
\end{equation}
\begin{equation}
\text{MEC-F1} = 2 \cdot \frac{\text{MEC-P} \cdot \text{MEC-R}}{\text{MEC-P} + \text{MEC-R}}.
\end{equation}

\subsection{Detailed Criteria for VLM-Judge}
\label{app:rubrics}
The detailed scoring criteria for the three dimensions of VLM-Judge are listed in Table~\ref{tab:metrics-rubrics}. All VLM-Judge evaluations, as well as human evaluations (Appendix~\ref{app:human}), are based on the criteria and Likert scales.

\begin{table*}[htbp]

\small
\centering
\renewcommand{\arraystretch}{1.3}
\begin{tabularx}{\textwidth}{@{} p{2.3cm} X @{}}
\toprule
\textbf{Dimensions} & \multicolumn{1}{c}{\textbf{Criteria}} \\
\midrule
\textbf{Content} \newline \textbf{Accuracy} &
\begin{itemize}[nosep, leftmargin=*, label={--}]
    \item All extracted concept names should be concise and accurate,
          generally proper nouns/noun phrases. The following situations
          should be avoided:
          \begin{itemize}[nosep, leftmargin=*, label={--}]
              \item Single abbreviations and broad words (e.g., introduction,
                    motivation, application, example) without referents;
              \item Uninformative modifier words/section indexes;
              \item Long sentences, questions, and expressions without
                    summarization or compression;
              \item One-phrase concatenation of multiple concepts.
          \end{itemize}
    \item Reasonable hierarchy and affiliation relation between parent and
          child concept, without confusing relation between parent-child/siblings.
    \item Concepts should be representative and recapitulatory, rather than
          lots of unimportant, course-irrelevant, noisy entities or mutually
          overlapped/similar concepts.
\end{itemize}
\\ \midrule
\textbf{Structure} \newline \textbf{Quality} &
\begin{itemize}[nosep, leftmargin=*, label={--}]
    \item Poor structures should be avoided, including:
          \begin{itemize}[nosep, leftmargin=*, label={--}]
              \item Too dense and disordered relations without clear organization;
              \item Overload child nodes under one parent without hierarchy
                    design, or cascaded long one-child chains;
              \item Too deep (with more than 7 levels) or shallow flat
                    hierarchy (with only 2--3 levels);
              \item Extremely unbalanced branch scale/depth;
              \item Multiple roots, ring/bi-directional edges, isolated nodes.
          \end{itemize}
    \item All local concepts and relations are organized into an overall
          coherent logic, with meaningful taxonomy and progressive hierarchy.
    \item Allow for proper structural flexibility and scale difference
          between branches, rather than a completely neat structure.
\end{itemize}
\\ \midrule
\textbf{Pedagogical} \newline \textbf{Efficiency} &
\begin{itemize}[nosep, leftmargin=*, label={--}]
    \item Proper overall concept scale with clear visual effect on
          hierarchy, rather than chaotic, dense graphs, or just superficial
          concepts without delving.
    \item Generalizable core topics and framework are presented obviously,
          can be captured quickly from the mind map, rather than a laborious
          search.
    \item Reasonable cross-links across branches that promote concept
          association, rather than a pure tree. The proportion is proper
          rather than dense.
    \item Core knowledge should be emphasized structurally (e.g., include
          more and deeper nodes) rather than all concepts being equally
          important.
    \item Overall organization is consistent with human cognitive habits,
          leading to less confusion.
\end{itemize}
\\ \bottomrule
\end{tabularx}
\caption{Detailed criteria for VLM-Judge. These criteria are based on cognitive science theories in Appendix~\ref{app:theory}.}

\label{tab:metrics-rubrics}
\end{table*}

\begin{table*}
\small
\setlength{\tabcolsep}{9pt}
\centering

\begin{tabular}{@{}l ccc cc cccc@{}}
\toprule
 \multirow{2}{*}{\textbf{Methods}} & \multicolumn{3}{c}{\textbf{GT-based}} & \multicolumn{2}{c}{\textbf{Structure conformity}} & \multicolumn{4}{c}{\textbf{VLM-Judge}} \\
\cmidrule(lr){2-4} \cmidrule(lr){5-6} \cmidrule(lr){7-10}
 & \textbf{HN-F1} & \textbf{EC-F1}&\textbf{MEC-F1} & \textbf{Pass@1} & \textbf{SLF$\downarrow$} & \textbf{CA} & \textbf{SQ} & \textbf{PE} & \textbf{Avg.} \\
\midrule
 \textbf{Direct}     & 22.18 & 11.55& 47.24 & 72.73 & 54.32 & 2.86 & 2.36 & 2.09 & 2.44 \\
 \textbf{Chunk-Merge}  & 18.20 & 11.26& \underline{57.66} & 66.67 & \underline{20.85} & \underline{3.29} & \underline{3.00} & 2.63 & 2.97 \\

 \textbf{Highlights-only}    & \underline{26.93} & \underline{15.09} & 53.15 & \underline{95.83} & 45.17& 3.21 & 2.92 & \underline{2.92} & \underline{3.01} \\
 \textbf{BookRAG}         & 23.58 & 9.78 & 42.44 & 82.60 & 22.55 & 1.74 & 1.48 & 1.35 & 1.52 \\
 \textbf{PageIndex}      & 25.85 & 8.54 & 36.25 & \textbf{100.00} & 40.52 & 2.33 & 2.00 & 1.71 & 2.01 \\
\midrule
 \textbf{AutoMindMap}   & \textbf{35.39} & \textbf{19.59} & \textbf{68.48} & \textbf{100.00} & \textbf{3.29} & \textbf{3.54} & \textbf{3.67} & \textbf{3.63} & \textbf{3.61} \\
\bottomrule
\end{tabular}
\caption{Experimental results of different methods, \textit{GPT-5.4-mini} serves as both the backbone model and VLM for refinement in AutoMindMap. Best results are \textit{bolded}, second-best results are \textit{underlined}.}
\label{tab:resgpt54}
\end{table*}

\begin{table*}
\small
\setlength{\tabcolsep}{5pt}
\centering

\begin{tabular}{@{}l c ccc cc cccc@{}}
\toprule
 \multirow{2}{*}{\textbf{Methods}} & \multirow{2}{*}{\textbf{Models}} & \multicolumn{3}{c}{\textbf{GT-based}} & \multicolumn{2}{c}{\textbf{Structure conformity}} & \multicolumn{4}{c}{\textbf{VLM-Judge}} \\
\cmidrule(lr){3-5} \cmidrule(lr){6-7} \cmidrule(lr){8-11}
& & \textbf{HN-F1} & \textbf{EC-F1}&\textbf{MEC-F1} & \textbf{Pass@1} & \textbf{SLF$\downarrow$} & \textbf{CA} & \textbf{SQ} & \textbf{PE} & \textbf{Avg.} \\
\midrule

\multirow{3}{*}{\textbf{Direct}} 
 &\textbf{GPT-5.4-mini}  & 22.18 & 11.55& 47.24 & 72.73 & 54.32 & 2.86 & 2.36 & 2.09 & 2.44 \\

 &\textbf{DS-V4-Flash}  & 33.43 & \underline{20.67} & 64.50 & 79.16 & 20.22 & 3.17 & 3.00 & 2.88 & 3.01 \\

 &\textbf{DS-V4-Pro}  & \underline{34.98} & 18.71 & \underline{68.61} & \underline{95.83} & \underline{4.30} & \underline{3.38} & \underline{3.38} & \underline{3.21} & \underline{3.32} \\

\midrule
\multirow{3}{*}{\textbf{AutoMindMap}}
 &\textbf{GPT-5.4-mini}  & 35.39 &  19.59 & 68.48 & \textbf{100.00} & 3.29 & 3.54 & 3.67 & 3.63 & 3.61\\

 &\textbf{DS-V4-Flash + GPT-4.1-mini}   & 37.80 & 22.96 & 74.01 & \textbf{100.00} &\textbf{1.82}  & 3.58 & 3.63 & \textbf{3.71} & 3.64 \\

 &\textbf{DS-V4-Flash + GPT-5.4-mini}    & \textbf{39.88} & \textbf{25.12} &\textbf{74.34} &\textbf{100.00} & 1.93  & \textbf{3.83} &\textbf{3.71} & 3.63 & \textbf{3.72} \\

\bottomrule
\end{tabular}
\caption{Performances of different models on the Direct method and AutoMindMap. The best result in the Direct method is \textit{underlined}, and the global optimum is in \textit{bold}. "DS" abbreviates "DeepSeek".}

\label{tab:diffmodel}
\end{table*}

\section{More Experiments}
\label{app:moreexp}
We discuss more experimental results, including the performance of AutoMindMap with different LLMs and on different course categories, the cost and stability, the ablation study table, and the detailed results of GT-based metrics. \textbf{If not mentioned otherwise, the model setting is DeepSeek-V4-Flash (baselines, and backbone of AutoMindMap) + GPT-4.1-mini (VLM for refinement in AutoMindMap).}

\subsection{Performance on Other LLMs}
To show the performance of AutoMindMap on different models, we first conduct the comparison experiments on pure GPT-5.4-mini (i.e., used as both the backbone model and VLM-based refinement model); the result is shown in Table~\ref{tab:resgpt54}. Besides, to examine the end-to-end generation performance of the recent advanced reasoning models on the Slides2MindMap task, we run the Direct method with DeepSeek-V4-Pro. The performances of different models in the Direct method and AutoMindMap are summarized in Table~\ref{tab:diffmodel}.

The results show that AutoMindMap still outperforms baseline methods with GPT-5.4-mini, but the performance of GPT-5.4-mini is worse than DeepSeek-V4-Flash when serving as the backbone LLM. A possible reason is that GPT-5.4-mini has limited capabilities in long context scenarios. In addition, more advanced reasoning models like DeepSeek-V4-Pro can significantly outperform other models in the Direct method, but there is still a gap with AutoMindMap (with DeepSeek-V4-Flash), especially in VLM-Judge and structure conformity metrics. Stronger LLMs can extract more faithful knowledge, but fail in further structural organization and pedagogy-oriented generation. The result shows the effectiveness of AutoMindMap.

\subsection{Performance Across Different Course Categories}
For the different course categories in S2M-Bench, we conduct comparisons in terms of the knowledge types and content modality. The results are shown in Table~\ref{tab:coursetype com} and Table~\ref{tab:modal com}.

AutoMindMap outperforms the Direct method and shows eligible generalization across different categories. 
However, the performance differences in different categories still exist.

Both the Direct method and AutoMindMap show significant differences in performance on different types of courses. In Table~\ref{tab:coursetype com}, the Direct approach performs best in \textit{math \& programming} courses, while performing worst in concept understanding, but AutoMindMap performs best in concept understanding, which shows concept-intensive courses inherently have a more complex conceptual hierarchical structure for direct identification, and AutoMindMap enhances the capacity to identify complex hierarchy.

In Table~\ref{tab:modal com}, the differences are comparatively smaller. The Direct method shows close performance across three categories, and AutoMindMap performs better in \textit{image-dominated} slides. The performance might benefit from the parsing effect of MinerU on visual information.

\begin{table*}
\small
\setlength{\tabcolsep}{8pt}
\centering

\label{tab:content_categories}
\begin{tabular}{@{}l l c c c c c c c@{}}
\toprule
\multirow{2}{*}{\textbf{Methods}} & \multirow{2}{*}{\textbf{Categories}} & \multicolumn{3}{c}{\textbf{GT-based}} & \multicolumn{4}{c}{\textbf{VLM-Judge}} \\
\cmidrule(lr){3-5}\cmidrule(lr){6-9}
& & \textbf{HN-F1} & \textbf{EC-F1} & \textbf{MEC-F1} & \textbf{CA} & \textbf{SQ} & \textbf{PE} & \textbf{Avg.} \\
\midrule

\multirow{4}{*}{\textbf{Direct}}
&  Math \& Programming & \textbf{38.92} & 20.92 & \textbf{65.83} & \textbf{3.29} & \textbf{3.14} & \textbf{3.14} & \textbf{3.19} \\

&  Concept Understanding & 28.42 & 15.78 & 63.68 & 3.00 & 2.67 & 2.50 & 2.72 \\

&  Synthetic Application & 32.67 & \textbf{23.17} & 64.10 & 3.18 & 3.09 & 2.91 & 3.06 \\
&  Total & {33.43} & {20.67} & {64.50} & {3.17} & {3.00} & {2.88} &{3.01} \\

\midrule
\multirow{4}{*}{\textbf{AutoMindMap}}
&  Math \& Programming & \textbf{40.68} & 23.85 & 69.71 & 3.57 & 3.57 & 3.86 & 3.67 \\

&  Concept Understanding & 40.23 & \textbf{25.33} & \textbf{76.17} & \textbf{3.83} & \textbf{4.00} & \textbf{4.00} & \textbf{3.94} \\

&  Synthetic Application & 34.64 & 21.10 & 75.57 & 3.45 & 3.45 & 3.45 & 3.45 \\

& {Total} & 37.80 & 22.96 & 74.01 & {3.58} & {3.63} & {3.71} & {3.64}\\
\bottomrule
\end{tabular}
\caption{Performance comparisons on knowledge type-based categories.}
\label{tab:coursetype com}
\end{table*}

\begin{table*}
\small
\setlength{\tabcolsep}{9pt}
\centering

\label{tab:modality_categories}
\begin{tabular}{@{}l l c c c c c c c@{}}
\toprule
\multirow{2}{*}{\textbf{Methods}} & \multirow{2}{*}{\textbf{Categories}} & \multicolumn{3}{c}{\textbf{GT-based}} & \multicolumn{4}{c}{\textbf{VLM-Judge}} \\
\cmidrule(lr){3-5}\cmidrule(lr){6-9}
& & \textbf{HN-F1} & \textbf{EC-F1} & \textbf{MEC-F1} & \textbf{CA} & \textbf{SQ} & \textbf{PE} & \textbf{Avg.} \\
\midrule

\multirow{4}{*}{\textbf{Direct}}
&  Image-dominated & \textbf{40.34} & \textbf{23.31} & 63.25 & \textbf{3.17} & 2.83 & 2.67 & 2.89 \\
&  Text-dominated & 30.44 & 19.72 & \textbf{66.86} & \textbf{3.17} & 3.00 & \textbf{3.00} & \textbf{3.06} \\
&  Hybrid & 31.47 & 19.82 & 63.95 & \textbf{3.17} & \textbf{3.08} & 2.92 & \textbf{3.06} \\
& {Total} & {33.43} & {20.67} &{64.50} & {3.17} & {3.00} & {2.88} & {3.01} \\

\midrule
\multirow{4}{*}{\textbf{AutoMindMap}}
&  Image-dominated & \textbf{43.33} & \textbf{25.54} & 73.95 & \textbf{3.83} & \textbf{3.83} & \textbf{3.83} & \textbf{3.83} \\
&  Text-dominated & 34.49 & 19.51 & \textbf{75.11} & 3.67 & 3.67 & 3.67 & 3.67 \\
&  Hybrid & 36.69 & 23.40 & 73.49 & 3.42 & 3.50 & 3.67 & 3.53 \\
& {Total} & 37.80 & 22.96 & 74.01 & {3.58} & {3.63} & {3.71} & {3.64}\\
\bottomrule
\end{tabular}
\caption{Performance comparisons on content modality-based categories.}
\label{tab:modal com}
\end{table*}

\subsection{Cost Analysis}
Table~\ref{tab:cost} lists prompt tokens, completion tokens, and runtime averaged per course. AutoMindMap incurs higher cost than single-pass methods (Direct, Highlights-only) due to its iterative agentic architecture, thus reducing the probability of illegal structure generation to 0 (with $\text{pass@1}=100\%$), and obviously narrowing the gap to human expert (e.g., average VLM-Judge metrics gap decreases from 0.88 to 0.17 in Table~\ref{tab:comparison}, and GT-based metrics gap also decreases in the sampled courses in Table~\ref{tab:human+human}). When compared against multi-step iteration methods (e.g., PageIndex), AutoMindMap is substantially more efficient. The theoretical upper bound of average API cost for AutoMindMap is approximately \$0.08 per course for the experimental model setting, compared with \$0.026 of the Direct method, and in practice the actual costs are often lower due to high cache hit rates from repeated prompt prefixes. Besides, the runtime of AutoMindMap is sufficient for offline generation scenarios. 

Nevertheless, we should admit that the cost efficiency is not a strength of AutoMindMap, and the quality improvement comes at a non-negligible cost, which is a trade-off inherent to the complexity of the Slides2MindMap task. Developing more cost-efficient generation algorithms remains an important future direction.

\begin{table}
\small
\setlength{\tabcolsep}{4pt}
\centering

\begin{tabular}{l c c c}
\toprule
\textbf{Methods} & \textbf{Tokens (P)} & \textbf{Tokens (C)} & \textbf{Runtime(s)} \\
\midrule
 \textbf{Direct} & 160.14K & 8.87K & 82\\
 \textbf{Chunk-Merge} & 178.22K & 50.00K & 461\\
 \textbf{Highlights-only} & 7.60K & 14.22K & 119\\
 \textbf{BookRAG} & 48.63K & 17.52K & 398\\
 \textbf{PageIndex} & 560.36K & 114.51K & 1482\\
\midrule
 \textbf{AutoMindMap} & 316.84K & 64.22K & 679\\
\bottomrule
\end{tabular}
\caption{Cost Analysis. The values are averaged by course. \textit{P} and \textit{C} abbreviate \textit{Prompt} and \textit{Completion}, respectively.}
\label{tab:cost}
\end{table}

\subsection{Stability Analysis}
To assess generation stability, we select three courses spanning different categories and run AutoMindMap five times per course with temperature raised to 0.5 to increase randomness. Table~\ref{tab:steady} reports the range, mean, and standard deviation of the results. \textbf{Range} averages the best and the worst values of each course, rather than the values averaged over a single run of all courses. The results show that multiple runs can still maintain eligible generation robustness even with increased randomness. In particular, Pass@1 shows reliable structure quality.

\begin{table*}
\small
\setlength{\tabcolsep}{3pt}
\centering

\begin{tabular}{@{}l ccc cc cccc@{}}
\toprule
 \multirow{2}{*}{\textbf{Indicators}} & \multicolumn{3}{c}{\textbf{GT-based}} & \multicolumn{2}{c}{\textbf{Structure conformity}} & \multicolumn{4}{c}{\textbf{VLM-Judge}} \\
\cmidrule(lr){2-4} \cmidrule(lr){5-6} \cmidrule(lr){7-10}
 & \textbf{HN-F1} & \textbf{EC-F1}&\textbf{MEC-F1} & \textbf{Pass@1} & \textbf{SLF$\downarrow$} & \textbf{CA} & \textbf{SQ} & \textbf{PE} & \textbf{Avg.} \\
\midrule

\textbf{Range} & [33.53, 38.78] & [24.55, 28.30] & [80.15, 85.40] &  100.00 & [1.81, 4.59] &  [3, 4] &  [3, 4] &  [3, 3.67]  &  [3, 3.67] \\

\textbf{Mean} & 36.72 & 25.86 & 82.81   & 100.00 & 2.99 & 3.40  &  3.40 &  3.13  & 3.31       \\ 

\textbf{Std.} & 1.93 & 1.53 & 2.26  & 0.00 &  0.99 &    0.14 & 0.21 & 0.16  &  0.14     \\

\bottomrule
\end{tabular}
\caption{Stability analysis for AutoMindMap. The experiments were run 5 times on 3 selected courses at temperature = 0.5.}
\label{tab:steady}
\end{table*}

\subsection{Detailed Results of Ablation Study}
The results of the ablation study are detailed in Table~\ref{tab:abl}, corresponding to Figure~\ref{fig:abl}.

\begin{table*}
\small
\setlength{\tabcolsep}{9pt}
\centering

\label{tab:ablation_main}
\begin{tabular}{@{}l ccc cc cccc@{}}
\toprule
\multirow{2}{*}{\textbf{Methods}} & \multicolumn{3}{c}{\textbf{GT-based}} & \multicolumn{2}{c}{\textbf{Structure conformity}} & \multicolumn{4}{c}{\textbf{VLM-Judge}} \\
\cmidrule(lr){2-4} \cmidrule(lr){5-6} \cmidrule(lr){7-10}
& \textbf{HN-F1} & \textbf{EC-F1} & \textbf{MEC-F1} & \textbf{Pass@1} & \textbf{SLF$\downarrow$} & \textbf{CA} & \textbf{SQ} & \textbf{PE} & \textbf{Avg.} \\
\midrule
 \textbf{w/o vision}      & 34.69 & \underline{20.99} & 73.50 & \underline{95.83} & \textbf{1.80} & \underline{3.50} & \underline{3.50} & \underline{3.38} & \underline{3.46} \\
 \textbf{w/o tools}       & 30.85 & 16.15 & 70.21 & \textbf{100.00} & 3.76 & 3.29 & 3.25 & 3.25 & 3.26 \\
 \textbf{w/o refinement}  & 35.09 & 18.40 & 64.20 & \textbf{100.00} & 2.34 & 3.17 & 3.25 & 2.96 & 3.13 \\
 \textbf{w/o summary}     & 31.46 & 18.44 & \textbf{76.66} & \textbf{100.00} & 2.02 & 3.38 & 3.17 & 3.00 & 3.18 \\
 \textbf{w/o skeleton}    & \underline{35.30} & 16.15 & 66.56 & 91.67 & 2.99 & 3.08 & 3.08 & 3.04 & 3.07 \\
\midrule
 \textbf{AutoMindMap}   & \textbf{37.80} & \textbf{22.96} & \underline{74.01} & \textbf{100.00} & \underline{1.82} & \textbf{3.58} & \textbf{3.63} & \textbf{3.71} & \textbf{3.64} \\
\bottomrule
\end{tabular}
\caption{Ablation study results. Best results are \textit{bolded}, second-best are \textit{underlined}.}
\label{tab:abl}
\end{table*}

\subsection{Detailed Results of GT-based Metrics}
\label{app:gtdetail}
As GT-based metrics HN-F1, EC-F1, and MEC-F1 cannot fully reflect the recall and precision of concept nodes and edges, we further list the full metrics in terms of comparison with the GT data in  Table~\ref{tab:gt_all_com} (comparison experiment) and Table~\ref{tab:gt_all_abl} (ablation study). N-P, N-R, and N-F1 are the node precision, recall, and F1-score without hierarchy-aware weights.

The two tables show a trade-off between precision and recall in terms of both nodes and edges. A detailed empirical statistic can be found in Figure~\ref{fig:corb}. The tables also indicate that the baselines can achieve high knowledge recall or precision, but cannot strike a balance and maintain global coherence. Some methods (e.g., Chunk-Merge) focus on too fine-grained knowledge, which may lead to lower precision, and overly conservative methods (e.g., Highlights-only) may lead to lower recall. In the ablation study, the trade-off also exists. Specifically, the node recall drops after refinement, but other metrics increase, which means a better balance between precision and recall, but also erroneously removes some nodes to streamline the structure. 

Besides, AutoMindMap cannot achieve the best performance in all the detailed metrics, but remains top-2 in most of the metrics, and can maintain a good balance between recall and precision. Compared with N-P, N-R, and N-F1, AutoMindMap performs even better than baselines in the metrics with hierarchy-aware weights, which further confirms that AutoMindMap can not only extract appropriate concepts from slides but also identify the hierarchies.

\begin{table*}
\small
\setlength{\tabcolsep}{4pt}
\centering

\label{tab:detailed_results}
\begin{tabular}{@{}l c c c c c c c c c c c c@{}}
\toprule
\bfseries Methods & \bfseries N-P & \bfseries N-R & \bfseries N-F1 & \bfseries HN-P & \bfseries HN-R & \bfseries \textcolor{red}{HN-F1} & \bfseries EC-P & \bfseries EC-R & \bfseries \textcolor{red}{EC-F1} & \bfseries MEC-P & \bfseries MEC-R & \bfseries \textcolor{red}{MEC-F1} \\

\midrule
 \bfseries Direct     & \textbf{44.77} & 49.96 & \underline{46.06} & \underline{32.62} & 36.19 & \underline{33.43} & \textbf{24.36} & 19.55 & \underline{20.67} & 75.02 & 58.93 & 64.50 \\

 \bfseries Chunk-Merge     & 26.45 & 59.64 & 34.22 & 19.43 & 43.33 & 25.03 & 12.80 & \underline{29.17} & 16.21 & 76.31 & \underline{62.83} & \underline{68.10} \\

 \bfseries Highlights-only & 39.63 & 30.16 & 32.03 & 29.31 & 22.23 & 23.63 & \underline{23.61} & 11.92 & 14.50 & \textbf{84.10} & 57.50 & 66.25 \\

 \bfseries BookRAG         & 26.03 & 52.51 & 32.20 & 20.39 & 40.41 & 25.12 & 9.47 & 13.68 & 10.12 & 73.01 & 37.74 & 47.85 \\

 \bfseries PageIndex       & 23.45 & \underline{60.08} & 29.88 & 18.98 & \underline{48.17} & 24.15 & 11.08 & 20.30 & 11.66 & \underline{81.50} & 45.83 & 57.11 \\

\midrule
 \bfseries AutoMindMap     & \underline{41.31} & \textbf{61.21} & \textbf{47.69} & \textbf{34.65} & \textbf{50.94} & \textbf{39.88} & 21.32 & \textbf{34.70} & \textbf{25.12} & 74.37 & \textbf{75.61} & \textbf{74.34} \\
\bottomrule
\end{tabular}
\caption{Full GT-based metrics in comparison experiment. The model combination is DeepSeek-V4-Flash + GPT-5.4-mini.}
\label{tab:gt_all_com}
\end{table*}

\begin{table*}
\small
\setlength{\tabcolsep}{4pt}
\centering

\label{tab:ablation_fine}
\begin{tabular}{@{}l c c c c c c c c c c c c@{}}
\toprule
\bfseries Methods & \bfseries N-P & \bfseries N-R & \bfseries N-F1 & \bfseries HN-P & \bfseries HN-R & \bfseries \textcolor{red}{HN-F1} & \bfseries EC-P & \bfseries EC-R & \bfseries \textcolor{red}{EC-F1} & \bfseries MEC-P & \bfseries MEC-R & \bfseries \textcolor{red}{MEC-F1} \\
\midrule
 \bfseries w/o vision      & \textbf{42.92} & 44.61 & 42.01 & \textbf{35.67} & 36.57 & 34.69 & \textbf{21.39} & 23.25 & \underline{20.99} & 75.86 & 72.37 & 73.50 \\

 \bfseries w/o tools       & \underline{40.52} & 36.17 & 37.38 & 33.39 & 29.87 & 30.85 & 17.75 & 16.16 & 16.15 & 74.33 & 68.35 & 70.21 \\

 \bfseries w/o refinement  & 35.80 & \textbf{63.31} & \underline{43.89} & 28.74 & \textbf{50.36} & 35.09 & 14.80 & \underline{28.62} & 18.40 & 69.33 & 61.65 & 64.20 \\

 \bfseries w/o summary     & 37.19 & 42.43 & 37.98 & 30.93 & 34.94 & 31.46 & 17.42 & 21.42 & 18.44 & \textbf{79.76} & \textbf{75.30} & \textbf{76.66} \\

 \bfseries w/o skeleton    & 38.87 & 49.65 & 42.50 & 32.28 & 41.24 & \underline{35.30} & 13.95 & 21.58 & 16.15 & 71.37 & 63.29 & 66.56 \\

\midrule
 \bfseries AutoMindMap   & 40.48 & \underline{57.14} & \textbf{45.59} & \underline{33.68} & \underline{47.20} & \textbf{37.80} & \underline{19.61} &\textbf{31.18}  & \textbf{22.96} & \underline{76.66} & \underline{72.42} & \underline{74.01} \\
\bottomrule
\end{tabular}
\caption{Full GT-based metrics in ablation study. The model combination is DeepSeek-V4-Flash + GPT-4.1-mini.}
\label{tab:gt_all_abl}
\end{table*}

\section{Evaluation Reliability Analysis}
\label{app:Evaluation Reliability Analysis}
\subsection{Human-Model Evaluation Agreement}
To demonstrate the evaluation effectiveness of VLM-Judge, we invite two human annotators to score 48 generated results for 8 courses. Human annotators and models are provided with the same criteria, which emphasize the quality of intrinsic structure, content, and pedagogy efficiency, ignoring pure visual rendering aesthetics. The scores are based on the 5-level Likert scale.

For the scoring results, we use the \textbf{Quadratic Weighted Kappa Coefficient}~\citep{Viera2005UnderstandingIA} to measure the consistency. Table~\ref{tab:Kappa} shows the results of human-human consistency and human-model consistency. The results demonstrate an eligible evaluation quality of VLM-Judge. Although the potential biases of VLM-Judge (e.g., aesthetic preferences, same-family model preferences) cannot be completely removed, the result already demonstrates effective mitigation. Besides, human-human consistency is slightly higher than human-model consistency. SQ shows the lowest consistency for both categories, indicating greater subjective variability.

\label{app:human}

\begin{table}[t]
\small
\setlength{\tabcolsep}{5pt}
\centering

\begin{tabular}{cccccc}
\toprule
\textbf{Categories} & \textbf{CA} & \textbf{SQ} & \textbf{PE} & \textbf{Avg.} &  \\
\midrule
\textbf{Human and VLM} & 0.811 & 0.774 & 0.835 & 0.807 &  \\
\textbf{Human and Human} & 0.862 & 0.790 & 0.883 & 0.845 &  \\
\bottomrule
\end{tabular}
\caption{Quadratic Weighted Kappa for consistency checking for VLM-as-a-Judge.}
\label{tab:Kappa}
\end{table}

\subsection{VLM-Judge with Other Models}
To further demonstrate the effectiveness of the VLM-Judge and eliminate same-family model preferences, we re-evaluate the results of the main experiment using Qwen3.5-Plus. Although there are small numerical differences, the overall ranking is still highly consistent with the main experiment (Table~\ref{tab:comparison}). Human experts still achieve the best results.

\begin{table}[t]
\small
\setlength{\tabcolsep}{7pt}
\centering

\label{tab:vlm_diff}
\begin{tabular}{l c c c c}
\toprule
\textbf{Methods} & \textbf{CA} & \textbf{SQ} & \textbf{PE} & \textbf{Avg.} \\
\midrule
 \textbf{Human} & 4.04 & 3.75 & 3.88 & 3.83\\
\midrule
 \textbf{Direct} & \underline{3.50} & 2.93 & 3.06 & 3.16\\
 \textbf{Chunk-Merge} & 3.18 & 2.75 & 2.69 & 2.87\\
 \textbf{Highlights-only} & 3.38 & \underline{3.00} & \underline{3.13} & \underline{3.17}\\
 \textbf{BookRAG} & 2.13 & 1.81 & 1.81 & 1.92\\
 \textbf{PageIndex} & 2.13 & 1.94 & 1.94 & 2.00\\ 
\midrule
    \textbf{AutoMindMap}\\
 + \textit{GPT-4.1-mini} & 3.80 & 3.23 & 3.23 & 3.42\\
 + \textit{GPT-5.4-mini} & \textbf{3.94} & \textbf{3.31} & \textbf{3.56} & \textbf{3.60}\\
\bottomrule
\end{tabular}
\caption{VLM-Judge with \textit{Qwen3.5-Plus} on the main experiment results. \textit{Bold} and \textit{underlined} results indicate the global optimum and the baseline optimum (except for Human), respectively.}
\label{tab:cost}
\end{table}

\subsection{Correlation Between Different Metrics}
\label{app:metcor}
To analyze the potential correlations among different metrics, we perform empirical statistical analysis on all the data of the main comparison experiment with DeepSeek-V4-Flash + GPT-4.1-mini.

The correlations between GT-based metrics and VLM-Judge metrics are low, and the mutual correlations within the VLM-Judge metrics are larger. Although the overly high correlation may indicate the influence of VLM's bias, the discretization and the too few levels of the scoring, and the inherent consistency of the experimental methods on these metrics could also be potential reasons. We further measure the correlation with \textbf{Kendall's Tau coefficient}~\citep{Sen1968EstimatesOT} in Table~\ref{tab:tau}, which indicates that the scoring mechanism is one of the factors leading to high correlation. Besides, the three dimensions target distinct pedagogical aspects (factual, structural, and cognitive), justifying keeping them separate for fine-grained diagnosis. The high \textit{quadratic weighted Kappa} with human evaluation also confirms that the bias has been effectively mitigated.

\begin{table}
\small
\setlength{\tabcolsep}{5pt}
\centering

\begin{tabular}{cccc}
\toprule
\textbf{Metrics} & \textbf{CA \& SQ} &  \textbf{CA \& PE}  &  \textbf{SQ \& PE}   \\
\midrule
\textbf{Pearson} & 0.894 & 0.806 & 0.896   \\
\textbf{Kendall's Tau} & 0.843 & 0.750 & 0.854  \\
\bottomrule
\end{tabular}
\caption{Comparison of Pearson and Kendall's Tau coefficients on VLM-Judge metrics.}
\label{tab:tau}
\end{table}

\begin{figure*}
    \centering
    \begin{subfigure}{0.5\linewidth}
        \centering
        \includegraphics[width=\linewidth]{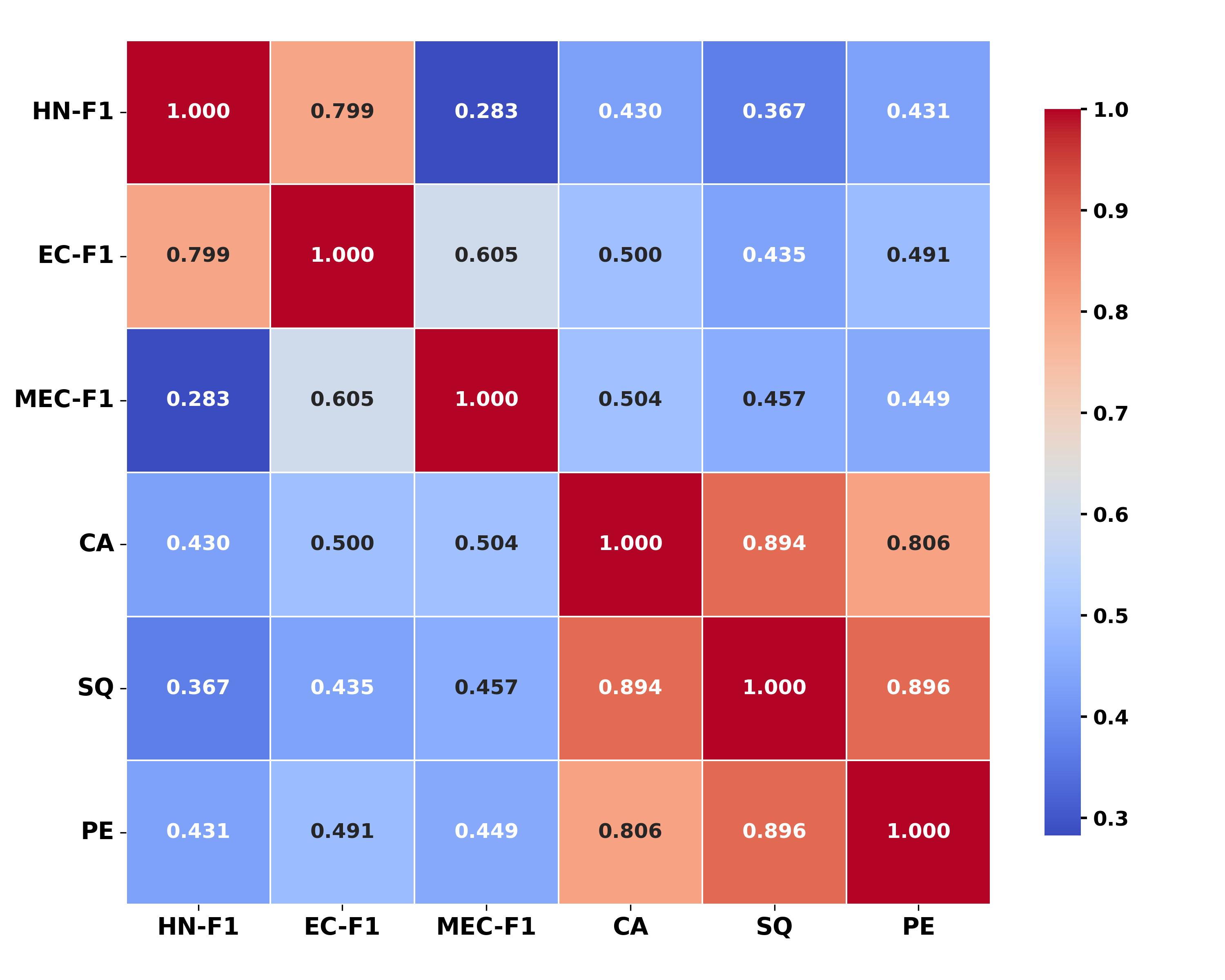}
        \caption{Pearson correlation matrix between GT-based metrics and VLM-Judge metrics.}
        \label{fig:cora}
    \end{subfigure}
    \hfill
    \begin{subfigure}{0.49\linewidth}
        \centering
        \includegraphics[width=\linewidth]{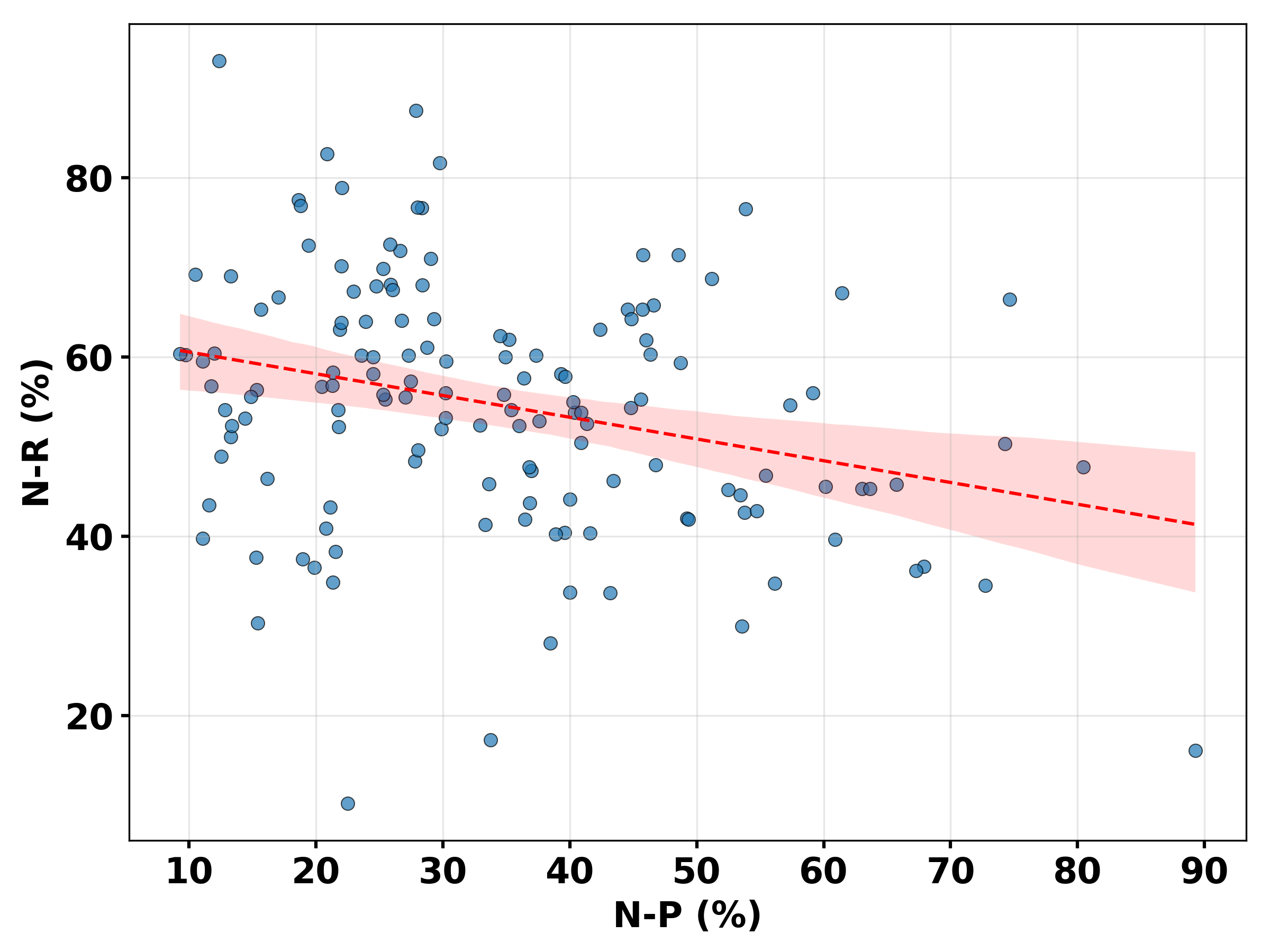}
        \caption{Correlation between N-R and N-P, with a Pearson correlation coefficient of -0.2913.}
        \label{fig:corb}
    \end{subfigure}
    \hfill
    \begin{subfigure}{0.33\linewidth}
        \centering
        \includegraphics[width=\linewidth]{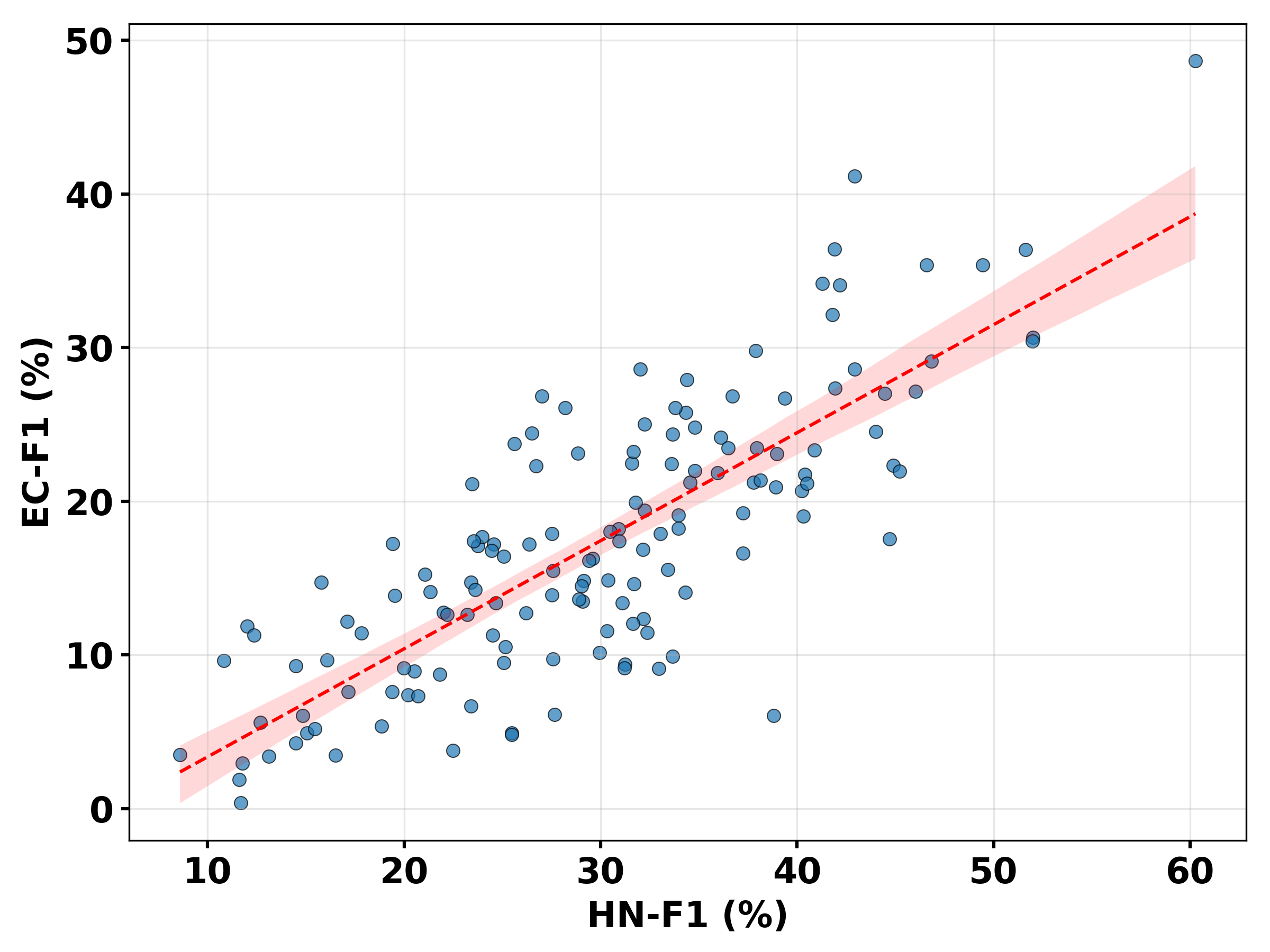}
        \caption{Correlation between EC-F1 and HN-F1.}
        \label{fig:corc}
    \end{subfigure}
    \hfill
    \begin{subfigure}{0.33\linewidth}
        \centering
        \includegraphics[width=\linewidth]{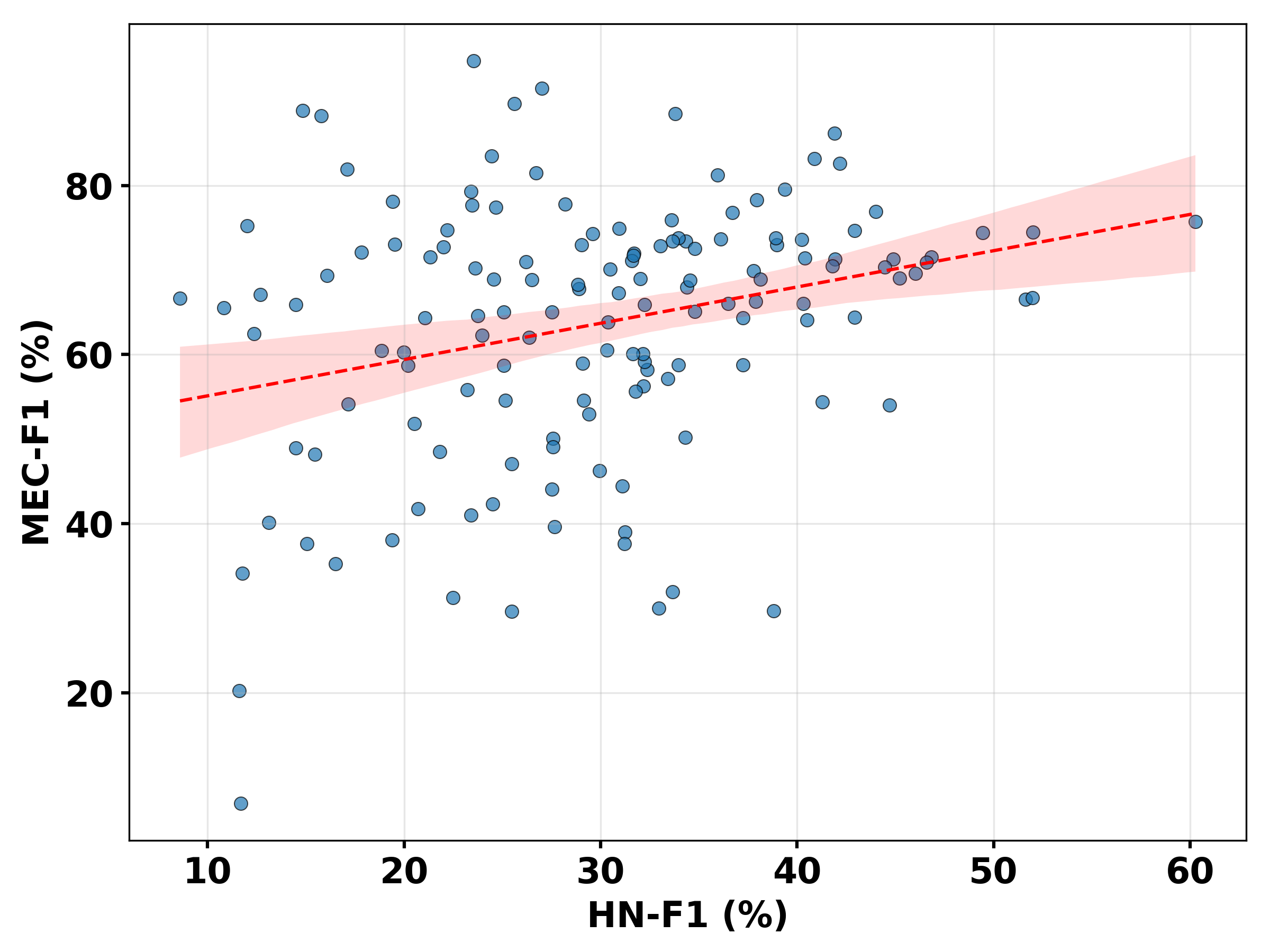}
        \caption{Correlation between MEC-F1 and HN-F1.}
        \label{fig:cord}
    \end{subfigure}
    \hfill
    \begin{subfigure}{0.32\linewidth}
        \centering
        \includegraphics[width=\linewidth]{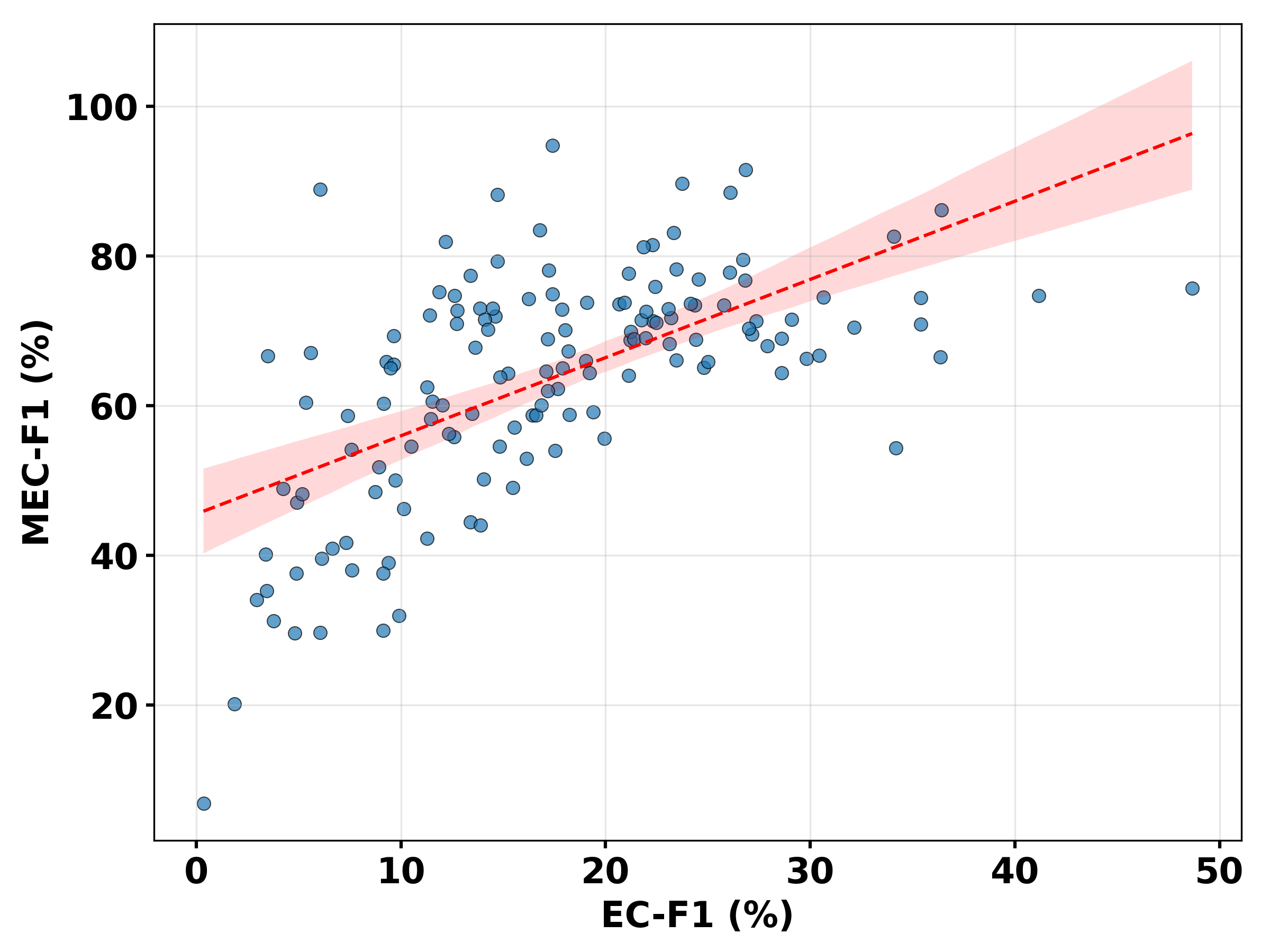}
        \caption{Correlation between MEC-F1 and EC-F1.}
        \label{fig:core}
    \end{subfigure}
    \caption{Visualization of correlation among main evaluation metrics, and scatter diagrams.}
    \label{fig:cor}
\end{figure*}

Among GT-based metrics, HN-F1 and MEC-F1 show the largest correlation but HN-F1 and MEC-F1 have the weakest. Specifically, HN-F1 and MEC-F1 have an obvious trade-off in the high-value interval of HN-F1. The result is consistent with previous experiments (especially in the ablation study): organizing hierarchies for fewer matching nodes naturally has lower difficulty and leads to higher MEC-F1. The trade-off makes MEC-F1 at risk of being \textbf{hacked}, but doesn't mean complete invalidity. MEC-F1 can still reflect the capacity of hierarchy identification more accurately than EC-F1 when the number of matching nodes is similar. The correlation between N-R and N-P is also visualized in Figure~\ref{fig:corb}; the trade-off is consistent with Appendix~\ref{app:gtdetail}. Besides, a single metric is easy to hack (e.g., some baselines achieve high MEC-F1 but have poor overall quality), but multi-facet metrics with low correlation increase the evaluation robustness.

\subsection{Analysis on GT-based Metrics}
\label{app:gt low ana}
Despite achieving the best performance among automated methods, the absolute GT-based metrics of AutoMindMap remain relatively low. For further analysis, we sample three courses from different categories, and invite new annotators to reconstruct mind maps without referencing the original ground truths. Evaluating these re-annotations against the original GT (Table~\ref{tab:human+human}) reveals substantial variability even among human experts, underscoring the inherent open-ended nature of the Slides2MindMap task. Additionally, the strict tuple-level matching rule may penalize semantically valid but structurally divergent hierarchies, further suppressing absolute scores. This analysis confirms that the comparison with ground truth is insufficient, justifying our multi-faceted evaluation framework.

However, the following conclusions are still tenable: (1)~Human re-annotations consistently outperform all automated methods and align with the results of VLM-Judge, confirming that GT-based metrics might be influenced by the task openness, but still retain discriminative power for knowledge faithfulness, particularly in the low-score regime. GT-based metrics also prevent other metrics from being hacked. (2) AutoMindMap substantially narrows the gap to the expert, demonstrating meaningful progress toward expert-level performance in this inherently open-ended task. This gap also confirms the inescapable challenge of the Slides2Mindmap task. There is still much room for improvement in future work.

\begin{table}
\small
\setlength{\tabcolsep}{2pt}
\centering

\begin{tabular}{cccc}
\toprule
\textbf{Methods} & \textbf{HN-F1} &  \textbf{EC-F1}  &  \textbf{MEC-F1}   \\
\midrule
\textbf{Direct}& 40.71 \textcolor{orange}{(-20.93)} & 17.54 \textcolor{orange}{(-37.19)} &  54.03 \textcolor{orange}{(-29.64)}\\ 
\textbf{AutoMindMap} & 51.64 \textcolor{blue}{(-10.00)} & 33.45 \textcolor{blue}{(-21.28)} & 66.30 \textcolor{blue}{(-17.37)}   \\
\textbf{Human} & 61.64 & 54.73 & 83.67  \\
\bottomrule
\end{tabular}
\caption{GT-based metrics of human re-annotation, compared with \textit{AutoMindMap} and \textit{Direct}, on the sampled courses. The differences between automated methods and human re-annotation are attached. The model setting is DeepSeek-V4-Flash + GPT-5.4-mini.}
\label{tab:human+human}
\end{table}

\section{More Structural Comparison}
\label{app:morestruct}
We employ more structural indicators to analyze the experimental results (with DeepSeek-V4-Flash + GPT-5.4-mini). These indicators represent the core structural characteristics of the generated mind maps but don't absolutely represent the quality of mind maps; thus, these extra structural comparisons only serve as soft guidelines and a quantified representation of the \textbf{SQ} metric for reference.

First, for N-R in Table~\ref{tab:gt_all_com}, we count the N-R of each layer's node in the ground truth (with the root node as layer-1, $L_1$) in Table~\ref{tab:layerrecall}. The N-R values of all methods decrease as the layers become deeper. The deeper nodes in the mind map tend to be less important~\citep{ausubel1960use}, which indicates that all the existing methods can achieve higher knowledge recall on more core concepts, and the Chunk-Merge method has the slowest N-R decline, which is caused by its mechanism that leads to many fine-grained concept extractions.

Grounded by the principle of cognitive balance and the criteria of VLM-Judge, to quantify the balance of branch depth and node distribution, we define a \textbf{branch imbalance coefficient} based on the complement of \textbf{normalized entropy}.
For a node $v$, let $\mathcal{E}(v)=\{u\mid (v,u)\in E\}$ be the set of direct child. Each $u\in\mathcal{E}(v)$ is described by a scalar indicator $\phi(u)$, including \textbf{sub-tree height} $h(u)=1+\max_{w\in V(u)}(d(w)-d(u))$ and \textbf{sub-tree size} $s(u)=|V(u)|$, where $V(u)$ is the node set of the sub-tree rooted at $u$.
After normalizing the children values into $p_u=\phi(u)/\sum_{w\in\mathcal{E}(v)}\phi(w)$, the local imbalance at $v$ is

\begin{equation}
\eta_\phi(v) = 1 - \frac{-\sum_{u\in\mathcal{E}(v)} p_u \log p_u}{\log |\mathcal{E}(v)|},
\label{eq:local_imbalance}
\end{equation}
to aggregate across the hierarchy, we fix a depth limit $L$ (the number of levels counted from the root).
Recall that the root $r$ has $d(r)=0$, so the node set at level $k$ is $V_k = \{v\in V \mid d(v)=k\}$.
The mean imbalance at level $k$ is $\bar{\eta}_\phi^{(k)} = \frac{1}{|V_k|}\sum_{v\in V_k}\eta_\phi(v)$.
The global branch imbalance coefficient is the average of the first $L$ level‑wise means:
\begin{equation}
\Psi_\phi^{(L)} = \frac{1}{L}\sum_{k=0}^{L-1} \bar{\eta}_\phi^{(k)}.
\label{eq:global_imbalance}
\end{equation}

Plugging in $\phi=h$ yields the \textbf{branch depth imbalance} $\Psi_{\text{depth}}^{(L)}$; $\phi=s$ yields the \textbf{branch size imbalance} $\Psi_{\text{size}}^{(L)}$. $\Psi_\phi^{(L)}$ captures structural skew that average depth/child fan-out alone cannot reveal.

The two coefficients, as well as the node size, cross-link ratio, and the average leaf-node layers in each mind map, are listed in Table~\ref{tab:morestruct}. Regarding human-annotated data as a benchmark, the balance of all methods is in a reasonable interval (as branches cannot be fully balanced due to the heterogeneity of knowledge organization), while Direct, Highlights-only, and AutoMindMap show stronger balance as a whole. The above three methods also produce significantly more concise node scales that are closer to human data, and AutoMindMap is comparatively worse. Although the mind map directly generated by LLMs may lack knowledge faithfulness, the structural \textbf{balance} and \textbf{scale} are closer to human preferences. However, in terms of cross-link ratio and layer number, the preferences of these two methods are farther from human. AutoMindMap performs more conservatively in cross-link ratio, and has the closest performance to human in hierarchical depth.

\begin{figure*}
    \centering
    \includegraphics[width=1\linewidth]{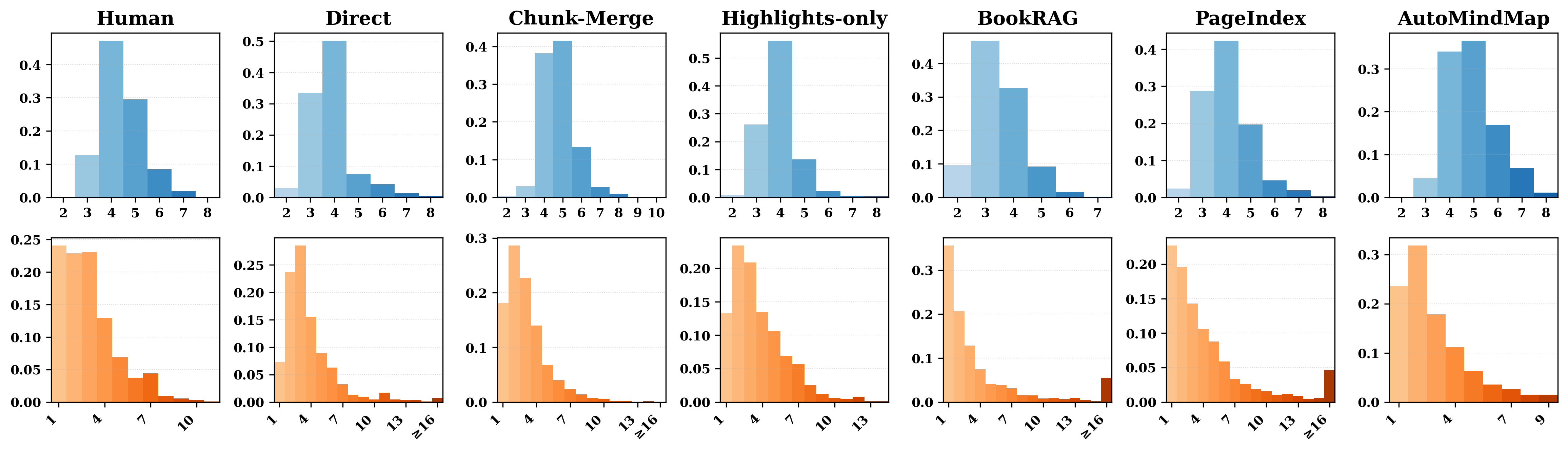}
    \caption{Distributions of leaf-node layers (top) and direct child-node fan-out (bottom) in mind maps generated by different methods. The vertical axis represents the proportion.}
    \label{fig:placeholder}
\end{figure*}

As average counting cannot reveal extreme values and overall distribution, we further count the distribution of leaf-node depth and child-node fan-out in Figure~\ref{fig:placeholder}. More intuitively, AutoMindMap shows more similar distributions to Human in both dimensions, while avoiding overly deep and overly many child-node fan-outs. Comparatively, many baselines generate an excess of child nodes (e.g., BookRAG and PageIndex) or depth (e.g., Chunk-Merge), or overly flat layers (e.g., Direct), which indicate poor structures. The result is consistent with the evaluation involving structure quality in the main experiment.

\begin{table}
\small
\setlength{\tabcolsep}{5pt}
\centering

\begin{tabular}{cccccc}
\toprule
\textbf{Methods} & \textbf{Overall} & \textbf{$L_2$} & \textbf{$L_3$}& \textbf{$L_4$} & \textbf{$\geq L_5$} \\
\midrule
 \textbf{Direct} & 49.96 & 63.12 & 58.59 & 48.60 & 42.97 \\

 \textbf{Chunk-Merge} & 59.64 & 68.59 & 64.01 & \textbf{59.38} & \textbf{53.73} \\

 \textbf{Highlights-only} & 30.16 & 50.93 & 41.72 & 26.63 & 21.44 \\

 \textbf{BookRAG} & 52.21 & 72.28 & 58.26 & 52.57 & 39.77 \\

 \textbf{PageIndex} & 60.08 & 69.98 & \textbf{68.22} & 58.60 & 50.07 \\
\midrule
 \textbf{AutoMindMap} & \textbf{61.21}& \textbf{72.31} & 58.97 & 54.35 & 40.27 \\

\bottomrule
\end{tabular}
\caption{N-R for different layers.}
\label{tab:layerrecall}
\end{table}

\begin{table*}
\small
\setlength{\tabcolsep}{6pt}
\centering

\label{tab:modality_categories}
\begin{tabular}{l c c c c c}
\toprule
\textbf{Methods} & \textbf{$\Psi_{\text{depth}}^{(2)}$}  & \textbf{$\Psi_{\text{size}}^{(2)}$} & \textbf{Size} & \textbf{Cross-link Ratio} & \textbf{Layers} \\
\midrule
\textbf{Human} & 0.71 & 12.17 & 121.38 & 9.26 & 4.40 \\

\midrule

\textbf{Direct} & 0.64 & 7.16 & 133.13 & 27.00 & 3.82 \\
\textbf{Chunk-Merge} & 1.07 & 11.80 & 295.08 & 10.00 & 4.77 \\
\textbf{Highlights-only} & 0.91 & 9.42 & 130.67 & 18.26 & 3.93 \\
\textbf{BookRAG} & 1.08 & 15.86 & 294.33 & 12.51 & 3.47 \\
\textbf{PageIndex} & 0.80 & 11.84 & 376.63 & 14.00 & 4.02 \\
\midrule
\textbf{AutoMindMap} & 0.91 & 10.14 & 185.75 & 5.53 & 4.33 \\
\bottomrule
\end{tabular}
\caption{More structural characteristics analysis. We set $L=2$. All the indicators are averaged by courses. Branch imbalance coefficients and cross-link ratio are scaled to $0\%$-$100\%$.}
\label{tab:morestruct}
\end{table*}

\section{Mind Map Visualization Case Study and Failure Analysis}
\label{app:casestudy}
We select a course named "Big Data Analytics" as a case and render the results of each method, with the help of Xmind\footnote{\url{https://xmind.cn/online-map/}}. This course has a total of 12 decks and 449 slide pages. The visualization results of human annotation, baseline methods, and AutoMindMap are shown in Figure~\ref{fig:com_vis}, and the results of different ablation study settings are shown in Figure~\ref{fig:abl_vis}.

\subsection{Analysis for Main Comparison Experiments}
The result of AutoMindMap is mostly similar to human annotation, having a clear hierarchical structure, accurate concept description and proper cross-links. The overall map is concise and clear with prominent focuses, which helps to reduce the cognitive load of learners. But shortcomings still exist in these two results. Although human-annotated data maintains good knowledge faithfulness, it is poor in branch balance, which indicates that ground truth is still imperfect, making multifaceted evaluations necessary. AutoMindMap also neglects some key concepts, especially in deeper layers.

The Direct method shows excellent overall structure and logic, but still has too many level-2 branches, the omission of some key concepts, and fewer layers, which are limitations of end-to-end long document understanding. Providing only important elements also weakens the understanding of the overall knowledge hierarchy, which leads to the disordered structure and logic of the Highlights-only method. Chunk-Merge, BookRAG, and PageIndex show even worse results. These three methods pay much attention to the local knowledge or mere retrieval-oriented document hierarchy, ignoring the coherence and progressiveness of the overall knowledge logic, resulting in excessive density of child nodes fan-out and cross-links, excessive flattening of the level, and a large number of factual errors, ambiguity, and noise in concept representation. For example, many concepts have "exercise" as their common child in Figure~\ref{fig:pageindex}, and lots of titles are directly used as node names without rephrasing.

The above baseline methods may be of high quality in the face of clearly structured textbooks, and BookRAG and PageIndex are efficient in building indexes for RAG systems, but all the above baselines perform poorly in Slides2MindMap, which presents the uniqueness of AutoMindMap: slide-specific document structure adaptation and the global-local trade-off of mind map generation.

\subsection{Analysis for Ablation Study}
\begin{figure*}
    \centering   
    \hfill
    \begin{subfigure}{0.49\linewidth}
        \centering
        \includegraphics[width=\linewidth]{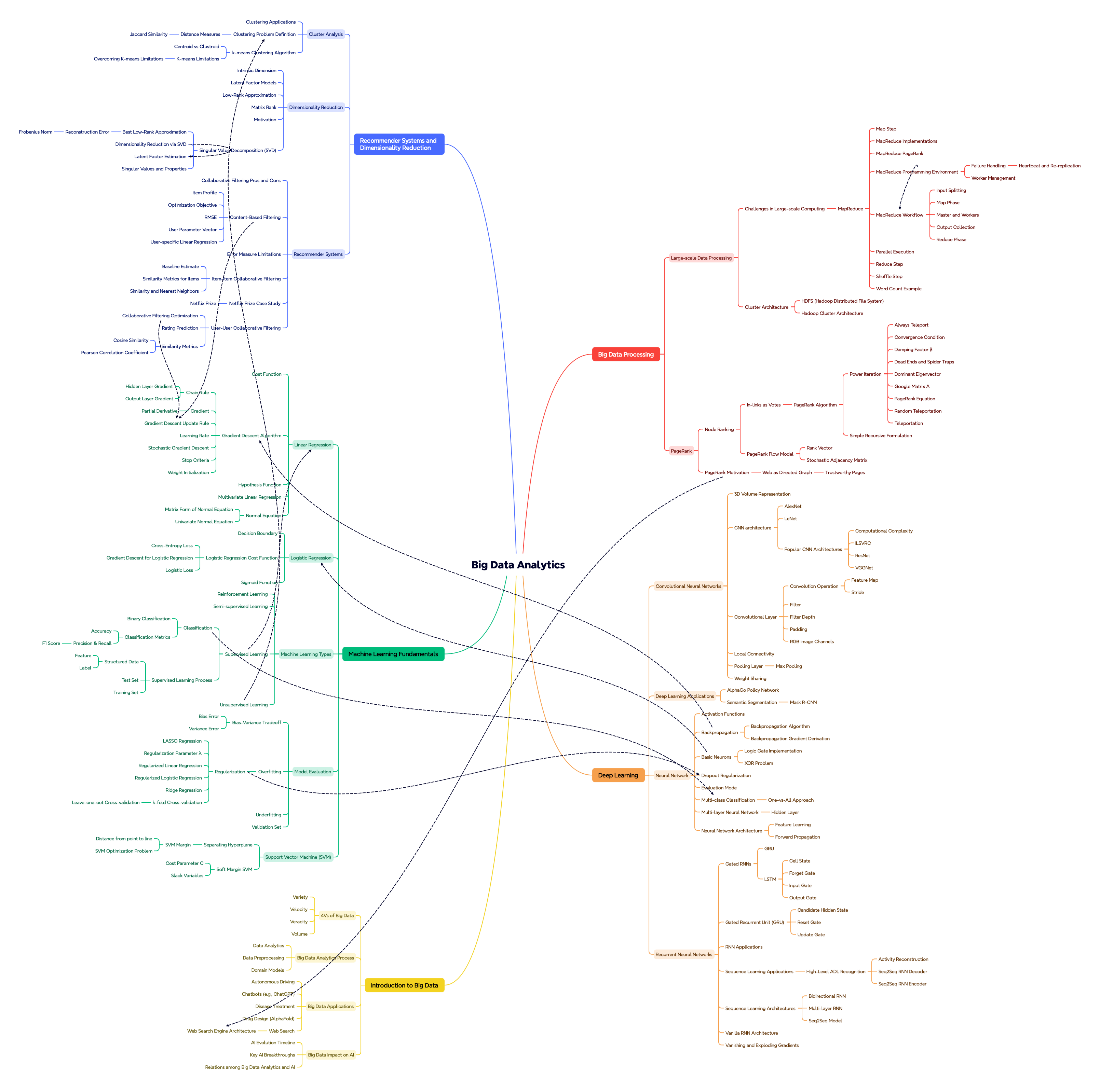}
        \caption{AutoMindMap w/o refinement.}
        \label{fig:histograms}
    \end{subfigure}
    \hfill
    \begin{subfigure}{0.49\linewidth}
        \centering
        \includegraphics[width=\linewidth]{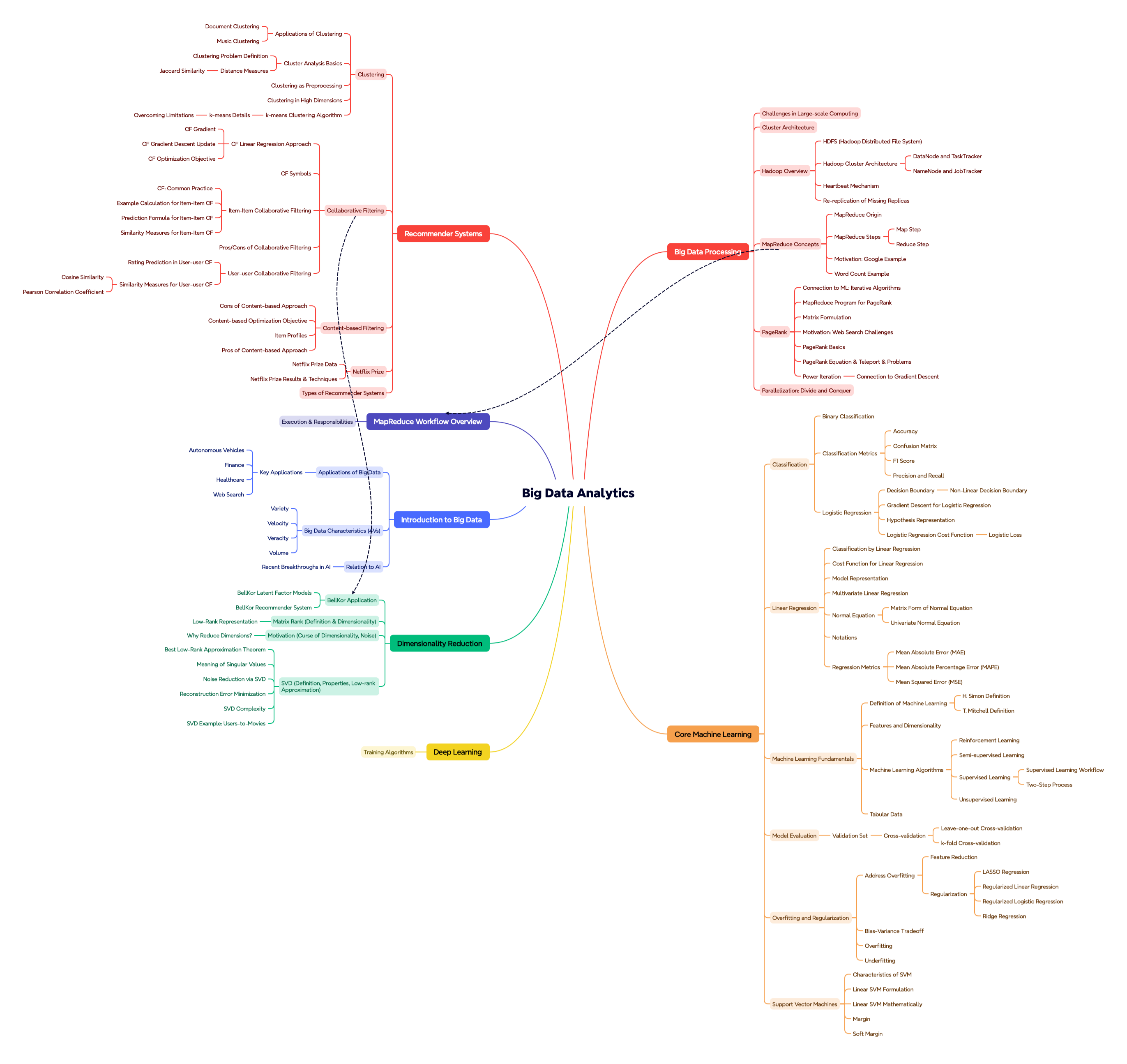}
        \caption{AutoMindMap w/o summary.}
        \label{fig:histograms}
    \end{subfigure}
    
     \hfill
         \begin{subfigure}{0.49\linewidth}
        \centering
        \includegraphics[width=\linewidth]{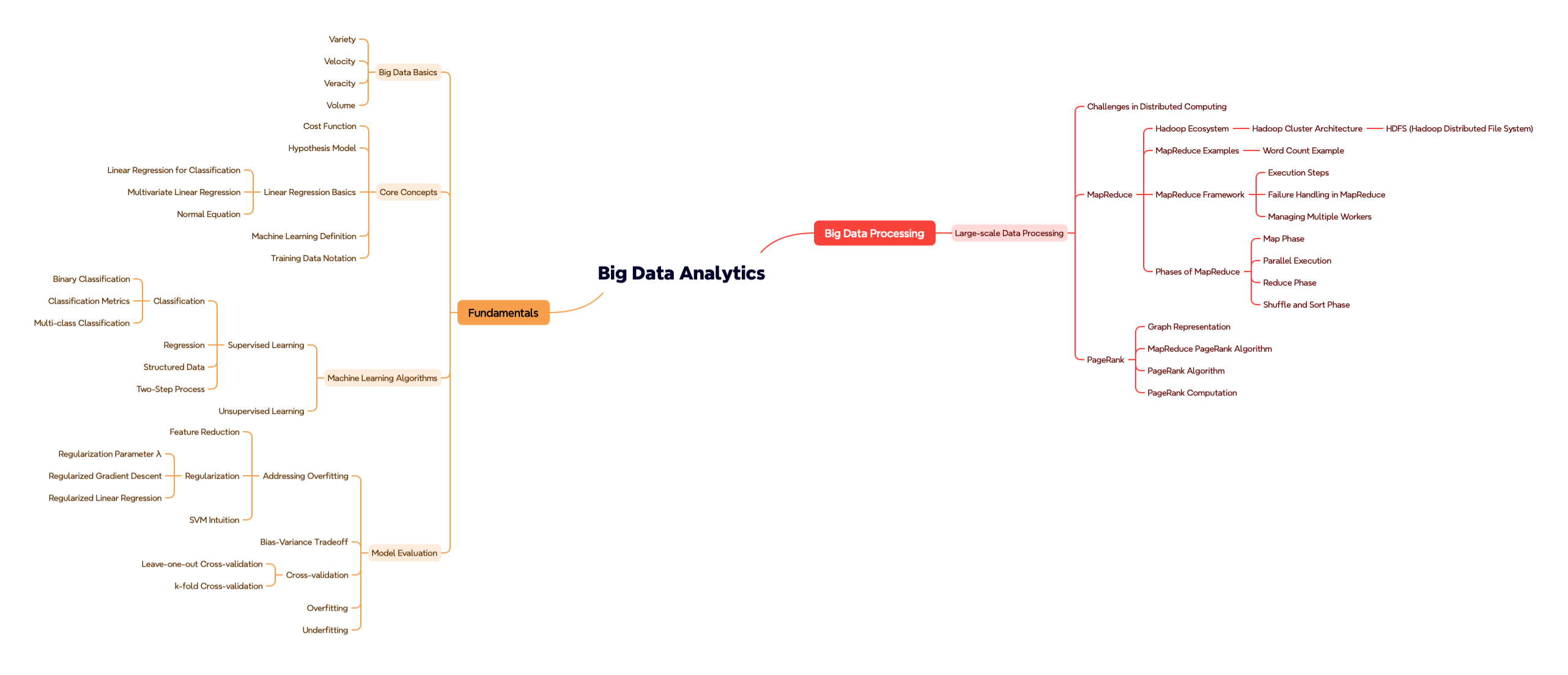}
        \caption{AutoMindMap w/o vision.}
        \label{fig:histograms}
    \end{subfigure}
    \hfill
    \begin{subfigure}{0.49\linewidth}
        \centering
        \includegraphics[width=\linewidth]{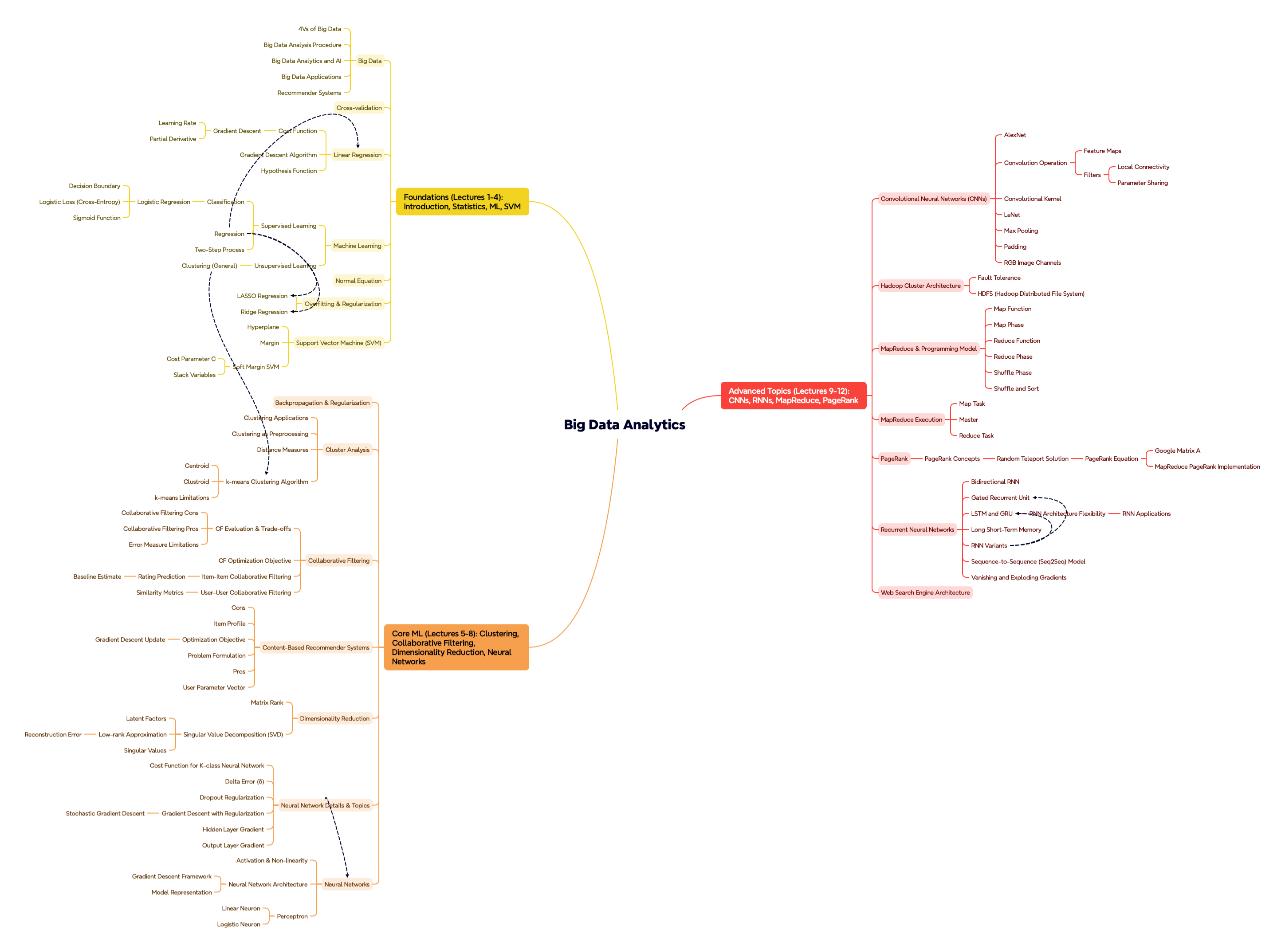}
        \caption{AutoMindMap w/o skeleton.}
        \label{fig:histograms}
    \end{subfigure}
     \hfill
    \begin{subfigure}{0.49\linewidth}
        \centering
        \includegraphics[width=\linewidth]{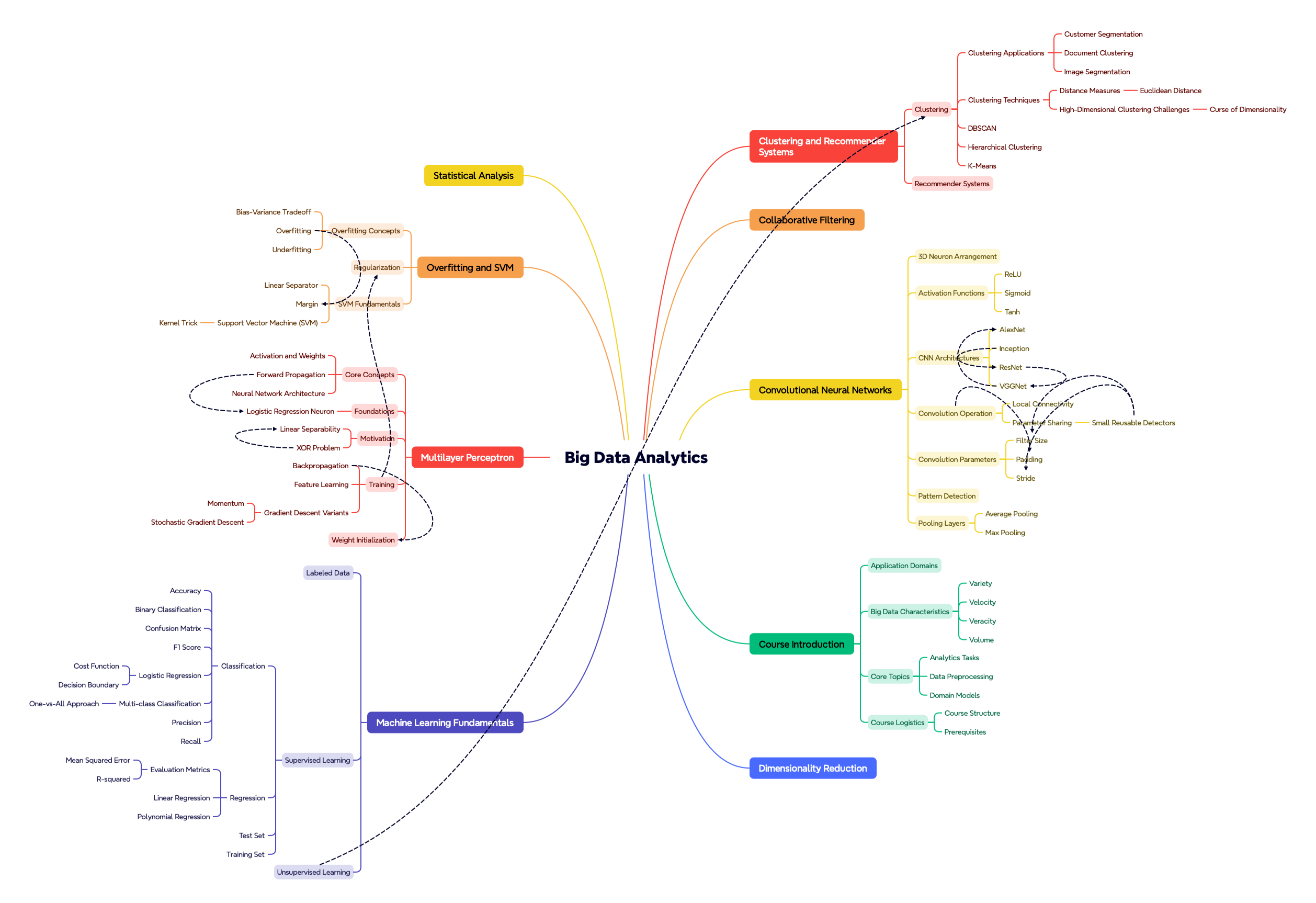}
        \caption{AutoMindMap w/o tools.}
        \label{fig:histograms}
    \end{subfigure}
    \caption{Visualization of the mind maps generated in the ablation study.}
    \label{fig:abl_vis}
\end{figure*}

Figure~\ref{fig:abl_vis} shows that the \textit{refinement mechanism} of AutoMindMap can significantly rebalance branches and prune fine-grained nodes to keep conciseness, and avoid mutual deconstruction of local-global structures. The result without the \textit{visual information} in refinement may lead to a misunderstanding of the mind map structures, resulting in inappropriate modification tool-use and collapsed structure. The result of \textit{w/o summary} shows logic breaks in some branches, and neglects key concept hierarchy extraction due to loss of context-associative bridge. AutoMindMap \textit{w/o skeleton} shows unnatural hierarchical progression, indicating the necessity to understand the overall content and plan the preliminary structure before the formal construction. AutoMindMap \textit{w/o tools} leads to confusing and collapsed local structures, which shows that the tool can promote structural constraints and understanding, and reduce the hallucinations generated by the pure-text way.

\begin{figure*}[htbp]
    \centering
    \begin{subfigure}{0.49\linewidth}
        \centering
        \includegraphics[width=\linewidth]{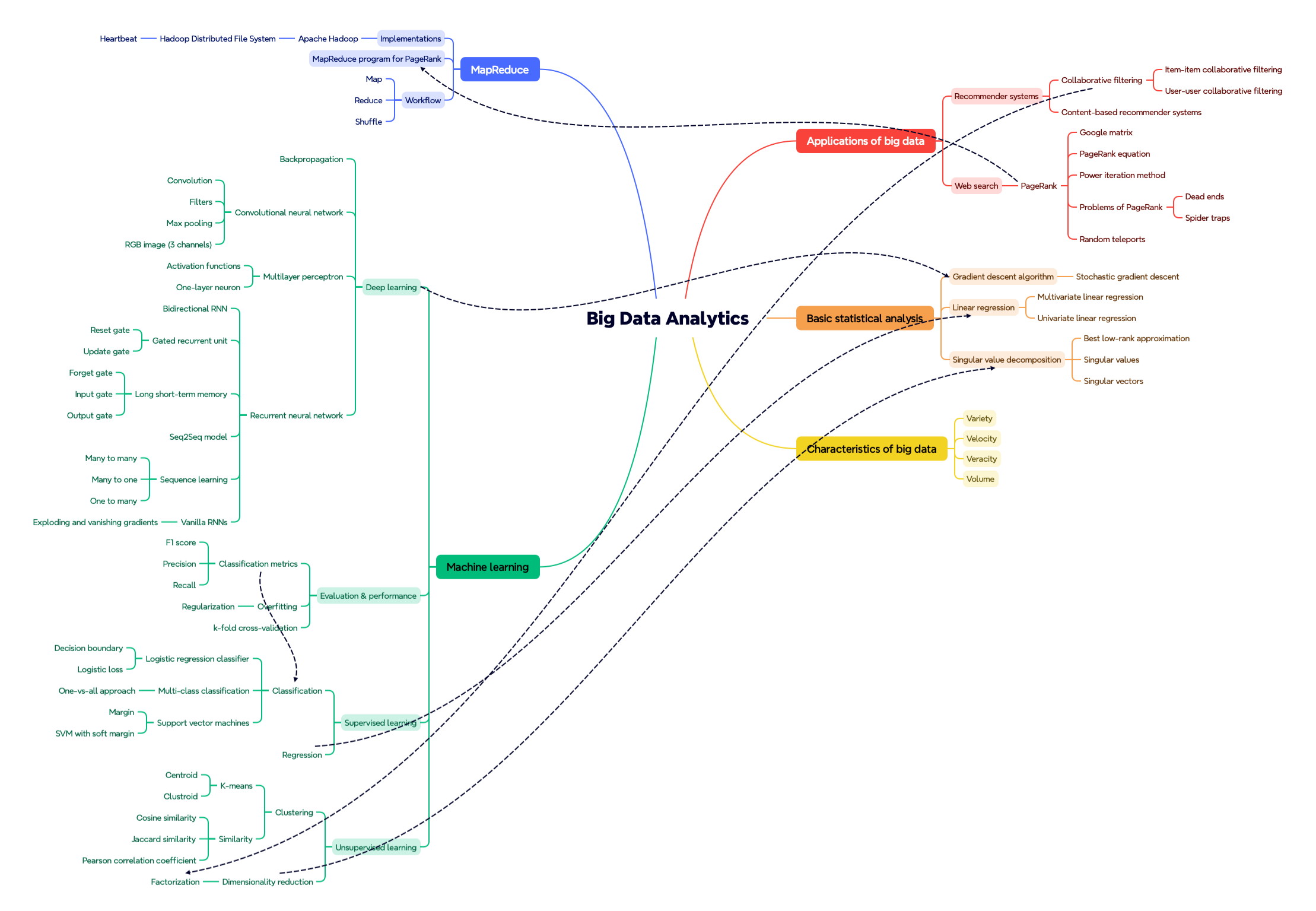}
        \caption{Human.}
        \label{fig:human}
    \end{subfigure}
    \hfill
    \begin{subfigure}{0.49\linewidth}
        \centering
        \includegraphics[width=\linewidth]{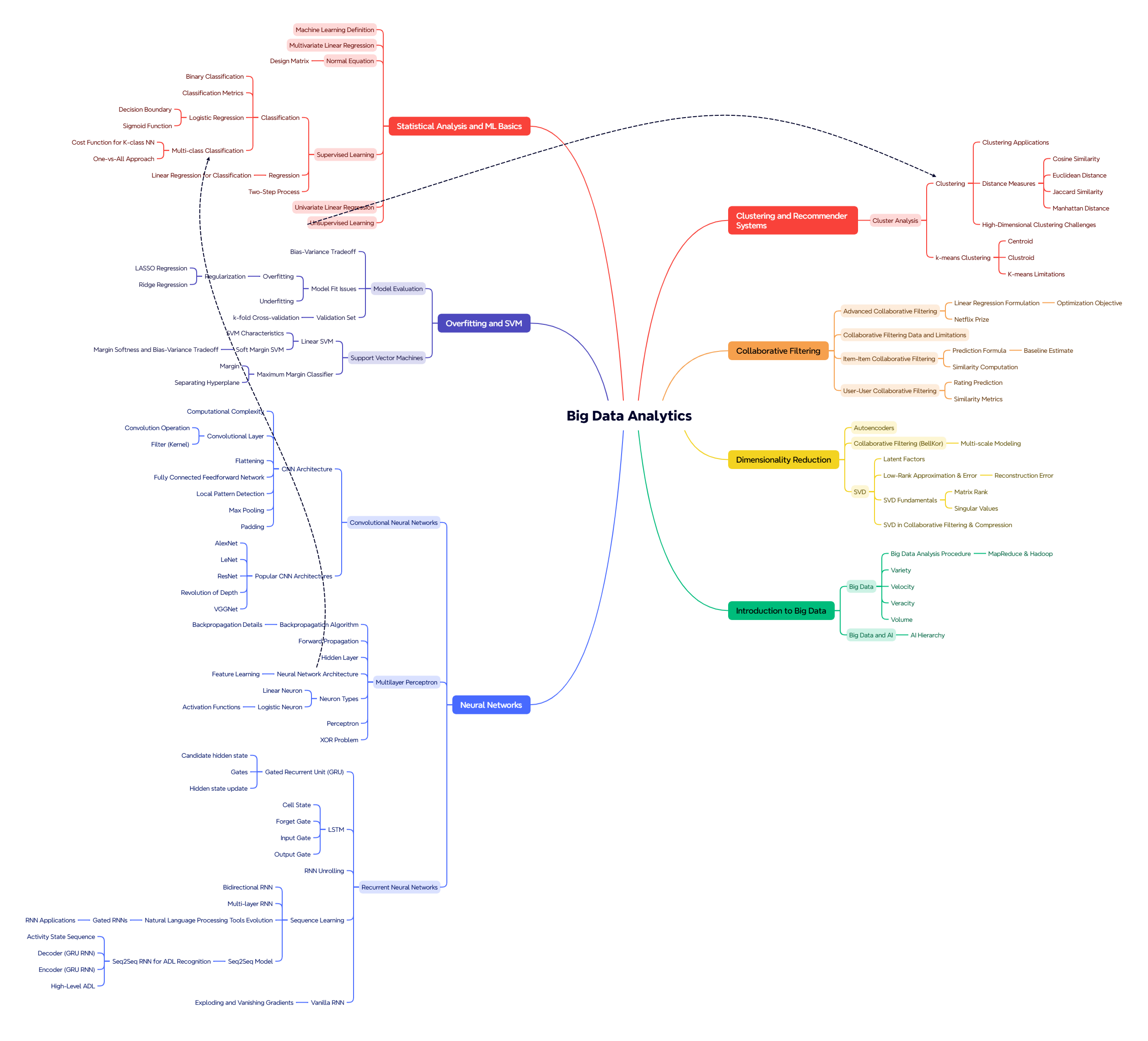}
        \caption{AutoMindMap.}
        \label{fig:automm}
    \end{subfigure}

    \medskip
    \begin{subfigure}{0.49\linewidth}
        \centering
        \includegraphics[width=\linewidth]{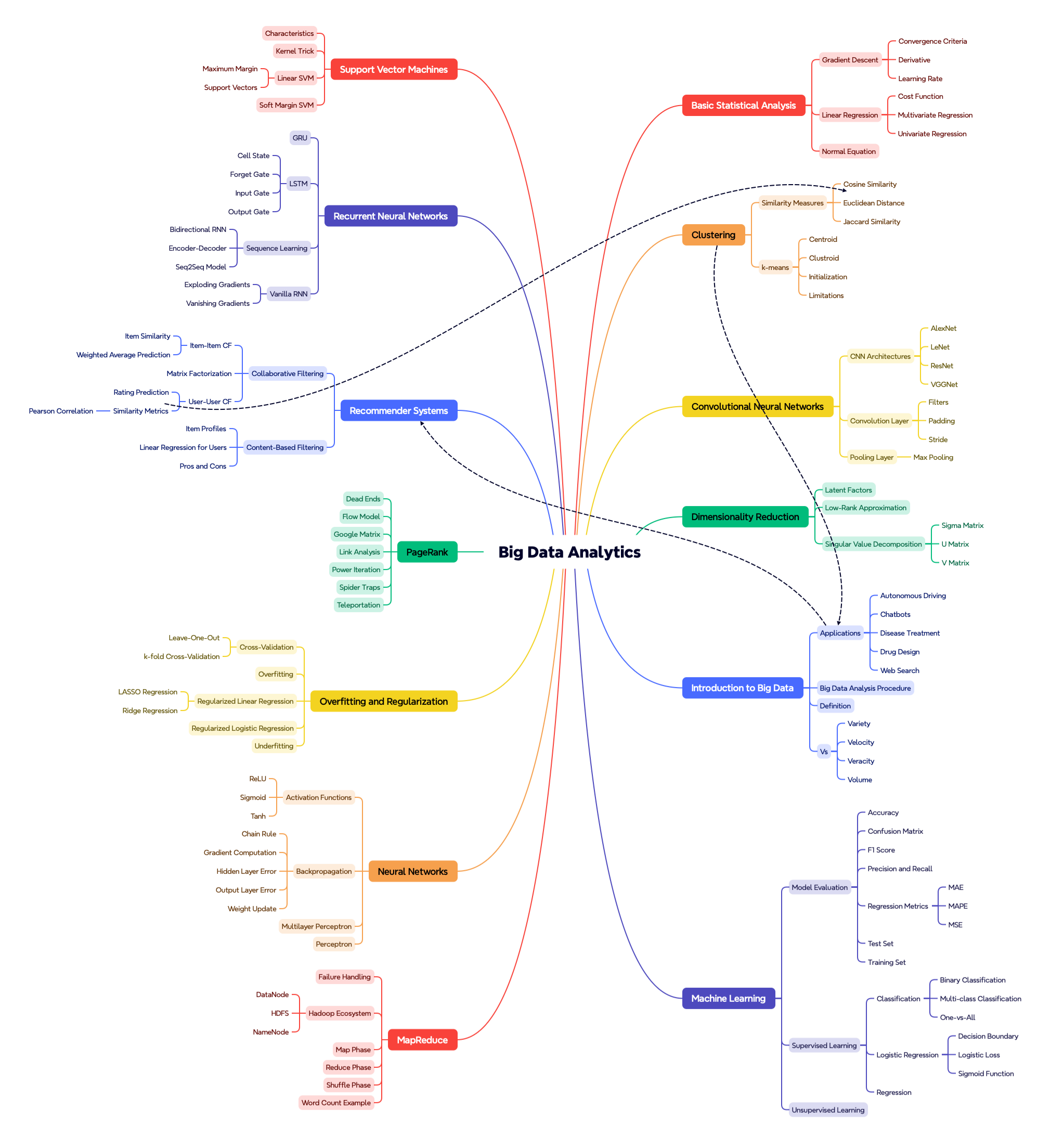}
        \caption{Direct.}
        \label{fig:direct}
    \end{subfigure}
    \hfill
    \begin{subfigure}{0.49\linewidth}
        \centering
        \includegraphics[width=\linewidth]{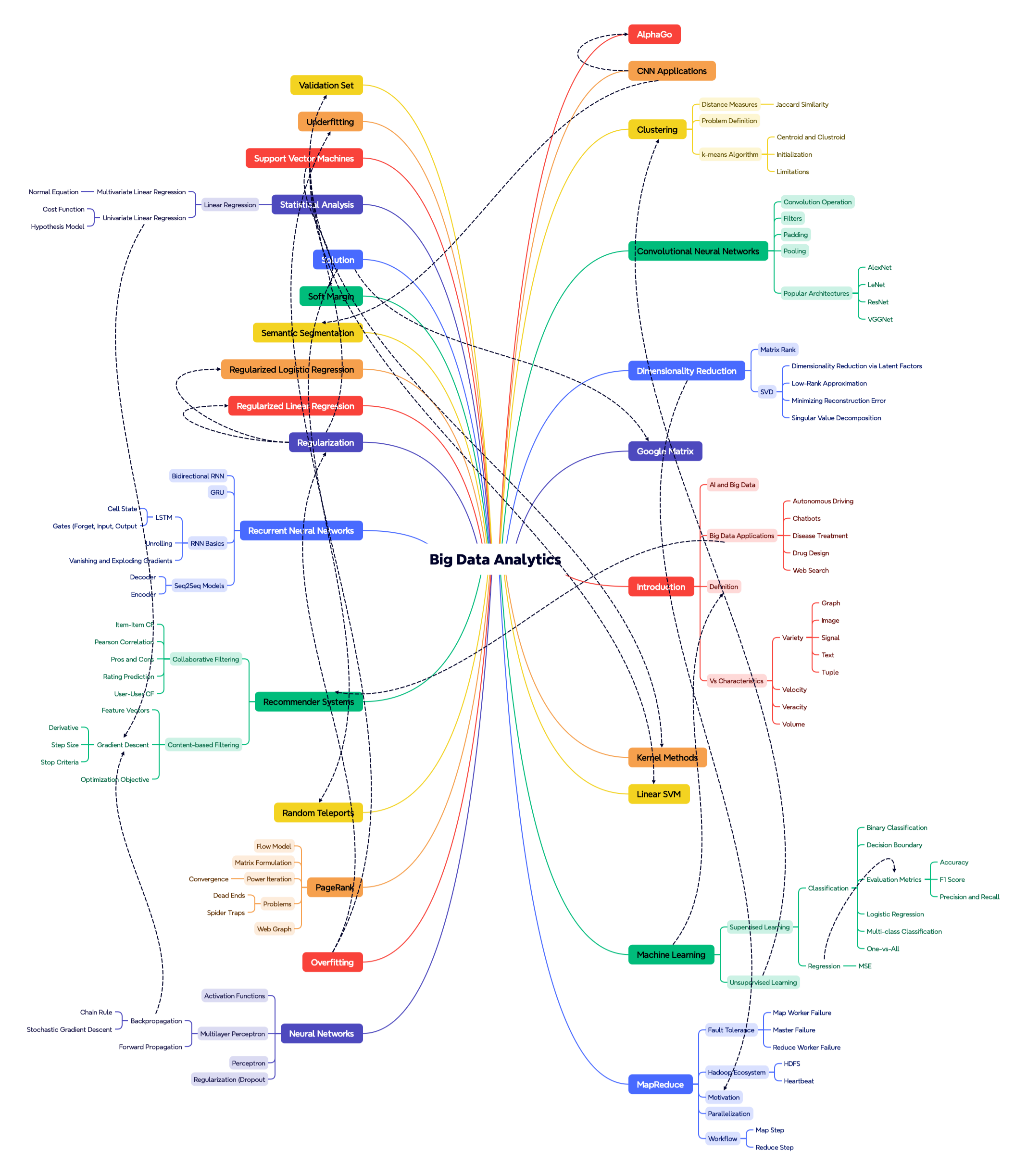}
        \caption{Highlights-only.}
        \label{fig:highlights}
    \end{subfigure}

    \caption{Visualization of the mind maps generated by different baselines.}
    \label{fig:com_vis}
\end{figure*}


\begin{figure*}[htbp]
    \centering
    \begin{subfigure}{0.35\linewidth}
        \centering
        \includegraphics[width=\linewidth]{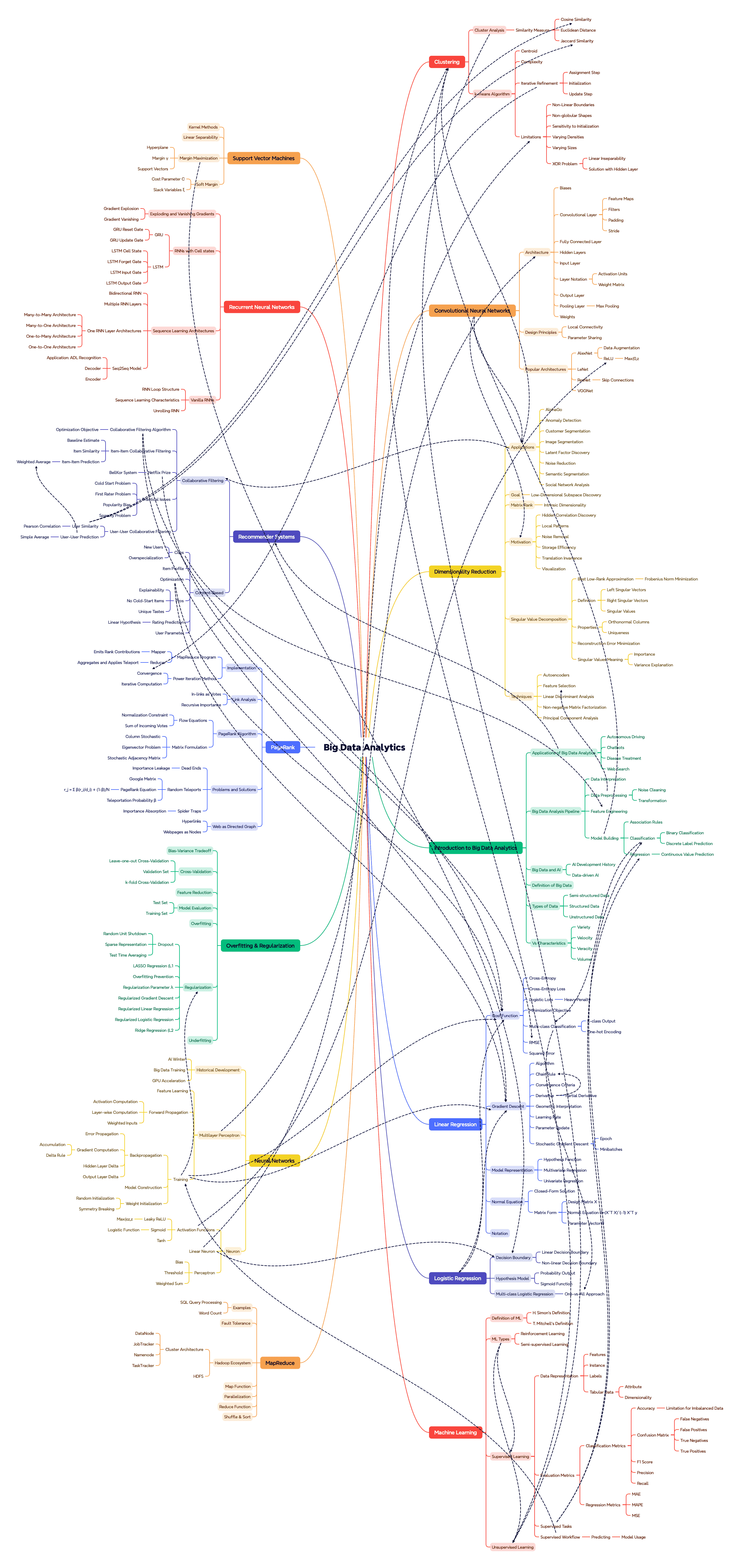}
        \caption{Chunk-Merge.}
        \label{fig:chunkmerge}
    \end{subfigure}
    \hfill
    \begin{subfigure}{0.27\linewidth}
        \centering
        \includegraphics[width=\linewidth]{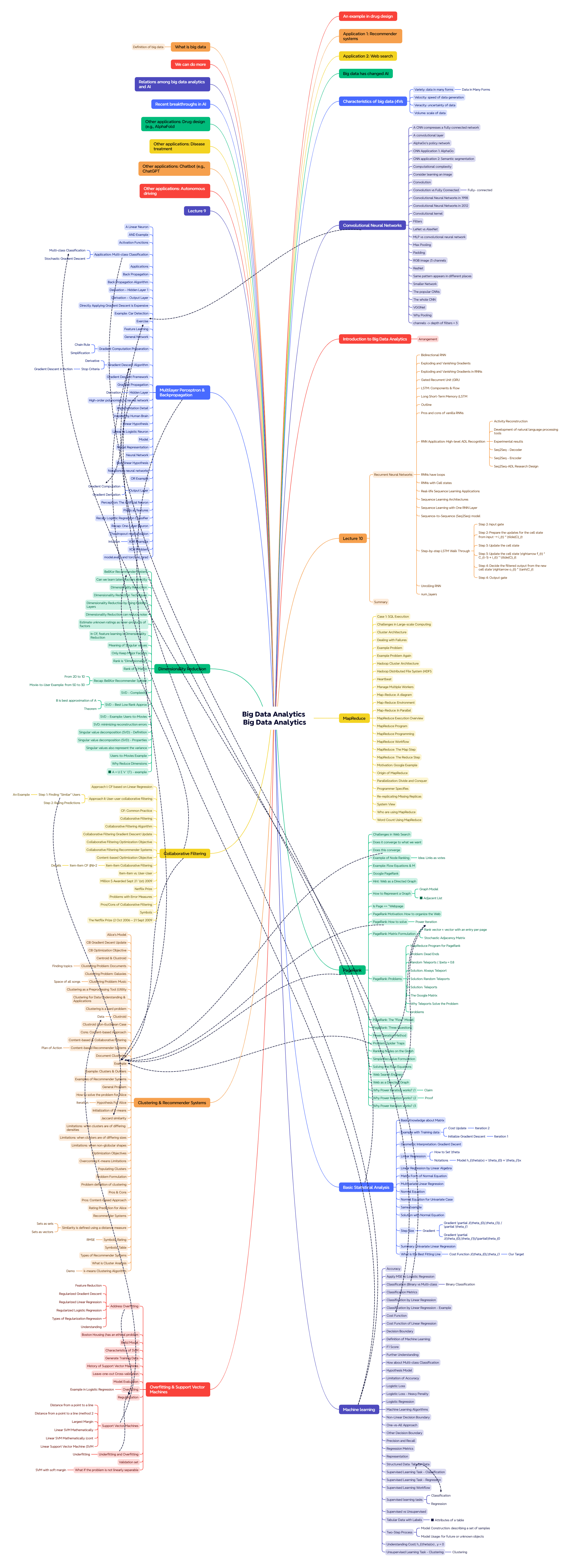}
        \caption{BookRAG.}
        \label{fig:bookrag}
    \end{subfigure}
    \hfill
    \begin{subfigure}{0.35\linewidth}
        \centering
        \includegraphics[width=\linewidth]{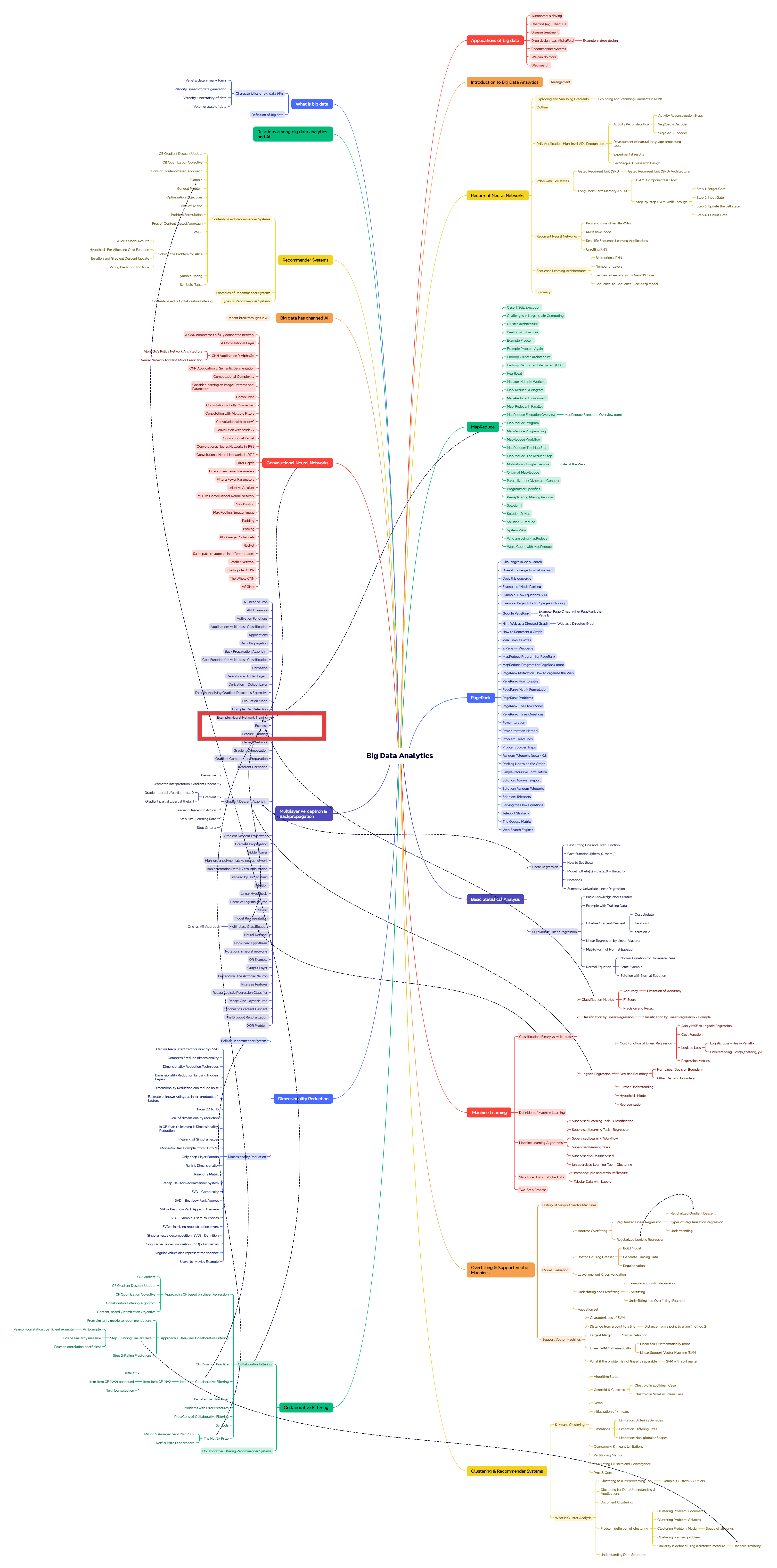}
        \caption{PageIndex.}
        \label{fig:pageindex}
    \end{subfigure}

    \caption{(Continued) Visualization of the mind maps generated by different baselines.}
\end{figure*}

\section{Tool Definitions for Mind Map Modification}
\label{app:tool}
The descriptions of all atomic tools for mind map modification are listed in Table~\ref{tab:tool}. Specs for registered tools are appended to the prompts in AutoMindMap.

\begin{table*}
\small
\setlength{\tabcolsep}{5pt}
\centering

\label{tab:tools}
\begin{tabularx}{\textwidth}{@{} l >{\raggedright\arraybackslash}p{5.5cm} X @{}}
\toprule
\textbf{Tool Name} & \textbf{Parameters} & \textbf{Description} \\
\midrule
\texttt{rename\_node}
  & \makecell[tl]{\texttt{old\_name} \textit{(string, required)} \\ \texttt{new\_name} \textit{(string, required)}}
  & Rename an existing node. \\
\midrule
\texttt{add\_node}
  & \makecell[tl]{\texttt{parent\_name} \textit{(string, required)} \\ \texttt{new\_node\_name} \textit{(string, required)} \\ \texttt{new\_node\_description} \textit{(string, optional)}}
  & Add a new node under an existing parent. A brief description incorporating the hierarchical context is encouraged. \\
\midrule
\texttt{add\_edge}
  & \makecell[tl]{\texttt{parent\_name} \textit{(string, required)} \\ \texttt{child\_name} \textit{(string, required)}}
  & Add a new directed edge between two existing nodes.\\
\midrule
\texttt{delete\_edge}
  & \makecell[tl]{\texttt{parent\_name} \textit{(string, required)} \\ \texttt{child\_name} \textit{(string, required)}}
  & Delete an existing directed edge without removing the associated nodes. Can be combined with \texttt{add\_edge} to reorganize local structure; carries the risk of creating isolated nodes. \\
\midrule
\texttt{delete\_node}
  & \makecell[tl]{\texttt{node\_name} \textit{(string, required)}}
  & Delete an existing node along with its incident edges. Avoid use unless the target is a leaf node or other tools are used together, to prevent orphaned child nodes. \\
\midrule
\texttt{merge\_nodes}
  & \makecell[tl]{\texttt{node\_names} \textit{(array<string>, required)} \\ \texttt{merged\_name} \textit{(string, required)}}
  & Merge \(N\) existing nodes into a single node while preserving all original edges. \\
\bottomrule
\end{tabularx}
\caption{Tool specifications for mind map operations.}
\label{tab:tool}
\end{table*}

\section{Important Cognitive Science Theories for Reference}
\label{app:theory}

\noindent \textbf{Relationship Between Mind Maps and Concept Maps}~\citep{davies2011concept}. Mind maps and concept maps present similar features; both are widely used in educational scenarios to promote knowledge understanding and organize concepts into divergent structures around a specific center. However, concept maps have stricter and more complex relational constraints than mind maps (i.e., the various relations between concepts should be identified explicitly), which contain more detailed knowledge association, but may cause higher cognitive costs, while mind maps are more flexible in organization and more obvious in hierarchy, facilitating more efficient knowledge overview. Under certain conditions, mind maps can be compatible with the features of concept maps. In our work, although the constructed mind maps have lots of features of concept maps, we focus on broader hierarchies instead of more concrete relations, which is still closer to the category of mind maps.

\noindent \textbf{Structure Building Framework (SBF)}~\citep{gernsbacher2013language}. The theory characterizes human discourse comprehension as the construction of coherent mental representations through three core processes: laying a foundation, mapping, and shifting. The human learners first lay a foundation using initial input segments, then map subsequent coherent information onto the developing structure. When incoming information is less coherent or related, they shift to initiate a new substructure, resulting in multi-branching mental representations. This framework directly inspires the design of AutoMindMap.

\noindent \textbf{Working Memory Capacity Theory}~\citep{miller1956magical}. Miller's classic formulation posits that human working memory can hold approximately $7\pm2$ information chunks at a time. This bounded capacity implies that knowledge structures with excessive child nodes under one parent or overly deep hierarchies impose cognitive overload on learners. The principle is adopted by previous mind map rubrics. In our evaluation, the \textbf{Structural Load Factor} directly operationalizes this constraint, and the \textbf{Structure Quality} explicitly discourages overloaded structures.

\noindent \textbf{Advance Organizer Theory}~\citep{ausubel1960use}. This theory emphasizes that meaningful learning occurs when new knowledge is integrated into existing cognitive structures via advance organizers, which are introductory materials presented at a higher level of abstraction. A key principle is \textbf{progressive differentiation}: knowledge should be presented from general to specific, building a stable scaffold before introducing details. Mind maps serve as visual advance organizers; in \textbf{Structure Quality}, \textbf{Pedagogical Efficiency}, and \textbf{Structural Load Factor}, we favor maps that present core topics prominently and have progressive hierarchy instead of a flat structure. The \textbf{Skeleton Laying} stage in AutoMindMap is also inspired by the theory.

\noindent \textbf{Principle of Cognitive Balance.} Drawing from information visualization~\citep{175815} and cognitive load research~\citep{Sweller2019CognitiveAA}, this principle holds that highly unbalanced hierarchical structures reduce navigation efficiency and impose uneven cognitive burdens. Our evaluation incorporates this principle in the \textbf{Structure Quality} dimension and extra structural analysis in Appendix~\ref{app:morestruct}. Notably, perfect balance is neither achievable nor desirable due to the intrinsic heterogeneity of knowledge, so our framework treats balance as a soft guideline allowing reasonable structural flexibility.

\noindent \textbf{Cognitive Flexibility Theory}~\citep{Spiro1988CognitiveFT}. Advanced knowledge acquisition in ill-structured domains requires learners to revisit concepts from multiple perspectives and construct alternative linkages among knowledge elements. Cross-links in mind maps serve as structural analogues to this principle, enabling learners to perceive interconnections beyond strict hierarchies. Our evaluation thus rewards \textit{"reasonable cross-links across branches that promote concept association"} in the \textbf{Pedagogical Efficiency} criteria, while cautioning against excessive density that would instead induce confusion.

\noindent \textbf{Memory-based Text Processing Theory}~\citep{o1998updating}. The theory emphasizes that discourse comprehension requires continuous coordination of local and global coherence, maintaining a situation model that integrates textual information across distal discourse units. This insight directly inspired the context-aware summarization mechanism in the \textbf{Iterative Knowledge Integration} module, which serves as a semantic bridge across disconnected slide groups, preventing the loss of cross-group associations in fragmented long-context processing.

\noindent \textbf{Cognitive Load Theory}~\citep{sweller2011cognitive}. Sweller's Cognitive Load Theory distinguishes intrinsic, extraneous, and germane load. A well-constructed mind map reduces \textit{extraneous load} by eliminating the need for learners to infer conceptual relationships from fragmented sources, and manages \textit{intrinsic load} through progressive differentiation. Our \textbf{Structural Load Factor} and \textbf{Pedagogical Efficiency} metrics are grounded in these principles, assessing whether generated maps impose unnecessary cognitive burdens or facilitate efficient knowledge acquisition.

\section{Prompts}
\subsection{Prompts of AutoMindMap Framework}

\begin{tcolorbox}[title=Deck summarization and page grouping, breakable]
\small
You are a planner for segmenting and summarizing course slide files.
Return STRICT JSON only. \\

\#\# Input: \\
- "course\_name" \\
- "file\_name" \\
- "page\_titles": [\{"page":1,"title":""\}] \\

\#\# Output schema: \\
\{ \\
  "thinking": "", \\
  "groups": [[start\_page, end\_page], ...], \\
  "file\_summary": "" \\
\} \\

\#\# Hard constraints: \\
- Groups must be contiguous, non-overlapping, and jointly cover all pages. \\
- Each range is inclusive [start, end], 1-based index. \\

\#\# Planning guidance: \\
- Think first based on title clue. Segment by topic continuity and pedagogical transitions. \\
- Try to make each group either: 
  1) centered on one core concept, or
  2) consistent in one discourse function (e.g., overview / detailed explanation / supplementary / examples or exercises). \\
- Keep proper length. Prefer medium-size unless title transitions strongly suggest otherwise. Make the grouping flexible. \\
- Keep "file\_summary" concise but structurally useful for downstream parsing, and well-founded, including the main title and ALL significant concepts logically and coherently. \\
\end{tcolorbox}

\begin{tcolorbox}[title=Skeleton Laying, breakable]
\small
You should output an initial plan for a course mind-map construction pipeline. The mind map construction should be based on several slide decks for one course.
Return STRICT JSON only. \\

\#\# Input:\\
- "course\_name": the root node\\
- "file\_infos": [\{"file\_name","page\_count","file\_summary"\}]\\

\#\# Goal:\\
- Propose a learning order over files.\\
- Build only level-2 (L2) skeleton nodes under mind map root, assign each node files and summary.\\

\#\# Output schema:\\
\{\\
  "teaching\_order": ["file\_a", "file\_b"],\\
  "thinking": "",\\
  "level2\_nodes": [{"name": "", "summary": "", "files": ["file\_a", "file\_b"]}]\\
\}\\

\#\# Requirements:\\
- For learning order, based on filenames if clues exist. Otherwise, use your prior knowledge.\\
- L2 nodes aim to abstract a higher knowledge hierarchy over file partition. Please:\\
\hspace*{0.5em} - Cluster files with sufficient clues. Think enough first to illustrate reason. \\
\hspace*{0.5em} - Cluster is not necessary. One file can be one L2 node if the files are few or naturally with a clear knowledge partition. Don't forcibly cluster unrelated files. \\
\hspace*{0.5em} -  Keep the granularity modest and each L2 branch balanced.\\
\hspace*{0.5em} - L2 node names must be concise and meaningful domain concept noun phrases, exclude meaningless modifiers, and section index.\\
- Summary will be used as a PRINCIPLE SKILL for downstream mind map expansion. Don't just summarize clustered file content. Please generate a BRIEF plan according to the file info:\\
\hspace*{0.5em} - What should be emphasized under this L2 branch, e.g., main concepts, hierarchical and branch structures plan.\\
\hspace*{0.5em} - Potential risks may lead to disorder/illogicality that should be avoided, and suggestions to make the mind map more concise and coherent.\\
\hspace*{0.5em} - Be insightful, traceable and actionable, not broad or shallow. Think outside the box.\\
\end{tcolorbox}

\begin{tcolorbox}[title=Knowledge Extraction, breakable]
\small
You need to extract and summarize knowledge from a slide fragment "group\_pages\_md" to support mind map generation.
Regard "group\_pages\_md" as visual slides instead of a text-intensive document. 
Return STRICT JSON only.\\

\#\# Output schema:\\
\{\\
  "reading\_memory\_update": " ",\\
  "discourse\_role": " ",\\
  "concepts": [\{"name": "", "description": ""\}],\\
  "local\_relations": [\{"parent":"", "child":"", "description":""\}]\\
\}\\

\#\# Goals:\\
- "Discourse\_role": briefly summarize discourse function.\\
- "reading\_memory" is the previous fragment's summarization. Anchor on the new fragment to update a context-aware memory. Keep key content of previous knowledge, the coherence, hierarchy/affiliation with the current fragment.\\
- Extract **key concepts** that prefer few but optimal, generalized ones for knowledge overview. EXCLUDE all fine-grained and noisy marginal concepts, e.g., prerequisites, intermediate transition, examples and exercises. Keep proper granularity. \\
- Concept names should be CONCISE domain noun phrases, better to extract from the original, avoid single broad word/abbr. without referents. Exclude meaningless modifiers (e.g., section index, uninformative words). \\
- Concept description includes hierarchical context relationship inferred from "reading\_memory", description amount in this group, and the concept importance.\\
- "local\_relations" must be direct parent -> child relations only. Don't extract parallel/similar knowledge relations. \\

\#\# Visual marker in "group\_pages\_md":\\
- "[visual\_summary:...]": Textual/markdown description for embedded image.\\
- "[image]": image exists, but no description is provided.\\
\end{tcolorbox}

\begin{tcolorbox}[title=Knowledge Integration, breakable]
\small
Please integrate a batch of concepts into an existing course mind map's branch to provide a knowledge overview.
ONLY mount general key concepts, not every detailed knowledge list.
Return a STRICT JSON object only. \\

\#\# Input:\\
- "exposed\_subtree": provided node information \\
- "branch\_guideline": branch-level principle to follow\\
- "reading\_memory\_update": context-aware section summarization for reference\\
- "discourse\_role": original discourse analysis\\
- "concepts", "local\_relations": local knowledge extracted\\
- "retry\_context" (if any): previous failure tool-call log for retry\\
- "tool\_registry": tool specs\\

\#\# Goals:\\
Plan the tool-call strategy first for new concept integration, based on all context, discourse roles, and guidelines.\\
- Exclude too fine-grained, meaningless, similar concepts. Mount the most representative ones with proper granularity. \\
- Plan the most proper position in "exposed\_subtree" to mount concepts by "add\_node", make the mind map hierarchical and concise. Avoid mounting to the root node.\\
- Finally, use other tools to refine local structure if poor overload structures/wrong relations exist, but not necessary.\\

\#\# Tool rules:\\
- Check the parameters' validity. Return "tool\_calls":[] if no action.\\
- Multiple tool-calls will be executed sequentially; the subsequent step is based on preorder modification.\\
- If `retry\_context` exists, rethink and regenerate a corrected FULL `tool\_calls` (NOT from the fail step). \\

\#\# Output schema:\\
\{"thinking":"", "tool\_calls": [\{"tool":"", "args":\{\}\}]\}\\
\end{tcolorbox}

\begin{tcolorbox}[title=Cross Branch Check, breakable]
    \small
You are an agent to judge cross-branch merge and cross-link relations in a course concept mind map. Return STRICT JSON only. \\

\#\# Input:\\
- "new\_node": brief of a newly added node\\
- "candidate\_nodes": one-hop briefs from other branches\\
- "retry\_context" (if any): previous failure tool-call log for retry\\

\#\# Output schema:\\
\{"thinking":"", "tool\_calls": [\{"tool":"", "args":\{\}\}]\}\\

\#\# Goals:\\
- Merge similar concepts in most cases if the names are detailed phrases, despite different descriptions/contexts.\\
- If the candidate concept names are very broad (e.g., application/introduction/example), rename them into detailed ones.\\
- Add cross-links ONLY WHEN the hierarchical relation is very explicit and meaningful with sufficient evidence.\\

\#\# Tool rules:\\
- Check the parameters' validity. Return "tool\_calls":[] if no action.\\
- Multiple tool-calls will be executed sequentially; the subsequent step is based on preorder modification.\\
- If "retry\_context" exists, rethink and regenerate a corrected FULL "tool\_calls" (NOT from the fail step). \\
\end{tcolorbox}

\begin{tcolorbox}[title=Structure Repair, breakable]
   \small 
You are a structural repair agent for a course mind map.
Return STRICT JSON only.\\

\#\# Input:\\
- "violation\_items": local structure to be repaired \\
- "constraints"\\
- "tool\_registry"\\
- "retry\_context" (if any)\\

\#\# Output schema:\\
\{\\
  "thinking":"",\\
  "tool\_calls": [\{"tool":"", "args":\{\}\}]\\
\}\\

\#\# Policy:\\
- Think first, prefer minimal local structure edits to fix the structural violation. Better reorganize hierarchical edges, avoid frequent node deletion and addition. \\
- Keep the mind map coherent and concise. Fix only the structure, don't break the whole logic.\\

\#\# Tool rules:\\
- Check the parameters' validity. Return "tool\_calls":[] if no action.\\
- Multiple tool-calls will be executed sequentially; the subsequent step is based on preorder modification.\\
- If "retry\_context" exists, rethink and regenerate a corrected FULL "tool\_calls" (NOT from the fail step). Simplify tool-call steps if the error is complex. \\
\end{tcolorbox}

\begin{tcolorbox}[title=Branch Level Critic, breakable]
    \small
You are a critic for improving the quality of a mind map's Level 2 branch.
Return STRICT JSON only.\\

\#\# Input:\\
- "l2 node"\\
- "branch\_guideline"\\
- "l2\_branch\_count": total l2-branch in mind map\\
- "branch\_node\_count","branch\_edge\_count": for this branch\\
- "branch\_relations"\\

\#\# Output schema:\\
\{\\
  "issues": [\{"problem\_analysis":"","feasible\_suggestion":""\}]\\
\}\\

\#\# Policy:\\
- If no problem, return "issues: []".\\
- Don't focus on visual rendering problem. Focus on structural or logical flaws revealed by visual way.\\
- Mainly focus on the following problems (not limited to):\\
  \hspace*{0.5em} - Branch violates the principle in "branch\_guideline". \\
  \hspace*{0.5em} - Too dense fine-grained/meaningless nodes that can be pruned to keep proper scale.\\
  \hspace*{0.5em} - Similar intermediate node/complex local structure that can be compressed into a concise one.\\
  \hspace*{0.5em} - Poor local structure, e.g., child-node overload, long chains, dense cross-link. A deeper hierarchy is preferred over a flat one.\\
  \hspace*{0.5em} - Unclear concept name and relation (sibling and parent-child).\\
- Each issue must be concrete and actionable, focusing on a specific part. You can output several problems, but be concise.\\
- Think outside the box. Focus on obvious, objective problems rather than trivial subjective preferences.\\
\end{tcolorbox}

\begin{tcolorbox}[title=Global Level Critic, breakable]
    \small
You are the global critic for improving the course mind-map quality. Please propose meaningful suggestions for mind map refinement. Return STRICT JSON only.\\

\#\# Input:\\
- mindmap image, dash lines are cross-links\\
- "mindmap\_relations"\\
- "total\_node\_count"\\
- "total\_edge\_count"\\

\#\# Output schema:\\
\{\\
  "issues": [\{"problem\_analysis":"","feasible\_suggestion":""\}]\\
\}\\

\#\# Critic policy:\\
- If no problem, return "issues: []".\\
- Don't focus on visual rendering problem. Focus on structural or logical flaws revealed by visual way.\\
- Use "total\_node\_count" and "total\_edge\_count" as hard references to judge global density.\\
- Focus on macro structure and logic first, including but not limited to:\\
  \hspace*{0.5em} - Imbalanced branches size and depth.\\
  \hspace*{0.5em} - Too dense nodes and cross-links, propose simplification, especially marginal leaf nodes.\\
  \hspace*{0.5em} - Poor global structure. \\
  \hspace*{0.5em} - Confusing logic design fails to present the course knowledge overview efficiently.\\
- Each issue must be concrete and actionable, focusing on a specific region. You can output several problems one time, but be concise.\\
- Think outside the box. Focus on obvious, objective problems rather than trivial subjective preferences.\\
\end{tcolorbox}

\begin{tcolorbox}[title=Branch Level Editor, breakable]
    \small
Please use suggestions from an expert critic to refine one branch (under a level-2 node) of a low-quality course concept mind map. Return STRICT JSON only.\\

\#\# Input:\\
- "branch\_l2"\\
- "issues"\\
- "branch\_relations"\\
- "retry\_context" (if any): previous failure tool-call log for retry\\

\#\# Output schema:\\
\{\\
  "thinking": "", "tool\_calls": [\{"tool":"", "args":\{\}\}],\\
\}\\

\#\# Policy:\\
- Based on "issues" as overall principle to plan modification and tool-use steps strictly with focused local improvements. Prefer fewer and concise tool steps to make minimal, local edits.\\
- Priority is removing non-core nodes and cross-links if too dense. Use "delete\_node" to iteratively prune the map.\\
- Generally, avoid creating intermediate concepts/structures. Better to reorganize within existing concepts.\\
- Every change should map to a specific critical issue. Do not do broad, unrelated edits.\\

\#\# Tool rules:\\
- You can refuse unreasonable suggestions. Return "tool\_calls: []" if no action.\\
- Multiple tool-calls will be executed sequentially; the subsequent step is based on preorder modification.\\
- If "retry\_context" exists, rethink and regenerate a corrected FULL "tool\_calls" (NOT from the fail step). Simplify tool-call steps if the error is complex. \\
\end{tcolorbox}

\begin{tcolorbox}[title=Global Level Editor, breakable]
    \small
Please use suggestions from an expert critic to refine a low-quality mind map. Return STRICT JSON only.\\

\#\# Input:\\
- "issues"\\
- "mindmap\_relations"\\
- mindmap image, dash lines are cross-links\\
- "retry\_context" (if any): previous failure tool-call log for retry\\

\#\# Output schema:\\
\{\\
  "thinking": "", "tool\_calls": [\{"tool":"", "args":\{\}\}],\\
\}\\

\#\# Policy:\\
- Based on "issues" (as overall principle) to plan modification and tool-use steps strictly with focused local improvements. Prefer fewer and concise tool steps to make minimal, local edits.\\
- Priority is removing non-core nodes and cross-links if too dense. Use "delete\_node" to iteratively prune the map.\\
- Generally, avoid creating intermediate concepts/structures. Better to reorganize within existing concepts.\\
- Pay less attention to rename and non-leaf node deletion (may influence overall structure), and avoid modifying the high-level skeleton.\\
- Every change should map to a specific critical issue. Do not do broad, unrelated edits.\\

\#\# Tool rules:\\
- You can refuse unreasonable suggestions. Return "tool\_calls: []" if no action.\\
- Multiple tool-calls will be executed sequentially; the subsequent step is based on preorder modification.\\
- If "retry\_context" exists, rethink and regenerate a corrected FULL "tool\_calls" (NOT from the fail step). Simplify tool-call steps if the error is complex. \\
\end{tcolorbox}

\subsection{Prompts of Baselines}
For Direct, Chunk-Merge, and Highlights-only, we design unified prompts that include the key mind map construction principles. The Chunk-Merge method has its own prompt for merging different branches, and the Highlights-only method has its own task-specific prompts as supplements. For BookRAG and PageIndex, we use the prompts in their source codes.

\begin{tcolorbox}[title={Shared Prompt for Direct, Chunk-Merge, and Highlights-only}, breakable]
    \small
Please construct a **Course Concept Mind map** based on the provided information.\\

\#\# Task Description\\
- **Input:** Multiple Markdown files parsed from slides for one specific course (or only parts of content in slides)\\
- **Output:** A well-organized educational mind-map\\

\#\# Mind-Map Schema\\
- Radiates around from a central node (course name, will be given first). Each node represents a **concept** in the course.\\
- Edges between nodes represent the hierarchical relation, without any attributes.\\

\#\# Principles and Constraints\\
- **Well-Organized**: A fully connected directed acyclic graph (DAG), cross-links (multiple parent concepts for one concept, instead of a pure tree) are encouraged but not dense.\\
- **Identifies Core Concepts**: Don't extract all knowledge concepts, exclude too-fine-grained/marginal ones. Keep the overall concept scale concise, less confusing.\\
- **Concise node name**: Name of each node should be a course-related concept Noun phrase (or word), exclude (or rephrase) any long sentence/questions/section index/abstract expressions. Better extracted from the original and to retain the semantically indispensable words.\\
- **Coherent Hierarchy and Branch**: Keep clear, coherent hierarchy and branch structure. Build hierarchy from global to parts, abstract to concrete logically, and design branches unambiguously and consistently.\\
- **Meaningful in Education**: Your mind map should be understood quickly and efficiently, and reduce students' cognitive loads. While remaining knowledge faithfulness, keep the map brief and structurally balanced. Don't add too many fan-out nodes under one, and too deep layers. Avoid too flat hierarchies and single-node long chains.\\

\#\# Output Format (Strictly)\\
1. (Optional but encouraged) Output your analysis, thinking process, and reasons for your answer, enclosed with <think>...</think> tags.\\
2. Output your final answer, enclosed with <result>...</result> tags. The output format is as follows:\\
{}[concept\_1, sub-concept\_2], \\
{}[concept\_3, sub-concept\_4], \\
\dots \\
3. Always use the course name as the ONLY root node. Don't output any content outside the tag pair, and don't add other tags by yourself.\\
\end{tcolorbox}

\begin{tcolorbox}[title=Merge Prompt in Chunk-Merge, breakable]
\small
There are several partial mind-map branches of this course. Please merge them into one. \\
- **Merge same/similar nodes and edges**: You can identify these nodes/edges then merge them.\\
- **Check the structure constraints and refine/prune the mind-map**: If the whole merged mind map doesn't satisfy the structure requirements (e.g., many fanout nodes under one, too deep hierarchy, not a fully-connected DAG expanded from one center node), please fix the mind map, but ensure the knowledge faithfulness, don't lose the core knowledge structure.\\
- **Be coherent and logical**: Final check: let the merged mind map have a coherent logic and hierarchy, and not too dense (to reduce the user's cognitive load).\\

You can still edit local structure (e.g., prune the dense structure or re-organize hierarchy/branch) where unreasonable, to make the mind map more perfect, concise, and less confusing.\\
\end{tcolorbox}

\begin{tcolorbox}[title=Extra Prompt for Highlights-only, breakable]
\small
You will be provided with the highlighted titles of every page in all slide files under this course.\\
You need to reorganize them into a hierarchical mind-map based on just those titles and your knowledge.\\
- Note that not all titles are important core concepts or can be used directly. Try to make filtering, denoise, and rephrase, to keep the core concepts, to meet the requirements.\\
- Try to refine the mind map into a coherent one, organize a clear and hierarchical mind map from the provided information.\\
\end{tcolorbox}

\subsection{Prompts of Evaluation}
\label{app:prompteval}

\begin{tcolorbox}[title=VLM Judge, breakable]
\small
You are a strict expert VLM judge to score the quality of a course's mind maps.
Return STRICT JSON only.\\

\#\# Based on:\\
- a rendered mindmap image (rendered by an external interface, so don't evaluate rendering and typography structures). Images only help to better understand.\\
- textual relations in "parent-> child" format.\\

\#\# Detailed rubric:\\
**Content accuracy**\\
- All concept names are concise and accurate, generally proper nouns/noun phrases with 2-3 words. Poor names are avoided, including:\\
  \hspace*{0.5em} - Single abbr., broad word (e.g., introduction / motivation / application / example) without referents, or uninformative modifier words/section indexes.\\ 
  \hspace*{0.5em} - Sentences/questions/expressions, any confusing and meaningless symbols/characters. \\
  \hspace*{0.5em} - Concatenation of multiple concepts into one node.\\
- Reasonable hierarchy and affiliation relation between parent and child concept, without confusing relation between parent-child/siblings.\\
- Concepts should be representative and recapitulatory, rather than lots of unimportant, course-irrelevant, noisy entities or mutually overlapped/similar concepts.\\

**Structure quality**\\
- Poor structures should be avoided, including but not limited to:\\
  \hspace*{0.5em} - Too dense and disordered relation without clear organization.\\
  \hspace*{0.5em} - Overload child nodes under one parent without hierarchy design, or cascaded long one-child chains.\\
  \hspace*{0.5em} - Too deep (with more than 7 levels) or shallow flat hierarchy (with only 2-3 levels).\\
  \hspace*{0.5em} - Very unbalanced branch scale/depth.\\
  \hspace*{0.5em} - Shallow leaf nodes under the root.\\ 
  \hspace*{0.5em} - Multiple root, ring/bi-direct edge, isolated nodes.\\
- All local concepts and relations are organized into an overall coherent logic, with meaningful taxonomy, progressive hierarchy.\\
- Allow for proper structural flexibility and scale difference between branches, rather than a completely neat structure.\\

**Pedagogical efficiency**\\
- Proper overall scale with clear visual effect on hierarchy, rather than chaotic dense graphs,  or just superficial concepts without delving.\\
- Generalizable core topics and framework are presented obviously, can be captured quickly from the mind map rather than a laborious search.\\
- Reasonable cross-links across branches that promote concept association, rather than a pure tree. The proportion is proper rather than dense.\\
- Core knowledge should be emphasized structurally (e.g., include more and deeper nodes) rather than all concepts being equally important.\\
- Overall organization is consistent with human cognitive habits, leading to less confusion.\\

\#\# Follow **Likert scale** to grade:\\
- 5: very perfect, almost no flaws\\
- 4: obvious but non-critical flaws won't affect the overall quality  \\ 
- 3: mixed quality with notable problems, but still obey most overall principles. 3 is a pass line \\ 
- 2: severely problematic, only meet a small number of requirements\\
- 1: meet no requirement, extremely poor quality\\

**For each dimension, think and analyze sufficiently to have more evidence, then grade.**\\
**Be strict, try to grade lower scores**.\\

\#\# Output schema:\\
\{\\
  "Content accuracy":\{"think":"", "score":1\},\\
  "Overall structure":\{"think":"", "score":1\},\\
  "Practicality and efficiency":\{"think":"", "score":1\}\\
\}\\
\end{tcolorbox}

\begin{tcolorbox}[title=Concept Matching, breakable]
\small
 You are a strict concept matcher for course mindmaps. For each query, pick at most one most relevant true match.\\
 - Reject when semantic scope differs (e.g., "application" vs "AI application"). But you can use the parent context to narrow broad scope. \\
 - Minor form variants are allowed (e.g., "machine learning" vs "machine learning algorithms"), more detailed name is also allowed. \\
 - Abbreviation/full-name variants may match only if they denote the same canonical concept. Use parent context as a tie-breaker ONLY WHEN ambiguous, instead of being more strict. \\
 - If the candidate one includes index info (e.g., section 1/lecture 1), very long sentences (not concise concept), or confusing symbols, reject it. DO NOT be too lenient with academic proper nouns. \\
 
 Return JSON only:\\
\{decisions:[{gt\_node:"",accept:true|false, pred:candidate\_or\_empty}]\} \\

\end{tcolorbox}

\section{Use of Large Language Models}
All research ideas, the framework of AutoMindMap, experimental design, and core paper-writing were conceived by humans. Codex (with GPT-5.3-Codex) was used to assist programming and prompt polishing. DeepSeek-V4-Pro and GPT-5.5 were used to polish the paper, without modifying the original idea. Some icons in the figures were generated by GPT-Image-2.

\section{Ethical Statement}
All slides used are approved by the lecturers of the courses as well as the relevant academic organization of the university. We remove the slides that may potentially leak personal privacy. All data annotators are informed about the purpose of the data they annotated. The datasets and code will be released upon publication.


\end{document}